\pdfoutput=1
\documentclass[11pt]{article}

\usepackage{emnlp2022}

\usepackage{times}
\usepackage{latexsym}
\usepackage{amsmath}
\usepackage{amsfonts}
\usepackage{caption}
\usepackage{subcaption}
\usepackage{graphicx}

\usepackage{bm}
\usepackage{amsthm}

\usepackage{amsfonts}
\usepackage{amsmath,amssymb}

\usepackage[T1]{fontenc}
\usepackage[utf8]{inputenc}

\usepackage{microtype}
\usepackage[utf8]{inputenc}
\usepackage[T1]{fontenc}
\usepackage{hyphenat}
\usepackage{xspace}
\usepackage{amsmath}
\usepackage{amsfonts}
\usepackage{hyperref}
\usepackage{url}
\usepackage{booktabs}
\usepackage{multirow}
\usepackage{makecell}
\usepackage{caption}
\usepackage{minibox}
\usepackage{bbm}
\usepackage{graphicx}
\usepackage{balance}
\usepackage{mathtools}
\usepackage{color}
\usepackage{marvosym}
\usepackage{ifthen}
\usepackage{textcomp}
\usepackage{enumitem}
\usepackage{verbatim}
\usepackage{algorithm}
\usepackage{numprint}
\usepackage{balance}

\usepackage{amsthm}
\theoremstyle{plain}

\newcommand{\chatoDisplayMode}[1]{#1}

\definecolor{MyRed}{rgb}{0.6,0.0,0.0} 
\definecolor{MyBlack}{rgb}{0.1,0.1,0.1} 
\newcommand{\inred}[1]{{\color{MyRed}\sf\textbf{\textsc{#1}}}}
\newcommand{\frameit}[2]{
  \begin{center}
  {\color{MyRed}
  \framebox[.9\columnwidth][l]{
    \begin{minipage}{.85\columnwidth}
    \inred{#1}: {\sf\color{MyBlack}#2}
    \end{minipage}
  }\\
  }
  \end{center}
}

\newcommand{\note}[2][]{\chatoDisplayMode{\def\@tmpsig{#1}\frameit{{\Pointinghand} Note}{#2\ifx \@tmpsig \@empty \else \mbox{ --\em #1}\fi}}}
\newcommand{\todo}[2][]{\chatoDisplayMode{\def\@tmpsig{#1}\frameit{{\Writinghand} To-do}{#2\ifx \@tmpsig \@empty \else \mbox{ --\em #1}\fi}}}

\newcommand{\Secref}[1]{Sec.~\ref{#1}}
\newcommand{\Eqnref}[1]{Eq.~\ref{#1}}

\newcommand{\Figref}[1]{Fig.~\ref{#1}}

\newcommand{\xhdr}[1]{\vspace{1.7mm}\noindent{{\bf #1.}}}

\newcommand{\textcite}[1]{\citeauthor{#1} \shortcite{#1}}

\newcommand{\hide}[1]{}

\makeatletter
\newcommand{\iffont}[2]{\ifthenelse{\equal{\f@family}{#1}}{#2}{}}
\makeatother

  \usepackage{mathptmx}

  \DeclareSymbolFont{greek}{OML}{cmm}{m}{n}
  \DeclareMathSymbol{\alpha}{\mathalpha}{greek}{"0B}
  \DeclareMathSymbol{\beta}{\mathalpha}{greek}{"0C}
  \DeclareMathSymbol{\gamma}{\mathalpha}{greek}{"0D}
  \DeclareMathSymbol{\delta}{\mathalpha}{greek}{"0E}
  \DeclareMathSymbol{\epsilon}{\mathalpha}{greek}{"0F}
  \DeclareMathSymbol{\zeta}{\mathalpha}{greek}{"10}
  \DeclareMathSymbol{\eta}{\mathalpha}{greek}{"11}
  \DeclareMathSymbol{\theta}{\mathalpha}{greek}{"12}
  \DeclareMathSymbol{\iota}{\mathalpha}{greek}{"13}
  \DeclareMathSymbol{\kappa}{\mathalpha}{greek}{"14}
  \DeclareMathSymbol{\lambda}{\mathalpha}{greek}{"15}
  \DeclareMathSymbol{\mu}{\mathalpha}{greek}{"16}
  \DeclareMathSymbol{\nu}{\mathalpha}{greek}{"17}
  \DeclareMathSymbol{\xi}{\mathalpha}{greek}{"18}
  \DeclareMathSymbol{\pi}{\mathalpha}{greek}{"19}
  \DeclareMathSymbol{\rho}{\mathalpha}{greek}{"1A}
  \DeclareMathSymbol{\sigma}{\mathalpha}{greek}{"1B}
  \DeclareMathSymbol{\tau}{\mathalpha}{greek}{"1C}
  \DeclareMathSymbol{\upsilon}{\mathalpha}{greek}{"1D}
  \DeclareMathSymbol{\phi}{\mathalpha}{greek}{"1E}
  \DeclareMathSymbol{\chi}{\mathalpha}{greek}{"1F}
  \DeclareMathSymbol{\psi}{\mathalpha}{greek}{"20}
  \DeclareMathSymbol{\omega}{\mathalpha}{greek}{"21}
  \DeclareMathSymbol{\varepsilon}{\mathalpha}{greek}{"22}
  \DeclareMathSymbol{\vartheta}{\mathalpha}{greek}{"23}
  \DeclareMathSymbol{\varpi}{\mathalpha}{greek}{"24}
  \DeclareMathSymbol{\varrho}{\mathalpha}{greek}{"25}
  \DeclareMathSymbol{\varsigma}{\mathalpha}{greek}{"26}
  \DeclareMathSymbol{\varphi}{\mathalpha}{greek}{"27}
  \DeclareSymbolFont{otone}{OT1}{cmr}{m}{n}
  \DeclareMathSymbol{\Gamma}{\mathalpha}{otone}{0}
  \DeclareMathSymbol{\Delta}{\mathalpha}{otone}{1}
  \DeclareMathSymbol{\Theta}{\mathalpha}{otone}{2}
  \DeclareMathSymbol{\Lambda}{\mathalpha}{otone}{3}
  \DeclareMathSymbol{\Xi}{\mathalpha}{otone}{4}
  \DeclareMathSymbol{\Pi}{\mathalpha}{otone}{5}
  \DeclareMathSymbol{\Sigma}{\mathalpha}{otone}{6}
  \DeclareMathSymbol{\Upsilon}{\mathalpha}{otone}{7}
  \DeclareMathSymbol{\Phi}{\mathalpha}{otone}{8}
  \DeclareMathSymbol{\Psi}{\mathalpha}{otone}{9}
  \DeclareMathSymbol{\Omega}{\mathalpha}{otone}{10}
  \DeclareSymbolFont{syms}{OML}{cmm}{m}{it}
  \DeclareMathSymbol{\partial}{\mathord}{syms}{"40}
  \DeclareMathAlphabet{\mathbold}{OML}{cmm}{b}{it}
  \DeclareSymbolFont{largesymbols}{OMX}{cmex}{m}{n}

\title{The Glass Ceiling of Automatic Evaluation\\in Natural Language Generation}

\author{
   Pierre Colombo$^{1}$, Maxime Peyrard$^2$, Nathan Noiry$^3$ \\ \textbf{Robert West}$^2$ \qquad \textbf{Pablo Piantanida}$^4$
   \\ $^1$MICS, CentraleSupélec, Université Paris-Saclay, $^2$ EPFL, $^3$ althiqa.io \\
    $^4$ ILLS,
McGill - ETS - MILA - CNRS - Université Paris-Saclay - CentraleSupélec\\
    \texttt{colombo.pierre@centralesupelec.fr}}

\begin{document}
\maketitle
\begin{abstract}
Automatic evaluation metrics capable of replacing human judgments are critical to allowing fast development of new methods. Thus, numerous research efforts have focused on crafting such metrics. In this work, we take a step back and analyze recent progress by comparing the body of existing automatic metrics and human metrics altogether. As metrics are used based on how they rank systems, we compare metrics in the space of system rankings. Our extensive statistical analysis reveals surprising findings: automatic metrics -- old and new -- are much more similar to each other than to humans. Automatic metrics are not complementary and rank systems similarly. Strikingly, human metrics predict each other much better than the combination of all automatic metrics used to predict a human metric. It is surprising because human metrics are often designed to be independent, to capture different aspects of quality, e.g. \emph{content fidelity} or \emph{readability}. We provide a discussion of these findings and recommendations for future work in the field of evaluation.
% Automatic evaluation metrics capable of replacing human judgments are critical to allowing fast development of new methods. Thus, numerous research efforts have focused on crafting such metrics. In this work, we take a step back and analyze recent progress by comparing the body of existing automatic metrics and human metrics altogether. Importantly, we compare metrics within the space of rankings, i.e., based on how metrics rank systems. Indeed, \emph{rankings} are more robust to statistical variations but also more aligned with the intended usage of automatic metrics, namely to rank systems by performance. Our extensive statistical analysis reveals surprising findings: automatic metrics -- old and new -- are much more similar to each other than to humans. Recent transformer-based metrics mostly capture the same \emph{simple} dimension of human judgments as the older, surface-level metrics. \textcolor{red}{\# I am not sure this sentence should be here, this could be misunderstood  \# For future researchers, we make the recommendation to craft metrics not only to maximize correlation with humans but also to minimize similarity with existing metrics. }This will ensure that future metrics contribute to  bridging the gap between automatic metrics and humans.
\end{abstract}

% what is missing to fill the gap? between human and automatic.

% Introduction (comment on compare les systems)
% choix methodologiques
% kemenny conscencus, tau de kendall
% utterance + system (difference entre les deux).
% cas extremes (de l'utterance)

% experience qu'on fait 
% mds
% clustering
% regression 

\section{Introduction}
% \paragraph{Problem and why it matters}
% \todo{Add citations throughout the intro}

Crafting automatic evaluation metrics (AEM) able to replace human judgments is critical to guide progress in natural language generation (NLG), as such automatic metrics allow for cheap, fast, and large-scale development of new ideas. The NLG fields are then heavily influenced by the set of AEM used to decide which systems are valuable. Therefore, a large body of work has focused on improving the ability of AEM to predict human judgments.

% \begin{itemize}
%     \item Crafting evaluation that can replace human judgments is critical to the improvement of the field
%     as automatic metrics allow cheap, fast and large scale development and testing of new ideas.
    
%     \item The NLG fields are heavily influenced by the set of automatic metrics used to decided which systems are valuable. 
    
%     \item Therefore, there has been a lot of work trying to (a) develop new and better automatic metrics, (b) understand how to choose the best automatic metrics.
% \end{itemize}

\begin{figure}[t]
    \centering
    \includegraphics[width=0.98\linewidth]{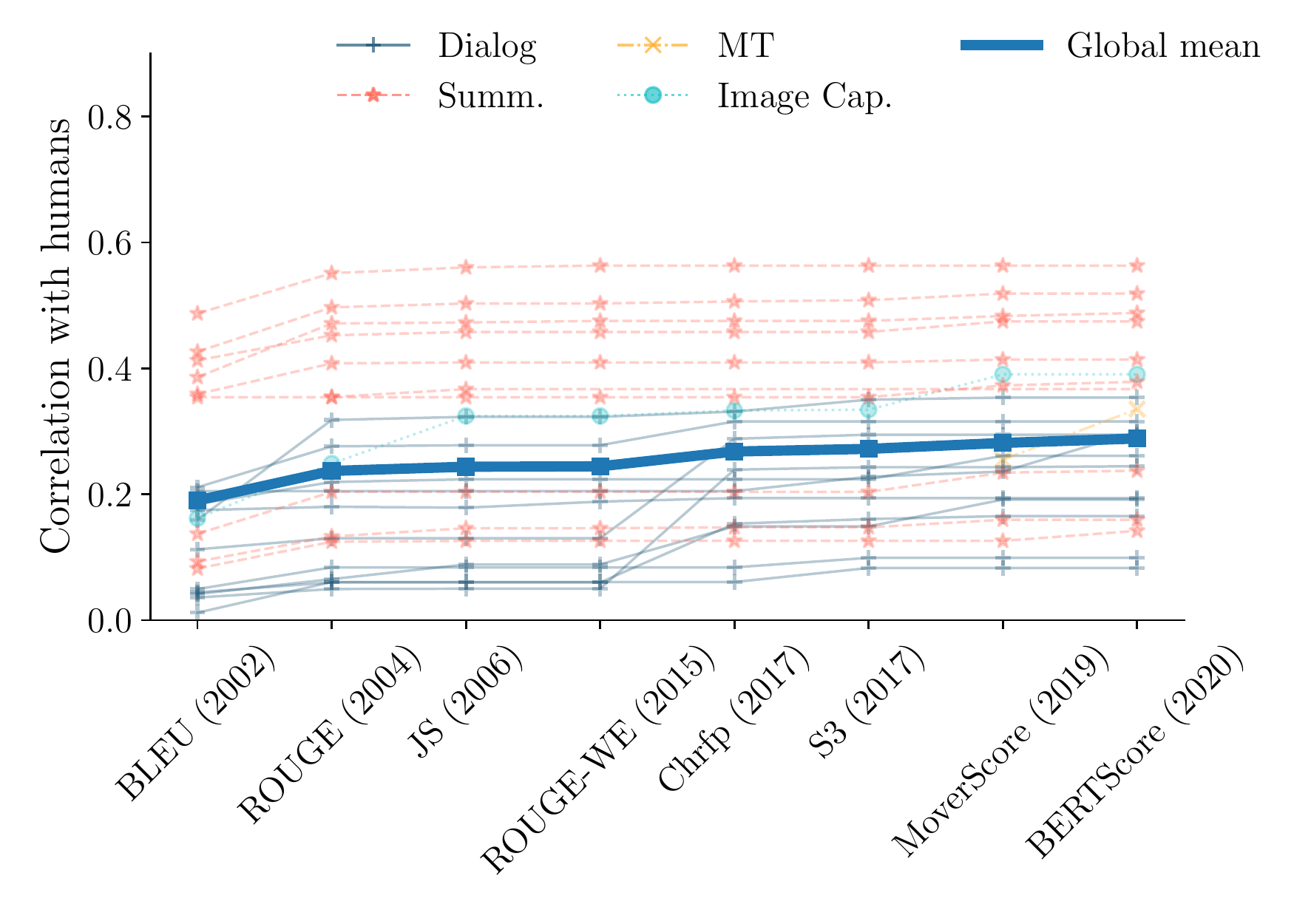}\vspace{-0.5cm}
    \caption{\textbf{Correlation with humans over time considering all existing metrics combined.} On the x-axis: evaluation metrics ordered by their release time; y-axis: utterance-level Kendall's $\tau$ with human when training a model to fit human judgments with all metrics available at the time (5-Fold cross-validation with XGBoost regressor). The dotted lines represent different human annotations and datasets. Different variants of the same metrics (like ROUGE-1 and ROUGE-2) are averaged. The datasets and metrics are described in \Secref{02_methodology}.}
    \label{fig:fig1}\vspace{-0.5cm}
\end{figure}

Human judgment data is typically employed to decide which metric to select based on correlation analysis with human annotations \cite{rankel-etal-2011-ranking, Owczarzak:2012,Graham:2015}.
In this work, we take a step back and investigate the relationship between existing AEM and human judgments globally. We do not make metric recommendation but reflect upon the global progress in the field of automatic evaluation. 
Our work is motivated by the findings of \Figref{fig:fig1}. It depicts the improvement over time, when new metrics were introduced, in the ability to fit human judgments when using all existing metrics as features. The fit is measured by the correlation with humans of a trained classifier in a 5-fold cross-validation setup. Surprisingly, we observe small marginal improvement and little progress over the years.

% \paragraph{Main idea of this work}
% \begin{itemize}
%     \item In this work, we propose to take a step back and investigate the relationship between existing automatic metrics and human judgments.
    
%     \item Previous works typically rely on human judgment data to compare correlations between evaluation metrics and some human metric of interest, to decide which metric is best. 
    
%     \item Here we look at the body of existing metrics (both human and automatic) and do not aim to make a new metric recommendation. Instead, we propose to analyze what are the relationship between existing automatic metrics and human metrics and reflect about the global progress of the field. 
    
%     \item Our analysis is motivated by the findings of \Figref{fig:fig1}. It depicts the improvement over time, when new metrics are introduced, in the ability to fit human judgments when using all existing metrics as features. This shows fairly small overall improvement when considering different datasets and different dimensions of human judgments. 
% \end{itemize}

Recent works emphasized the importance of viewing metrics in terms of how they rank systems instead of just comparing score values  \cite{novikova-etal-2018-rankme, peyrard2021better,colombo2022best}. Indeed, not only ranking is a more robust framework of comparison, it is also more aligned with the way metrics are used: identifying and extracting the "best system". Thus, we perform our analysis in the space of rankings. i.e., how do metrics rank systems? By analyzing 9 datasets covering 4 tasks and 270k scores, we made the following observations:

% \paragraph{What is done, why is it new}
% \begin{itemize}
%     \item Inspired from recent work that emphasized the importance of ranking-based measurements, we perform our analyses not directly by comparing automatic and human scores but by comparing metrics in the space of rankings, i.e., how do metrics rank systems? 
%     Indeed, not only ranking is a more robust framework of comparison, it is also more aligned with the way metrics are used: identifying and extracting the "best metric".
%     \item By analyzing xx datasets covering xx tasks, xx systems and xx judgments, we made the following observations:
% \end{itemize}
\noindent\textbf{Findings.} (i) Automatic metrics are much more similar to each other, in terms of how they rank systems, than they are to human metrics. It means that AEM, even the more recent transformer-based ones are similar to the older ones when used in practice (ROUGE and BLEU). 
(ii) This lack of complementarity results in the inability to fit human judgments even when all these metrics are taken together as features for a classifier predicting humans. 
(iii) Quite surprisingly, different human dimensions -- different annotations guidelines such as \emph{readability}, or \emph{content fidelity} -- are very predictive of each other, whereas AEM are much less predictive of humans. 
This finding is striking because human metrics are designed to capture different and independent aspects of quality whereas AEM have been selected precisely for their ability to match humans. We would expect human metrics to be uncorrelated and automatic metric to be highly correlated with humans but we observe the opposite. First, it casts serious doubt about the ability of AEM to replace human judgments. Then, the correlation between independent human annotations of quality hints at some latent inherent \emph{goodness} of systems: good systems are good in different aspect whereas bad systems are bad across all aspects. % It is particularly striking to observe that human metrics are, in fact, quite predictive of each other while automatic metric are much less predictive of humans.
Our findings have several consequences that can inform future research. Newly introduced metrics are not complementary to previous ones, resulting in small global improvements. As a way forward, we propose that research, instead of crafting metrics that maximize correlation with humans, focus on making metrics that \textbf{also} aim to be explicitly complementary to the set of existing metrics. This would enforce maximal marginal gain and ensure that the field, as a whole, makes progress towards capturing the complexity of human annotations.

 For practitioners, it is common practice to report several AEM in the hope to get a better view of system performances. However, reporting several metrics that all produce similar rankings does not bring useful additional information. With our proposal, reporting a set of complementary metrics would better serve the intended purpose.

To help research build upon our work and use our measure of complementarity, we make our code available at \url{github}.

\section{Methodology}\label{02_methodology}
\textbf{Terminology.}
%Notations + introduction of the notion of seeing human judgments as human metrics.
Let $\mathcal{X}$ be the space of possible outputs for an NLG task. 
An NLG metric is a function $m: \mathcal{X} \times \mathcal{X} \rightarrow \mathbb{R}_+$ which, from a given textual candidate $C \in \mathcal{X}$ and corresponding reference $R \in \mathcal{X}$, computes a score $m(C,R)$ reflecting the properties that $C$ should satisfy ({\it e.g.} fluency, fidelity...). Of course, it is illusory to summarize subtle semantic properties by a single scalar and one is rather seeking for metrics that are able to discriminate between different systems. In fact, crafted AEM are evaluated by comparison to human judgments: one usually computes ranking correlations such as the Kendall's $\tau$. 
Higher correlations indicating that the AEM is a better replacement for the human metrics. 
% The more an automatic metric returns the same rankings as a human metric, the more relevant it is for automatic NLG evaluation.

\xhdr{Encoding metrics with rankings} 
Since the usage of NLG metrics is to rank systems, we choose to represent an NLG metric, automatic or human, by the ranking it induces on a set of systems or of utterances. More formally, for $N \geq 1$ NLG systems evaluated on a dataset made of $K \geq 1$ utterances, there exists a natural ranking representations of $m$:

Each utterance $k \in \{1, \ldots, K\}$ induces a ranking $\sigma_k^m \in \mathbb{R}^N$ of the $N$ systems seen as a vector $\sigma_{k}^m$, where $\sigma_k^m(S)$ is the rank of system $S \in \{1, \dots, N\}$. 
For a system $S$, the representation of a metric $m$, noted $\sigma^{m, S}$, is sum of \emph{rankings over the utterances}:
    \begin{equation} \label{eq:sys-level}
        \sigma^{m, S} := \sum\limits_{k=1}^K \sigma_k^m(S) \in \mathbb{R}^N. 
    \end{equation}  
We call this {\bf System level representation.} 

\noindent Symmetrically, each system $k \in \{1, \ldots, N\}$ induces a ranking $\sigma_n^m \in \mathbb{R}^K$ of the $K$ utterances, where $\sigma_n^m(k)$ is the rank of utterance $k$. The {\bf Utterance level representation} of $m$ is sum of rankings over the systems:
    \begin{equation} \label{eq:ut-level}
         \sigma_{utt}^m := \sum\limits_{n=1}^N \sigma_k^n \in \mathbb{R}^K. 
     \end{equation}

%the {\bf Utterance level representation} can be defined (see \autoref{section:sup_etendred_metho}). Due to space limitation results on Utterance level representation can be found in Appendix.
Using the space of rankings has been shown to be more robust than the raw scores as it is less sensitive to outliers and statistical variations \cite{novikova2017we, peyrard2021better, colombo2022best}. Furthermore, this representation is closely tied to Borda counts, which enjoys theoretical properties: the ranking induced by $\sigma^{m,S}$ is a $5$-approximation of the Kemeny-consensus which is a good notion of average in the symmetric group \cite{kemeny1959mathematics, young1978consistent, coppersmith2006ordering}. It is moreover the fastest approximation of the Kemeny-consensus whose computation is NP-hard \cite{ali2012experiments}. 

% \todo{quick high-level summary and citation of the Kemeny consensus paper. \cite{novikova2017we}
% Borda Count Representation relying on ranks to represent the scores}

\xhdr{Complementarity}
We measure the \emph{complementarity} between two metrics -- humans or automatic -- by the average over utterances of the distance between their rankings of systems. Formally, for two metrics $m_0$ and $m_1$, complementarity is given by:
\begin{equation}
    C(m_0, m_1) := \frac{1}{K} \sum\limits_{k = 1}^K d_{\tau}(\sigma_{k}^{m_0}, \sigma_{k}^{m_1}),
    \label{eq:comple}
\end{equation}
where $d_{\tau}$ is the normalized Kendall's distance between the vectors of rank. It is related to the Kendall's rank correlation $\tau$ by: $\tau = 1-2 d_{\tau}$.

Similarly, we define the complementarity between a metric $m_0$ and a set of other metrics $\boldsymbol{m} := \{m_i\}_{i = 1, \dots, l}$, as the average pairwise complementarity:
\begin{equation}
    C(m_0, \boldsymbol{m}) = \frac{1}{l}\sum\limits_{i = 1, \dots, l} C(m_0, m_i).
    \label{eq:comple_set}
\end{equation}
Complementarity measures the extent to which a metric ranks systems differently than another metrics or a set of other metrics. Whether comparing two metrics or a metric with set, it is a number between 0 and 1 where 0 indicates that the metrics rank systems in the exact same order and 1 indicates the exact opposite order. In between, it counts the number of inversions between the two rank lists normalized by the number of possible pairs of systems.

\subsection{Dataset description}\label{sec:dataset} To ensure a wide coverage of NLG we focus on four different problems \textit{i.e.}, dialogue generation (using PersonaChat (PC) and TopicalChat (TC) \cite{mehri2020usr}), image description (relying on FLICKR \cite{young2014image}), summary evaluation (via TAC08 \cite{dang2008overview}, TAC10,  TAC11 \cite{owczarzak2011overview}, RSUM \cite{bhandari2020re} and SEVAL \cite{fabbri2021summeval}), and translation (focusing on multilingual quality estimation (MLQE) \citet{ranasinghe2021exploratory}). 

\noindent For each task, we gather datasets and rely on AEM such as  JS [1-2] \cite{lin2006information}, BLEU \cite{bleu,sacrebleu}, Chrfpp \cite{popovic2017chrf++}, S3 (both variant pyr/resp) \cite{peyrard2017learning}, ROUGE \cite{lin2004rouge} (including 5 of its variants \cite{ng2015better}),  BERTScore \cite{zhang2019bertscore}, MoverScore \cite{zhao2019moverscore}.  For MLQE we solely consider several version of BERTScore, MoverScore and ContrastScore. The human evalutions criterion are specific to each dataset and will be identified by starting with an H:. Overall, our final datasets gather over 270k scores.

\section{Experiments}
\begin{figure}
    \centering
    \includegraphics[width=0.98\linewidth]{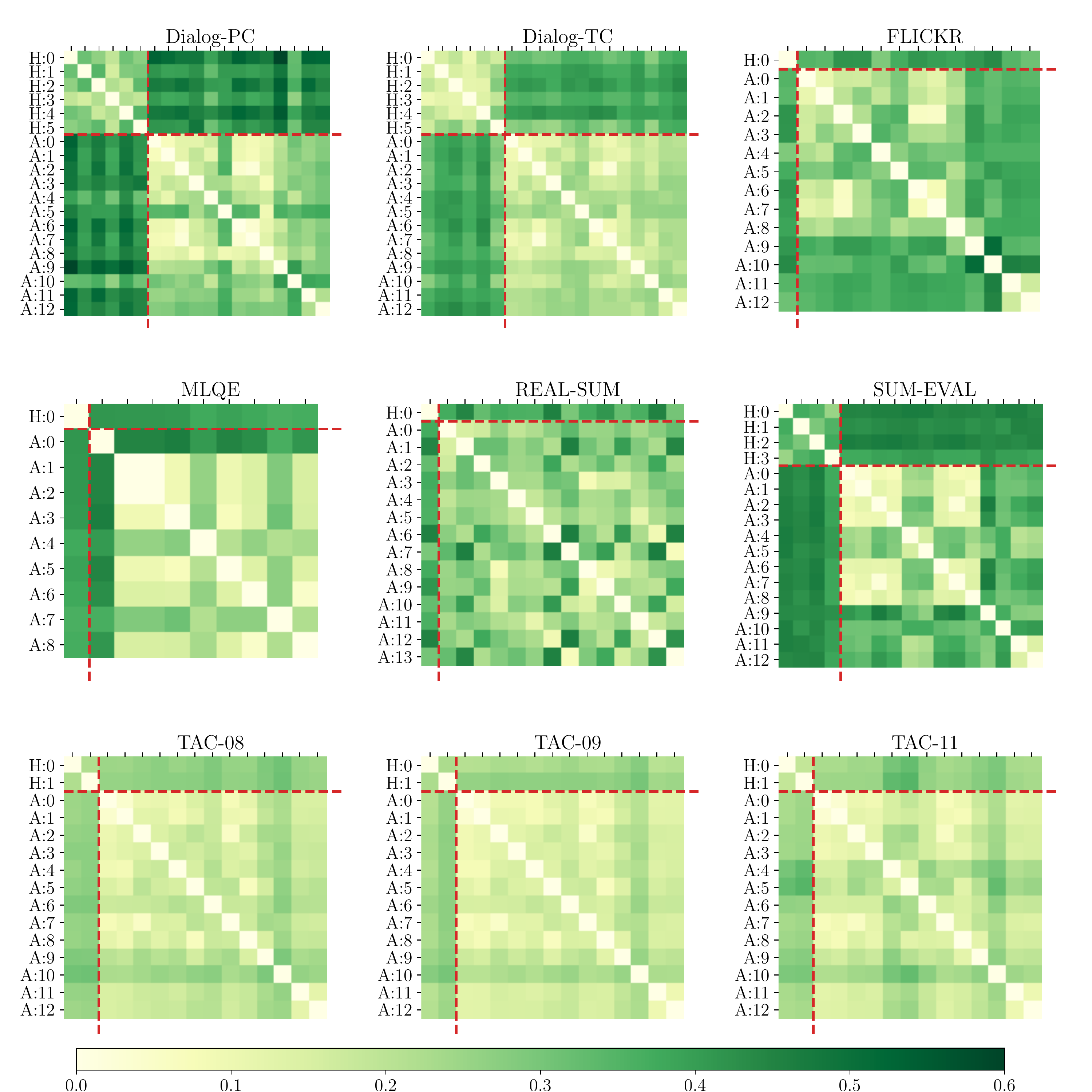}
    \caption{\textbf{Complementarity}: For each dataset, the pairwise complementarity between each pair of metrics as computed by \Eqnref{eq:comple} both human and automatic. In these matrix plot, symmetric by design, we ordered metrics to have the human one first and the automatic ones after, the red lines trace the limit between humans and AEM. % Detailed legend in \Appref{}.
    }
    \label{fig:global_pairwise}
\end{figure}
% \todo{Fix legend in the plot}

\xhdr{Finding 1: Automatic metrics are similar to each other much more than they are to human metric} In \Figref{fig:global_pairwise}, we report the pairwise complementarity between each pair of metrics as computed by \Eqnref{eq:comple} for both human and AEM. 
When aggregated over pairs and over datasets, we obtain an average complementarity between: (i) two human metrics of $.16 \pm .01$, (ii) two AEM of $.20 \pm .01$  and (iii) a human and an automatic metric of $.35 \pm .02$.

Importantly, we observe across datasets low complementarity, i.e., strong similarity, between AEM, low complementarity between human metrics but high complementarity, i.e., low similarity, between automatic and human metrics. 

We draw two conclusions from this analysis: (i) AEM rank systems similarly but (ii) differently than humans. 
There is some nuances across datasets. The effect described above is particularly strong in the Dialog, MLQE and SUM-Eval datasets.
In particular, we notice that TAC datasets, from the summarization task, have lower complementarity in general, meaning that all metrics, human and automatic, are more similar. Indeed, a lot of works have relied on these datasets to develop new metrics. Interestingly, the more recent REAL-SUM and SUM-Eval reveal much lower metric similarity.

\begin{figure}
    \centering
    \includegraphics[width=0.95\linewidth]{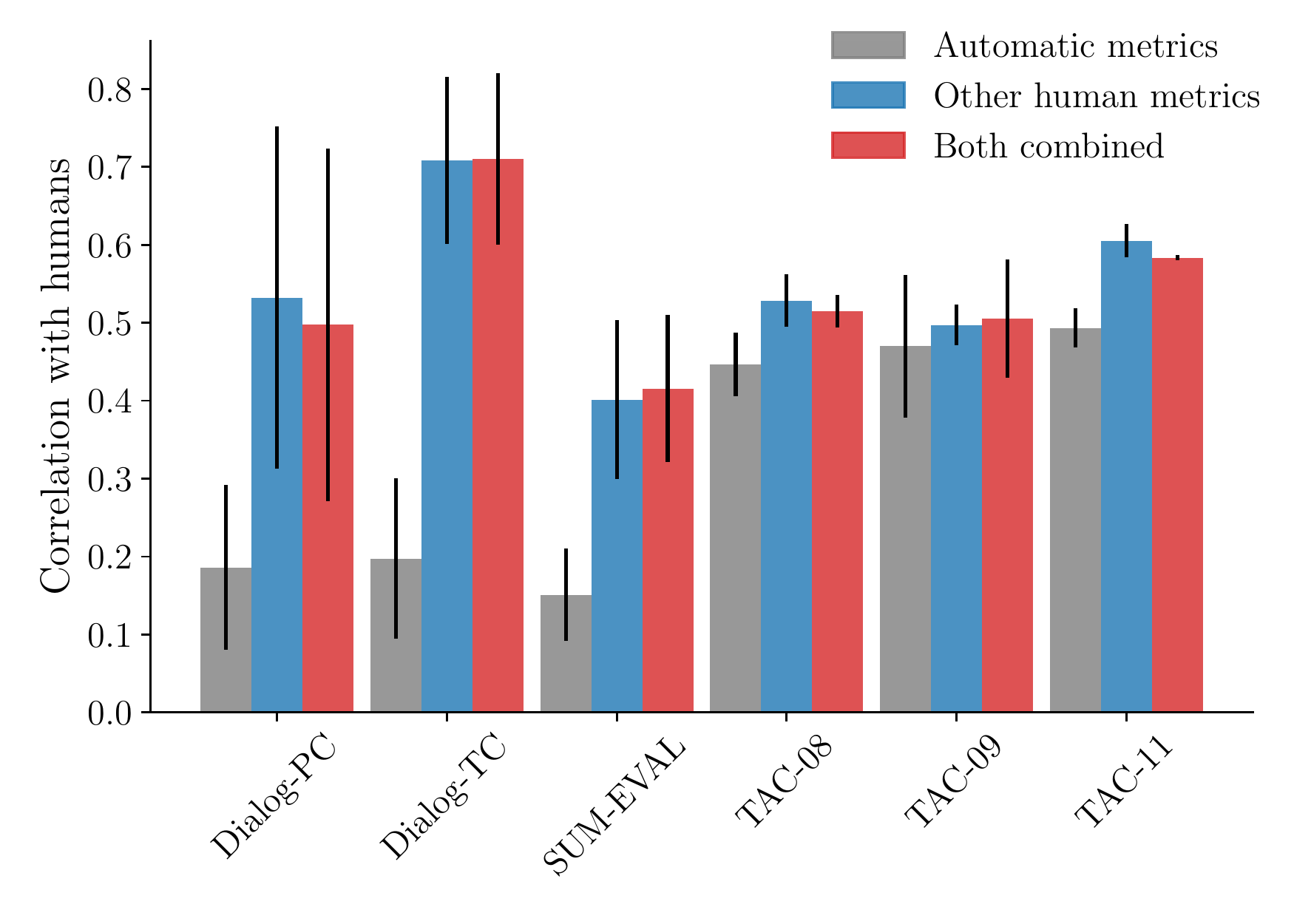} % 87
    \caption{\textbf{Human metrics are significantly more predictive of each other than AEM.} 
    On this plot, we report the 5-fold cross-validated result of fitting an XGBoost regressor on various feature sets: (i) all available AEM, (ii) other human metrics when available, and (iii) both automatic and human metrics. The fit is measured as the average instance-level correlation in the test set.
    % On this plot, we report the ratio between the MSE-error of a linear regression trained with AEM together with human metrics and a linear regression trained only with AEM. For each dataset, we provide mean and variance corresponding to the prediction of available human metrics.
    \vspace{-0.5cm}
    }
    \label{fig:predicted_power}
\end{figure}

% \begin{figure}
%     \centering
%     \includegraphics[width=0.48\linewidth]{figures_main/all_SUM_EVAL.csv_borda_Hcoherence_system.pdf}\includegraphics[width=0.48\linewidth]{figures_main/all_SUM_EVAL.csv_borda_Hfluency_system.pdf}
%     \caption{\textbf{Human metrics are more use-full than AEM when predicting other human metrics.}. On this plot we report the weights (y-axis) obtained by each metric when training a Lasso Regression for different regularization coefficients (x-axis). }
%     \label{fig:lasso_plots}
% \end{figure}

\xhdr{Finding 2: Automatic metrics even all combined do not explain human metrics} 
If AEM are rather different than human metrics, we might wonder whether it is possible to get a good approximation of human judgments by combining existing AEM together.
% \indent 
To account for possible correlations, we rely on XGBoost regressors with 5-fold cross-validation to predict human judgments. The training is performed on three different features space: {\it (i)} AEM only, {\it (ii)} other human metrics only and {\it (iii)} both sets of metrics combined. We compute the Kendall's $\tau$ between predictions and ground truths and report the results in \Figref{fig:predicted_power}. \\
% \indent 
The plot confirms that AEM struggle to capture human judgment subtlety: correlation rarely exceeds .4 on held-out data. In contrast, human metrics are much more predictive of each others, even if they are often supposed to capture different concepts. Finally, it is worth noting that adding AEM to human ones do not marginally improve the prediction power.

%In \Figref{fig:predicted_power}, we train an XGBoost regressor with 5-fold cross-validation to predict human judgment on several datasets. We 

%As shown in \Figref{fig:predicted_power}, even when combining all available metrics into a regression model, correlations with human judgments remain fairly small on held-out data.
%There are datasets and dimensions of human judgments that are still difficult to capture for automatic metrics. Even the in the best cases, correlation on held-out data rarely exceeds .4 of Kendall's $\tau$ with humans.

%Furthermore, in \Figref{fig:}, we report the ability to fit human judgments while varying the set of features: (i) using all available AEM, (ii) using other human metrics when available, (iii) using both all automatic and other human metrics. 
%As in \Figref{fig:fig1} we use 5-fold cross-validation with an XGBoost regressor and measure the instance-level correlation with the target human metric on the test set.
%This plot confirms that AEM, even when combined together, struggle to fit human metrics.
%Interestingly, other human metric are significantly more predictive of each other than AEM (in 5 out of 6 datasets) even though human metrics are usually designed to capture different, independent, aspects of quality. Finally, adding automatic metrics on top of human ones does not bring any benefit.

These findings cast shadows over recent progress in the field. In next section, we discuss the implications and make a proposition for future work.

\section{Discussion}
Our analysis reveals that automatic metrics are not complementary, and recent automatic metrics actually capture the same properties of human judgments as older ones. 
Furthermore, the existing metrics are not strong predictors of human judgments. Quite surprisingly, other human metrics which are often designed to be independent of each other end-up being more predictive of each other than automatic metrics. This predictability of human metrics from one another can be explained due to the available datasets: when a system is good at extracting content, it is also often good at making the content readable, when a system is bad it is often bad across the board in all human metrics.
However, the fact that automatic metrics are less predictive than other human dimensions casts some shadow over recent progress in the field.
It shows that the current strategy of crafting metrics with slightly better correlation than baselines with one of the human metrics has reached its limit and some qualitative change would be needed.

A promising strategy to address the limitations of automatic metrics is to report several of them, hoping that they will together give a more robust overview of system performance. However, this makes sense only if automatic metrics measure different aspects of human judgments, i.e., if they are complementary. In this work, we have seen that metrics are in fact not complementary, as they produce similar rankings of systems.

\noindent \textbf{Proposition for future work}
To foster meaningful progress in the field of automatic evaluation, we propose that future research craft new metrics not only to maximize correlation with human judgments but also to minimize the similarity with the body of existing automatic metrics. This would ensure that the field progresses as whole by focusing on capturing aspects of human judgments that are not already captured by existing metrics.
Furthermore, the reporting of several metrics that have been demonstrated to be complementary could become again a valid heuristic to get a robust overview of model performance.
In practice, researchers could re-use our code and analysis to enforce complementarity by, for example, enforcing new metrics to have low complementarity as measured by \Eqnref{eq:comple}.

% \begin{itemize}
%     \item Currently reporting several metrics in the hope that they give a more robust overview of system performances
%     \item However, new metrics introduced are not very complementary to the previous ones, when creating a new metrics we tend to relearn what was already covered by previous metrics
%     \item Our analysis in the space of rankings shows all these automatic metrics even together do not explain well choices made by humans.
%     \item \textbf{Proposition}: We would propose, to craft new metrics not only to maximize correlation with human judgments but also to minimize similarity with the body of existing automatic metrics.
%     \item This would ensure that the field progresses as a whole and make the reporting of several automatic metrics much more meaningful if the chosen metrics are proven to be complementary.
% \end{itemize}

% \begin{itemize}
%     \item In future work, these analysis could be repeated on new datasets with incoming evaluation metrics and evaluation strategies to re-assess the state of the field.
% \end{itemize}

\section{Limitations}
Even though we have considered a representative set of automatic evaluation metrics, new ones are constantly introduced and could be added to such an analysis. Similarly, new datasets could be added to the analysis and impact the results. 
In an effort to make our findings relevant in the long run, we release an easy-to-use code base to replicate our analysis with new metrics and datasets.

Like the majority of analysis on automatic evaluation metrics, ours rely on the assumption that human judgments are valid and meaningful. However, some works have questioned the quality of human judgments in standard datasets.

% \begin{itemize}
%     \item Even though we considered a large and representative set of automatic evaluation metrics, new ones are constantly proposed and could be added to such an analysis. A similar limitation holds for datasets. In order to make our findings relevant in the long run, we release an easy-to-use code to reproduce our findings with other automatic metrics and datasets
%     \item Like the majority of analysis on automatic evaluation metrics, ours rely on the assumption that human judgments are valid and meaningful. Some works have questioned the quality of human judgments in standard datasets.
% \end{itemize}

\section*{Acknowledgments}
This work was also granted access to the HPC resources of IDRIS under the allocation 2021-AP010611665 as well as under the project 2021-101838 made by GENCI. This work has been supported by the project PSPC AIDA: 2019-PSPC-09 funded by BPI-France.

\bibliography{anthology,custom}
\bibliographystyle{acl_natbib}

\appendix

\onecolumn
\section{Extended Methodology}\label{section:sup_etendred_metho}

\subsection{Utterance level Representation}
In the main paper, we focus on {\bf System level representation.} Each utterance $k \in \{1, \ldots, K\}$ induces a ranking $\sigma_k^m \in \mathbb{R}^N$ of the $N$ systems, where $\sigma_k^m(n)$ is the rank of system $n$. The system level representation of $m$ is the sum of rankings over the utterances:
     \begin{equation} \label{eq:sys-level}
         \sigma_{sys}^m := \sum\limits_{k=1}^K \sigma_k^m \in \mathbb{R}^N. 
     \end{equation}  

In the supplementary, we also provide an analysis at {\bf Utterance level representation.} Each system $k \in \{1, \ldots, N\}$ induces a ranking $\sigma_n^m \in \mathbb{R}^K$ of the $K$ utterances, where $\sigma_n^m(k)$ is the rank of utterance $k$. The utterance level representation of $m$ is the sum of rankings over the systems:
    \begin{equation} \label{eq:ut-level}
         \sigma_{utt}^m := \sum\limits_{n=1}^N \sigma_k^n \in \mathbb{R}^K. 
     \end{equation}

\subsection{A remark on the rank representations}

 For a given family of $l\geq 1$ objects, the formal mathematical object describing a ranking is a permutation $\sigma \in \mathfrak{S}_l$ which describes how the objects must be interchanged to be ordered. The set of permutations is a group where the notion of mean is not straightforward since the addition of two permutations is not a well defined object. For a given family $\sigma_1, \ldots, \sigma_p$, the classical surrogate is called a {\bf Kemeny consensus}, defined by:
 \begin{equation}
\sigma^* \in \underset{\sigma \in \mathfrak{S}_l}{\mathrm{argmin}} \sum\limits_{i=1}^p d(\sigma_i, \sigma),
  \end{equation}
 where $d$ the Kendall distance, given by:
 %\[
  \begin{equation}
 d(\eta, \tau) := \sum_{1\leq i,j \leq N} \mathbf{1}_{ (\eta_i - \eta_j)( \tau_i - \tau_j) < 0 }. 
   \end{equation}
 %\]
 However, computing a Kemeny consensus is a NP hard problem \cite{bartholdi1989computational, dwork2001rank}. It turns out that the Borda count, defined as the sum of ranks induced by the permutations, is a very good approximation of the Kemeny consensus \cite{ali2012experiments}, justifying our choices \eqref{eq:sys-level} and \eqref{eq:ut-level}.

\section{Extending Finding 1 using clustering analysis}\label{ssec:extending_finding_one}
In this section, we want to obtain a visual and interpretable representation of both automatic and human metrics to understand their relationships better. Formally, we study the abstract space of metrics when encoded at the System or Utterance level. We ask the two following questions:
\begin{itemize}
    \item What is the {\it effective dimension} of this space? 
    \item Does it exist clusters of metrics?
\end{itemize}

\subsection{Representing the metrics in a 2D space}
\begin{figure*}
             \begin{subfigure}[b]{0.5\textwidth}
         \centering
         \includegraphics[width=\textwidth]{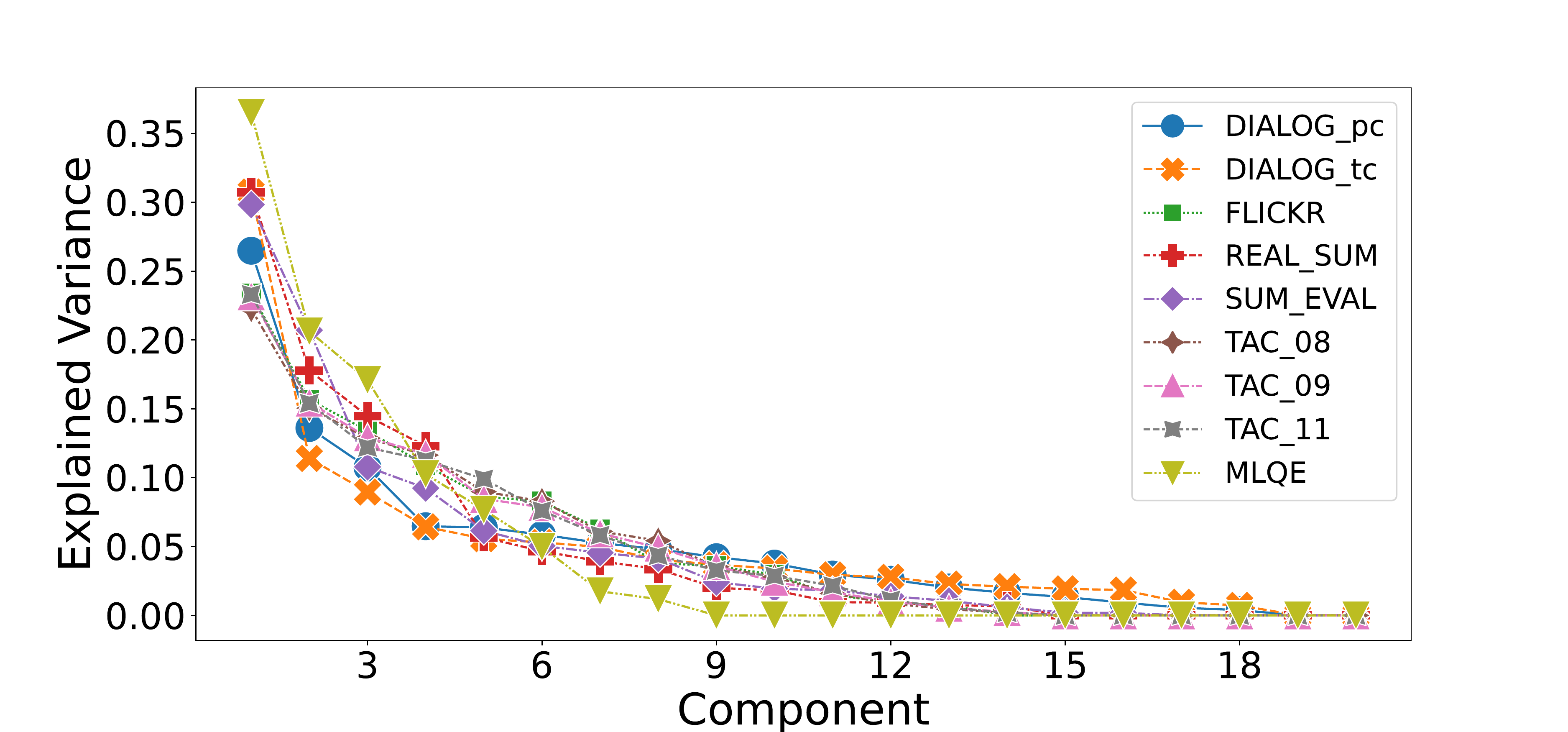}
         \caption{PCA Sys level}
         \label{fig:pca_system}
     \end{subfigure}
      \begin{subfigure}[b]{0.5\textwidth}
         \centering
         \includegraphics[width=\textwidth]{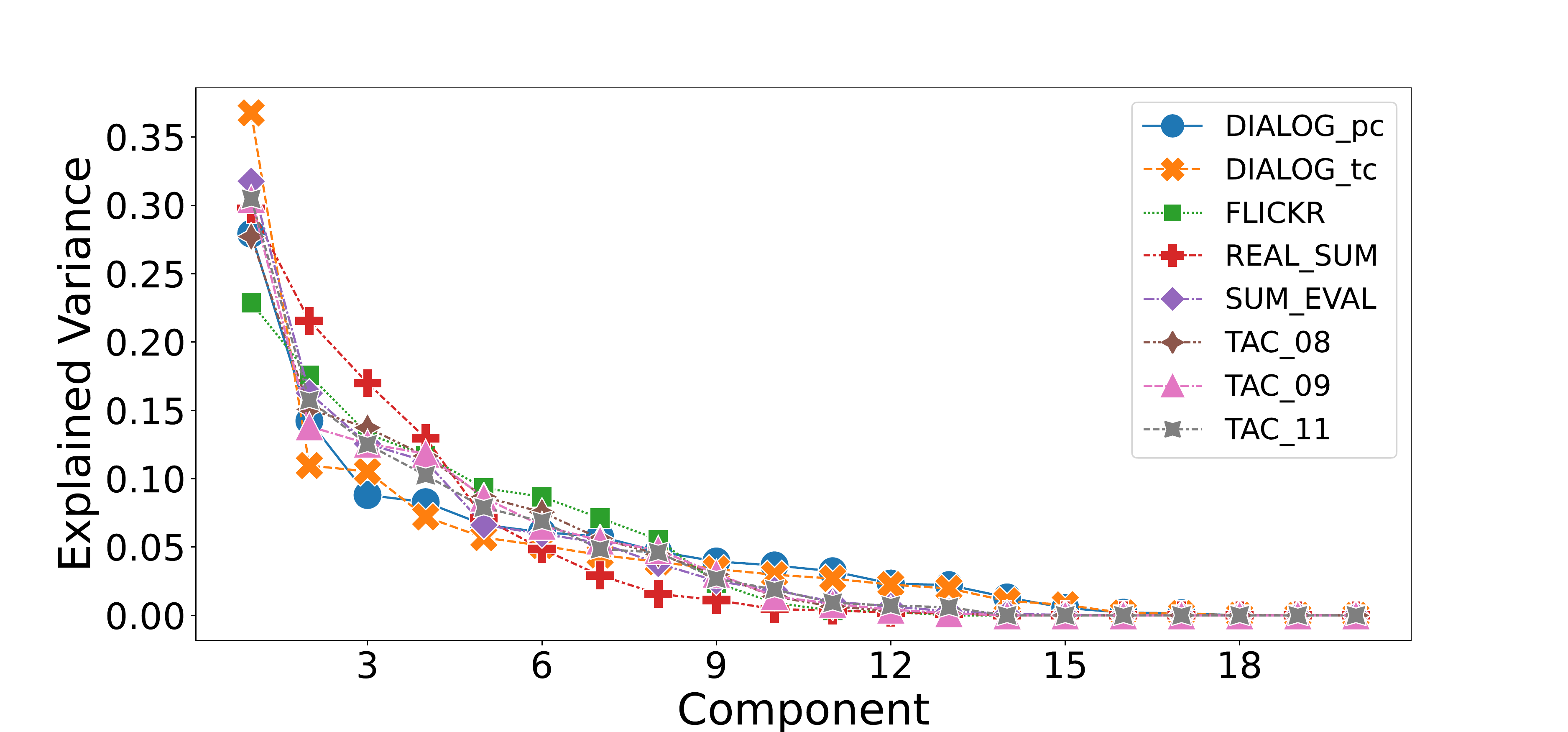}
         \caption{PCA utt level}
         \label{fig:pca_utter}
     \end{subfigure}
     \caption{Few components are need to explain the variance} 
\end{figure*}

In \autoref{fig:pca_system} and \autoref{fig:pca_utter}, we report the variance analysis given by a PCA \cite{jolliffe2016principal} for each dataset at the System and Utterance levels, respectively.
\\\texttt{Analysis:} We observe that only a few components (less than 6) are needed to explain over 80 \% of the variance. This behavior is typical to all considered datasets and can be observed when studying the ranks at the System and Instance levels.
\\\texttt{Takeaways:} Automatic and human metrics present in our datasets can be represented in a low-dimensional space. This confirms the low complementarity already observed in the main paper: the effective dimension of metrics is small. We will use the two first components in the next experiments to represent the metrics in a 2D space.

\subsection{Finding similar groups of metrics}
\begin{figure*}
     \centering
     \begin{subfigure}[b]{0.3\textwidth}
         \centering
         \includegraphics[width=\textwidth]{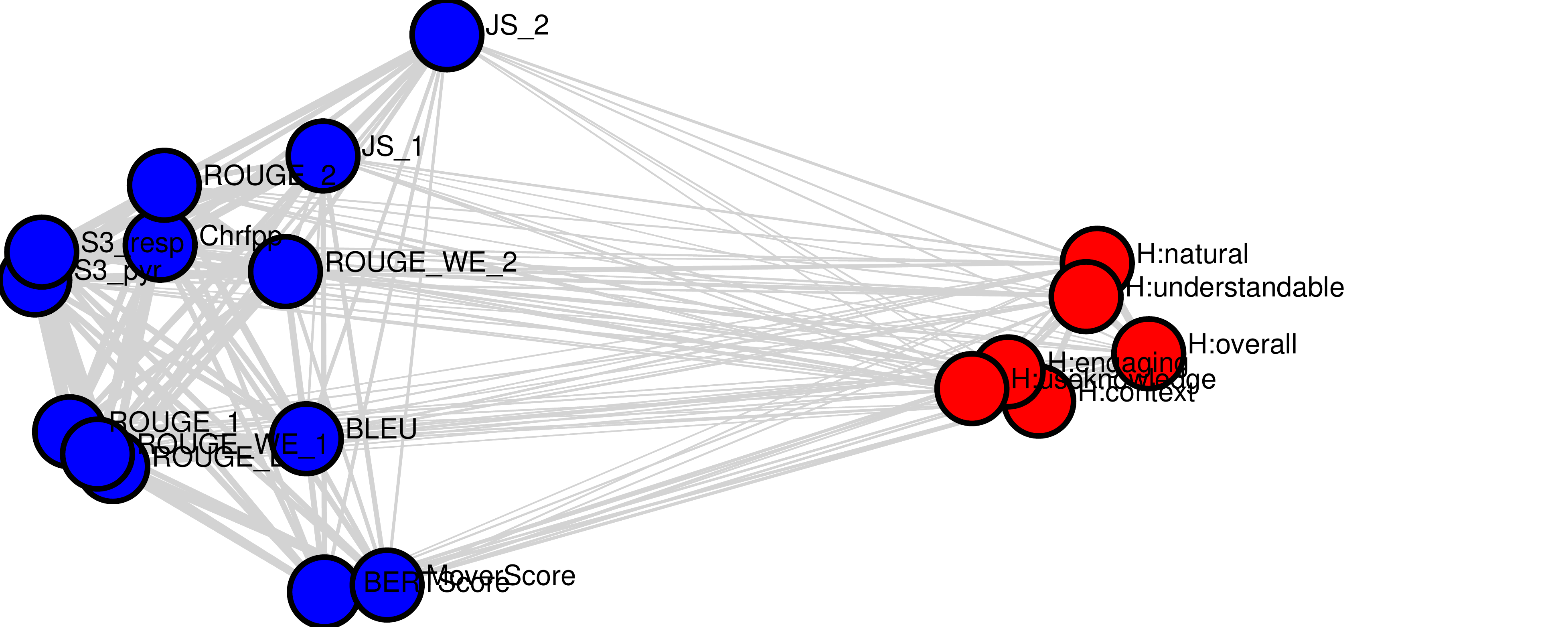}
         \caption{Dialog PC}
         \label{fig:y equals x}
     \end{subfigure}
     \hfill
     \begin{subfigure}[b]{0.3\textwidth}
         \centering
         \includegraphics[width=\textwidth]{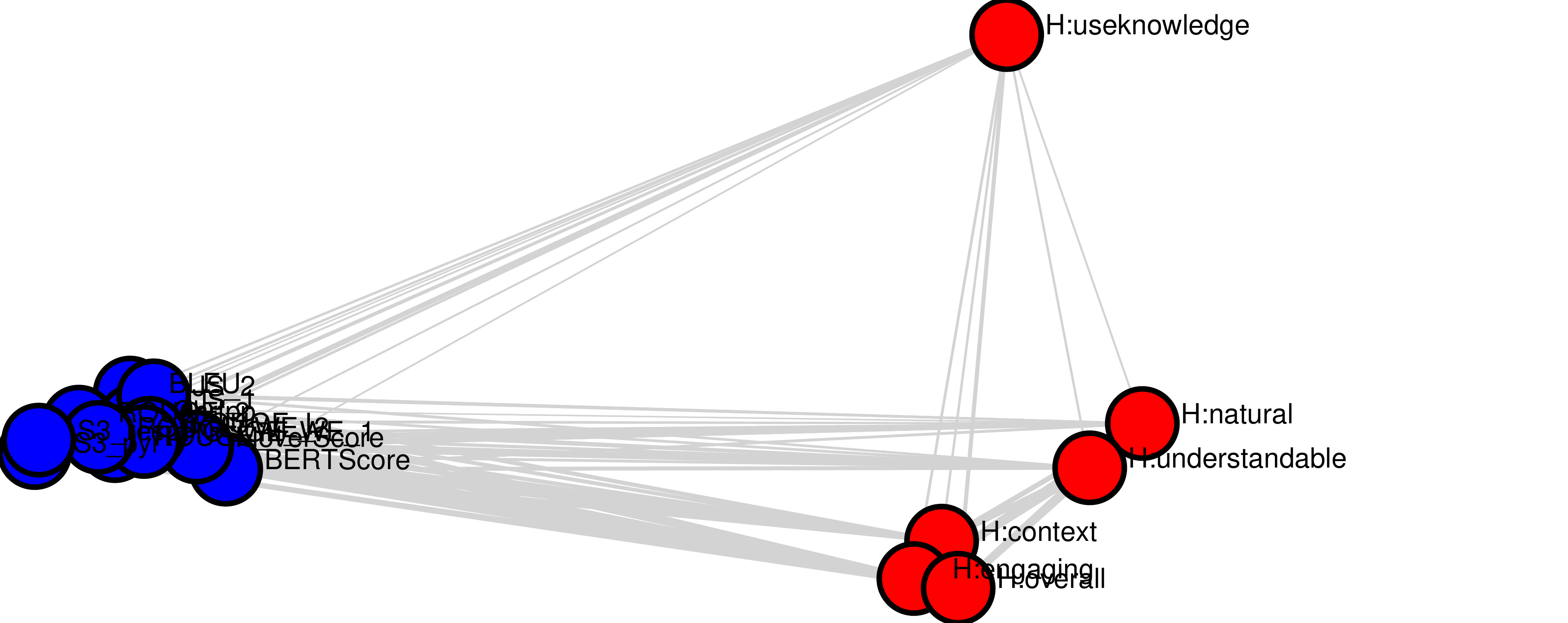}
         \caption{Dialog TC}
         \label{fig:three sin x}
     \end{subfigure}
     \hfill
     \begin{subfigure}[b]{0.3\textwidth}
         \centering
         \includegraphics[width=\textwidth]{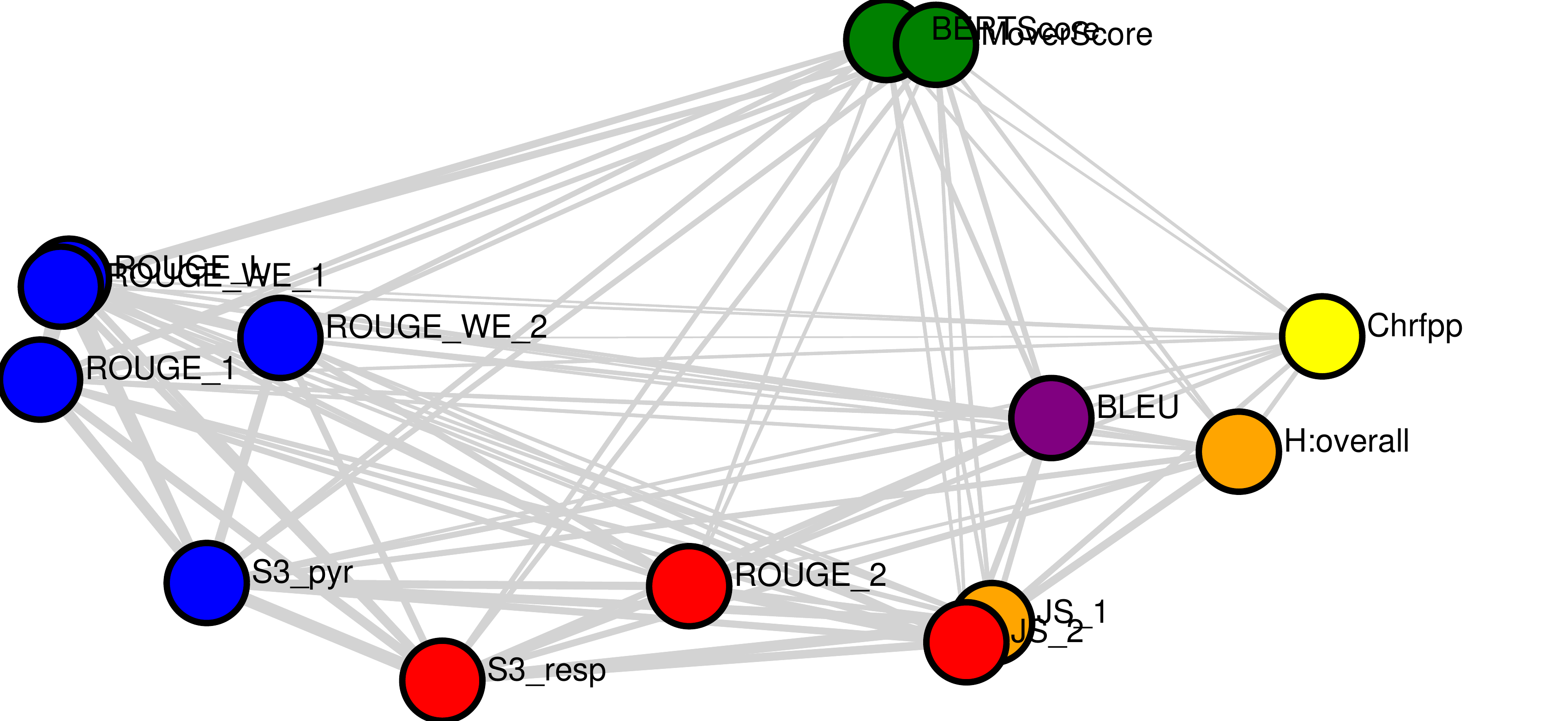}
         \caption{FLICKR}
         \label{fig:five over x}
     \end{subfigure}
 \\
             \begin{subfigure}[b]{0.3\textwidth}
         \centering
         \includegraphics[width=\textwidth]{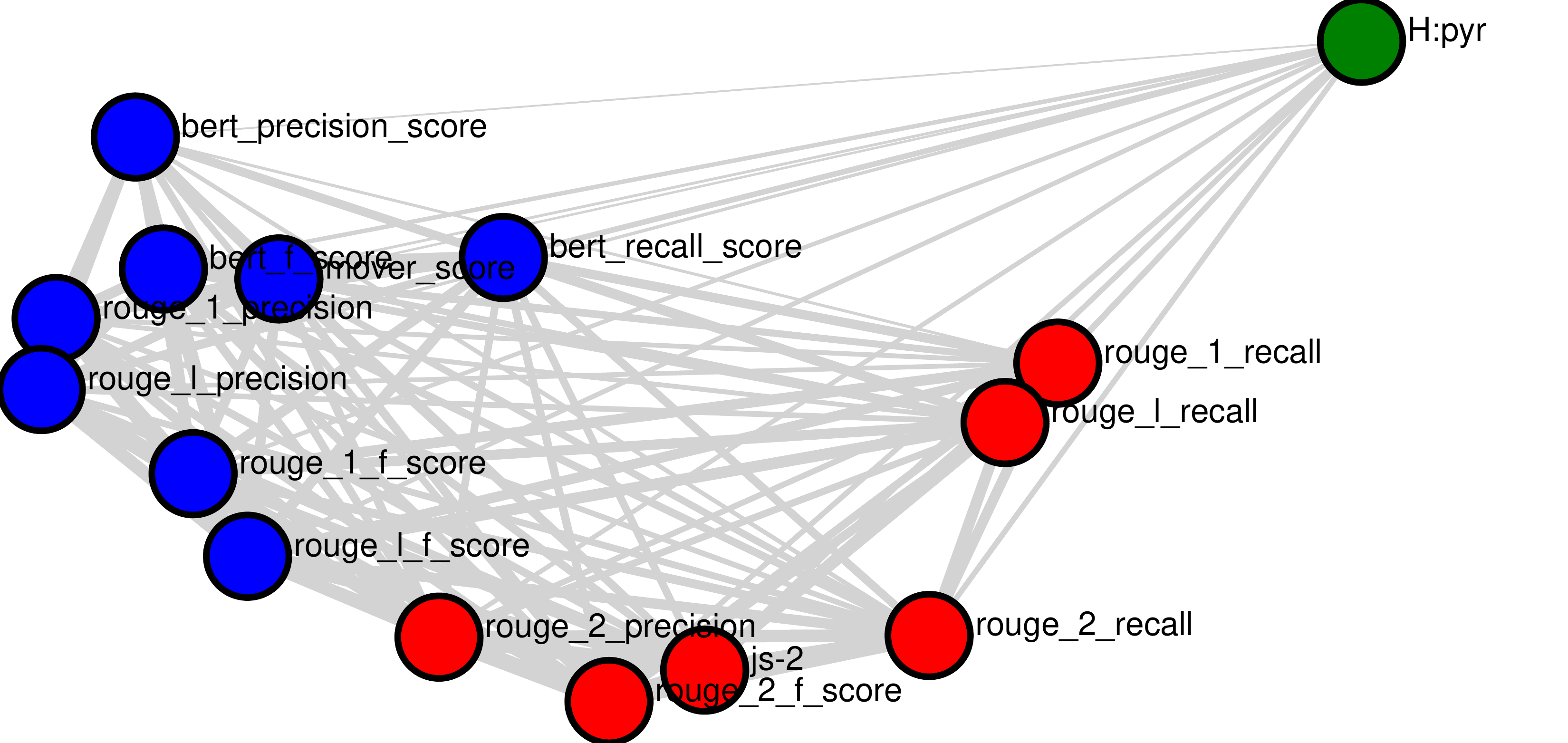}
         \caption{REAL SUM}
         \label{fig:y equals x}
     \end{subfigure}
     \hfill
     \begin{subfigure}[b]{0.3\textwidth}
         \centering
         \includegraphics[width=\textwidth]{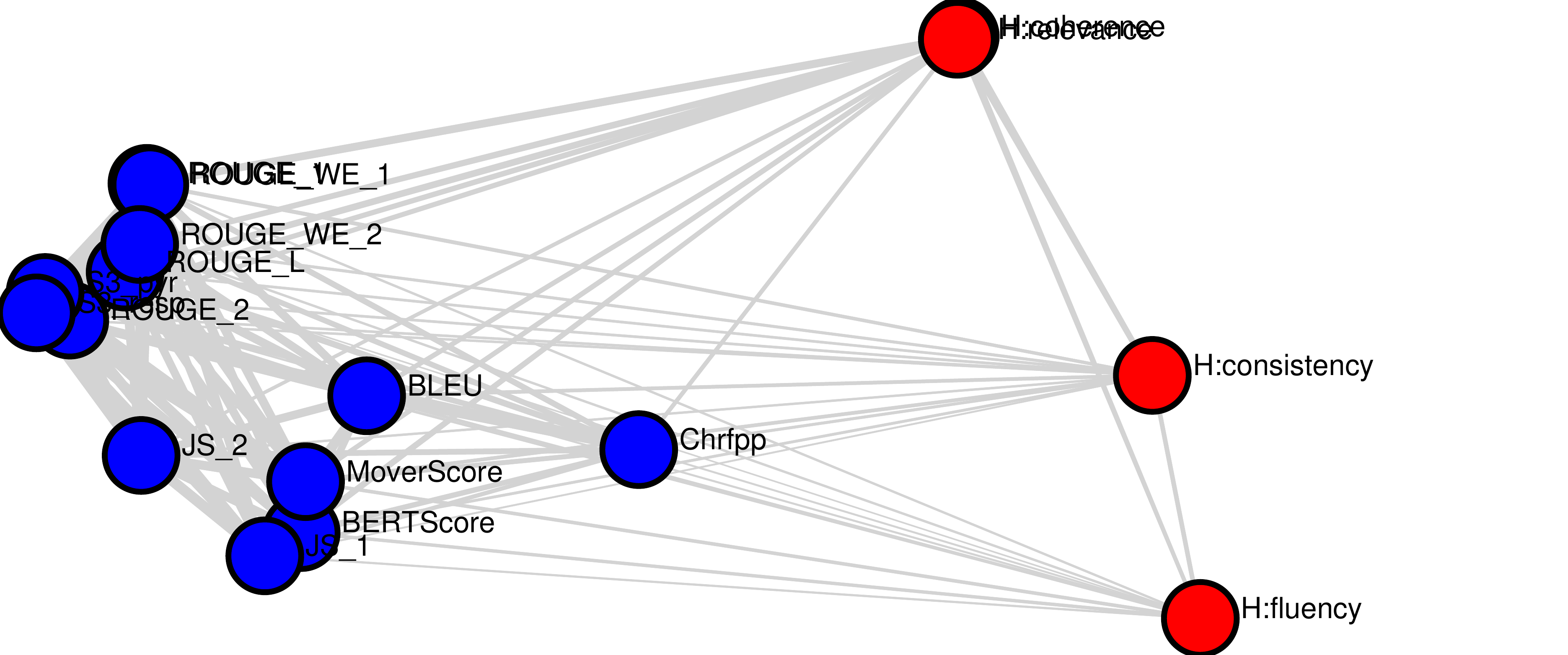}
         \caption{SUM EVAL}
         \label{fig:three sin x}
     \end{subfigure}
     \hfill
     \begin{subfigure}[b]{0.3\textwidth}
         \centering
         \includegraphics[width=\textwidth]{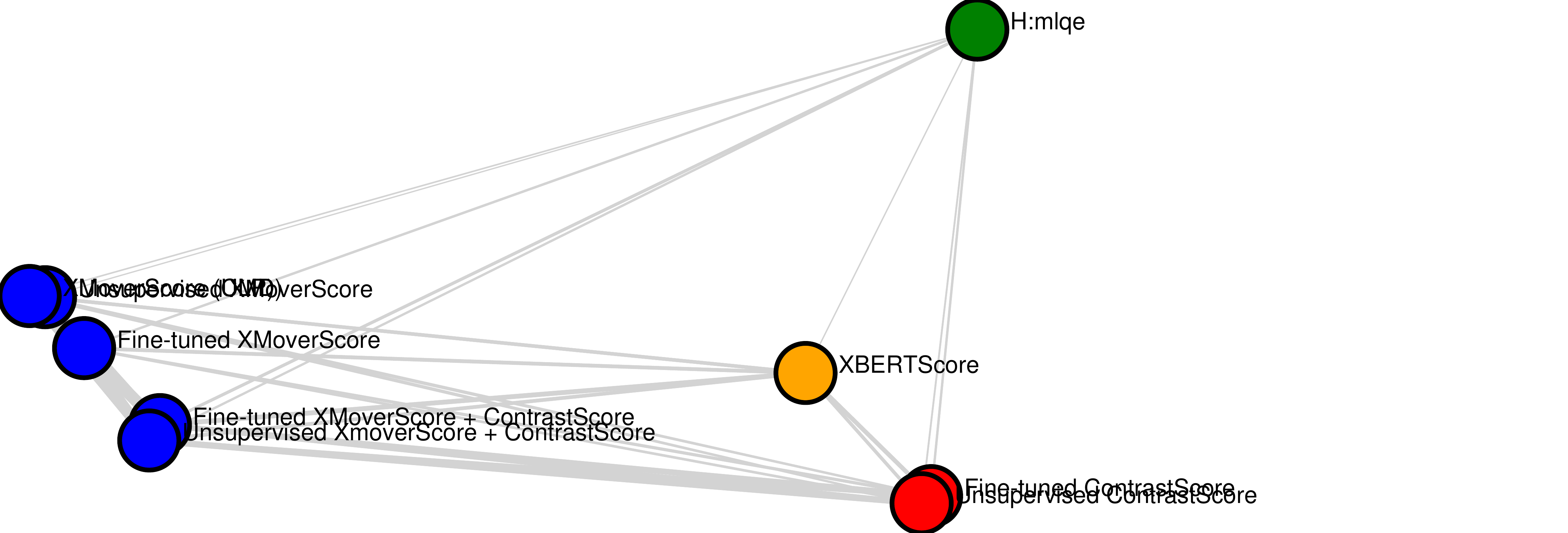}
         \caption{MLQE}
         \label{fig:five over x}
     \end{subfigure}
 \\
             \begin{subfigure}[b]{0.3\textwidth}
         \centering
         \includegraphics[width=\textwidth]{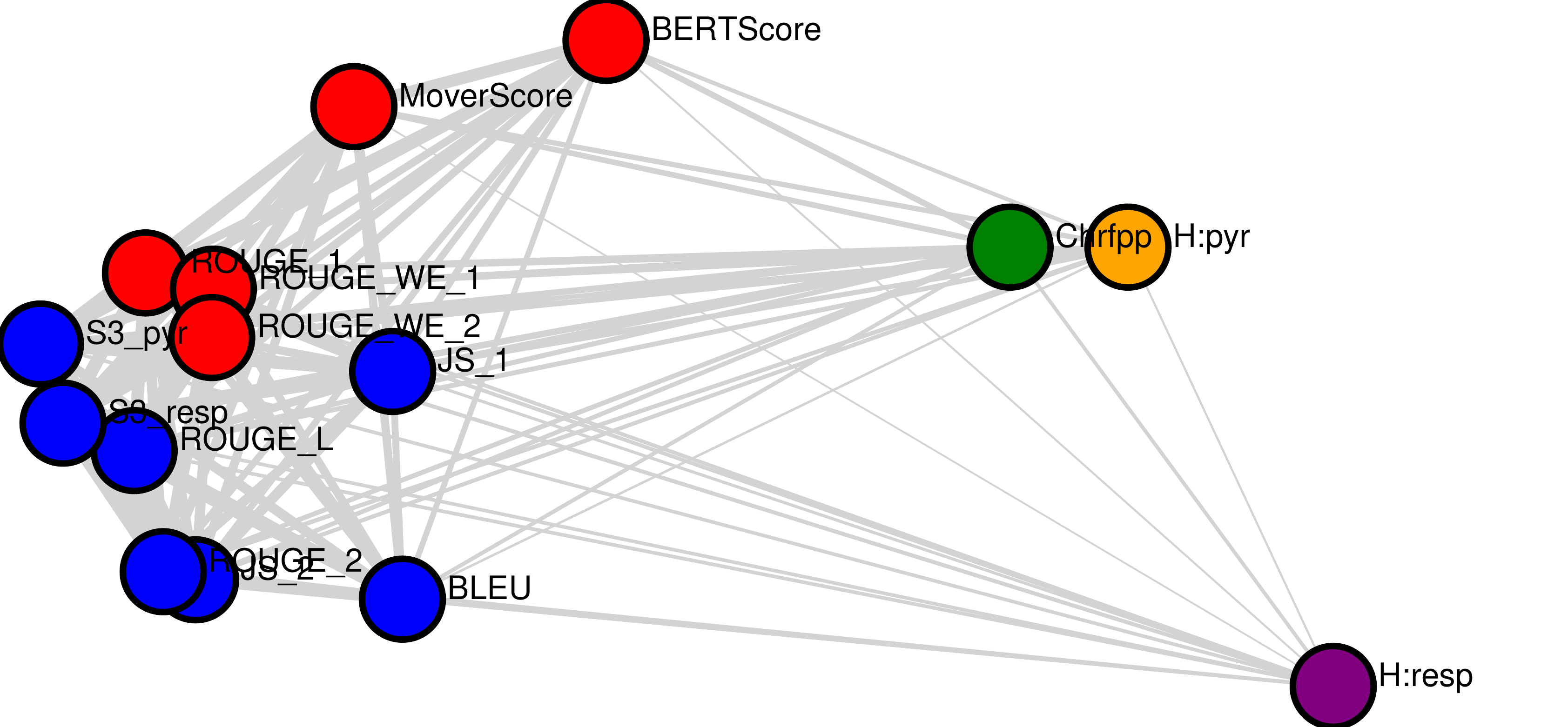}
         \caption{TAC8}
         \label{fig:y equals x}
     \end{subfigure}
     \hfill
     \begin{subfigure}[b]{0.3\textwidth}
         \centering
         \includegraphics[width=\textwidth]{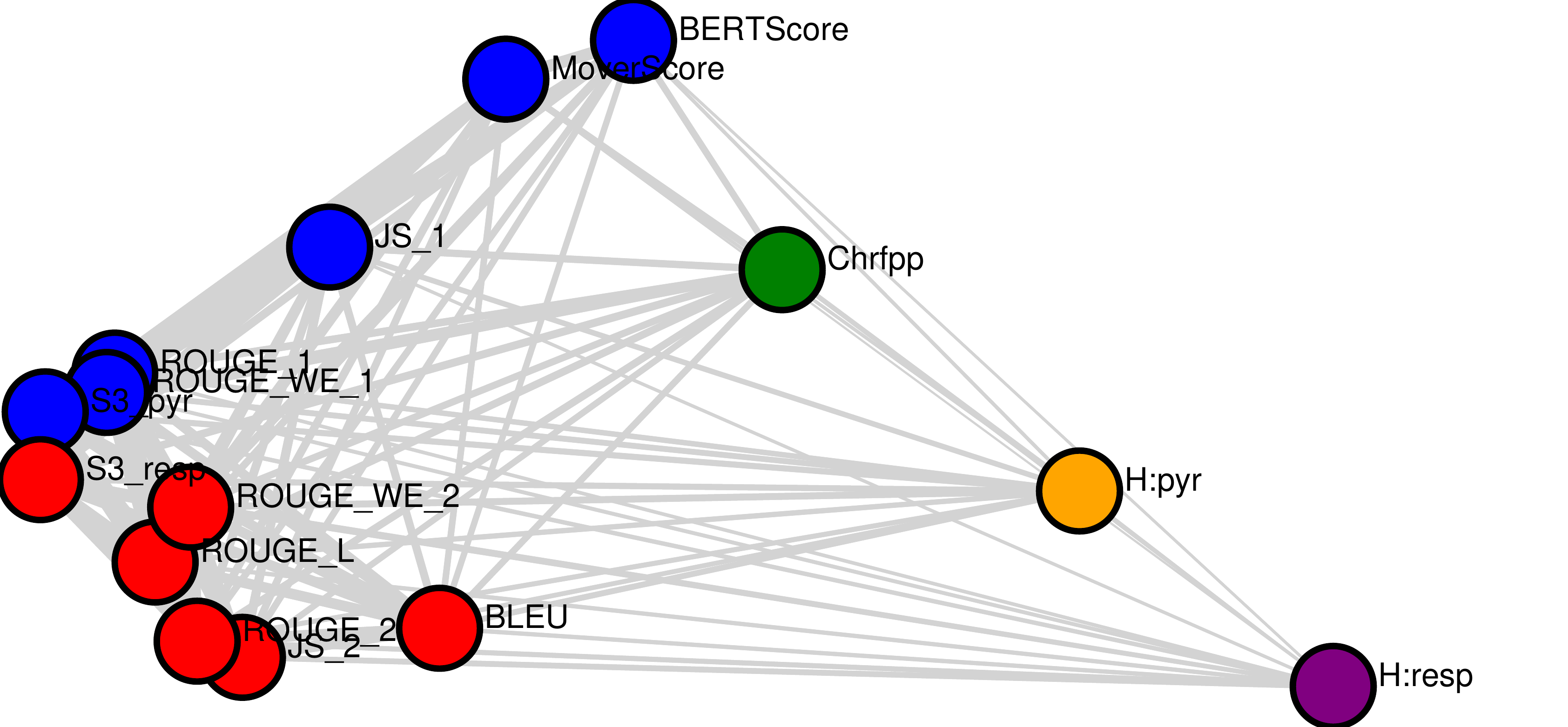}
         \caption{TAC9}
         \label{fig:three sin x}
     \end{subfigure}
     \hfill
     \begin{subfigure}[b]{0.3\textwidth}
         \centering
         \includegraphics[width=\textwidth]{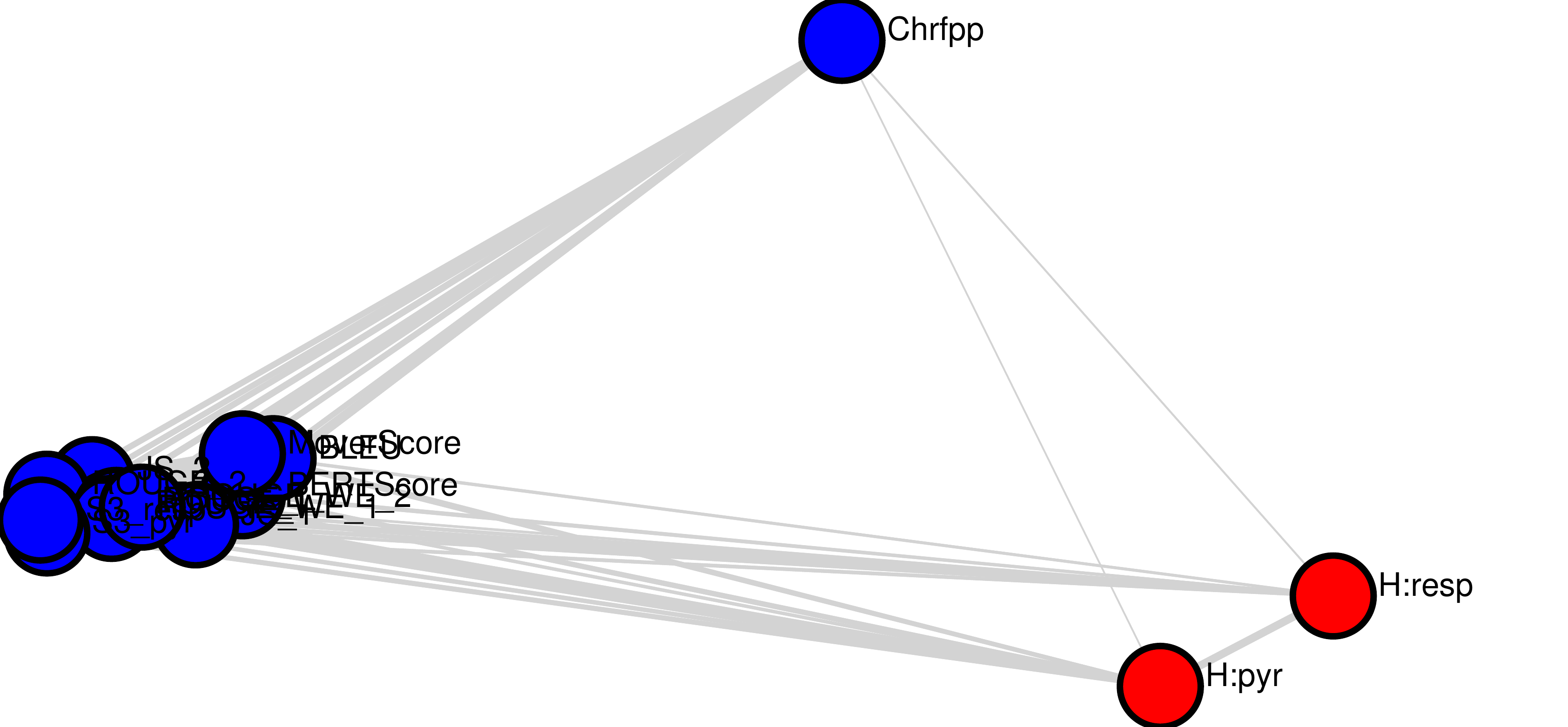}
         \caption{TAC11}
         \label{fig:five over x}
     \end{subfigure}
     \caption{\textbf{Metric visualization at the System Level in a 2D space with clustering analysis.} For each dataset, the metric representations are obtained by considering the two first components of the PCA. To get the cluster of similar metrics, the Louvain Algorithm is applied.}\label{fig:pca_louvain_sys}
\end{figure*}

\begin{figure*}
     \begin{subfigure}[b]{0.3\textwidth}
         \centering
         \includegraphics[width=\textwidth]{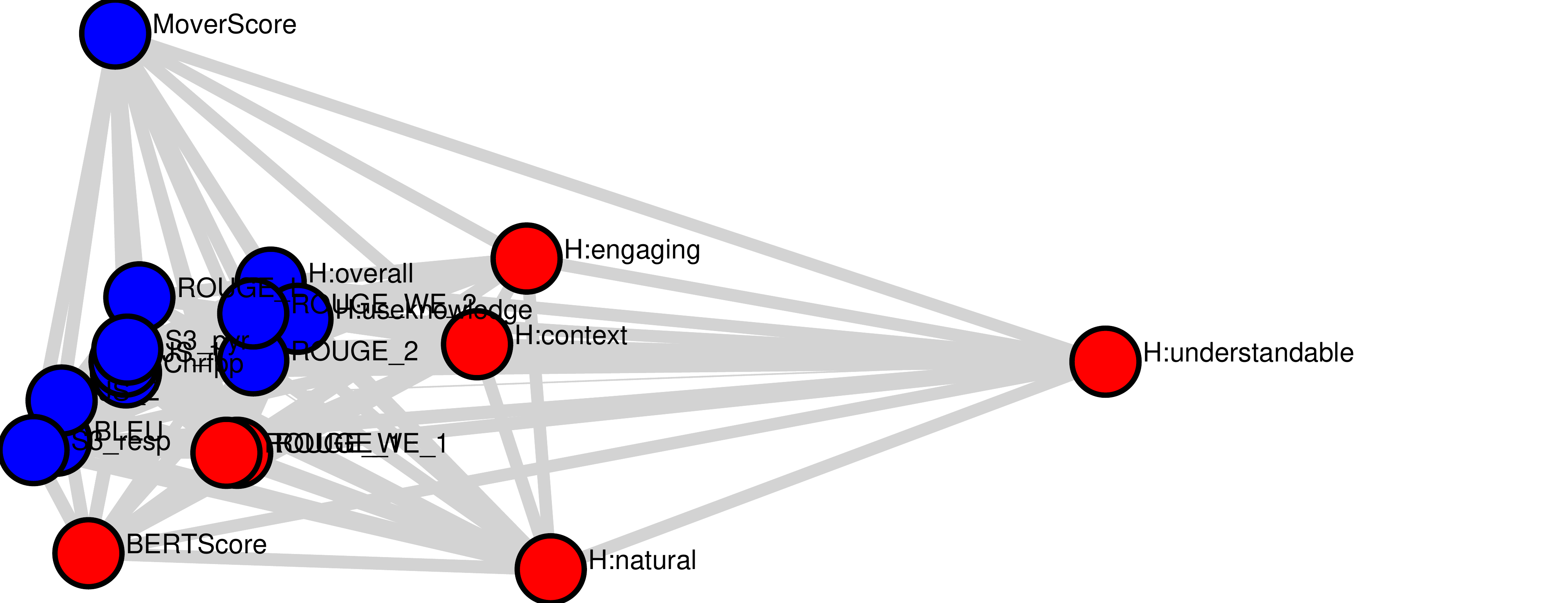}
         \caption{DIALOG TC}
         \label{fig:three sin x}
     \end{subfigure}
     \hfill
     \begin{subfigure}[b]{0.3\textwidth}
         \centering
         \includegraphics[width=\textwidth]{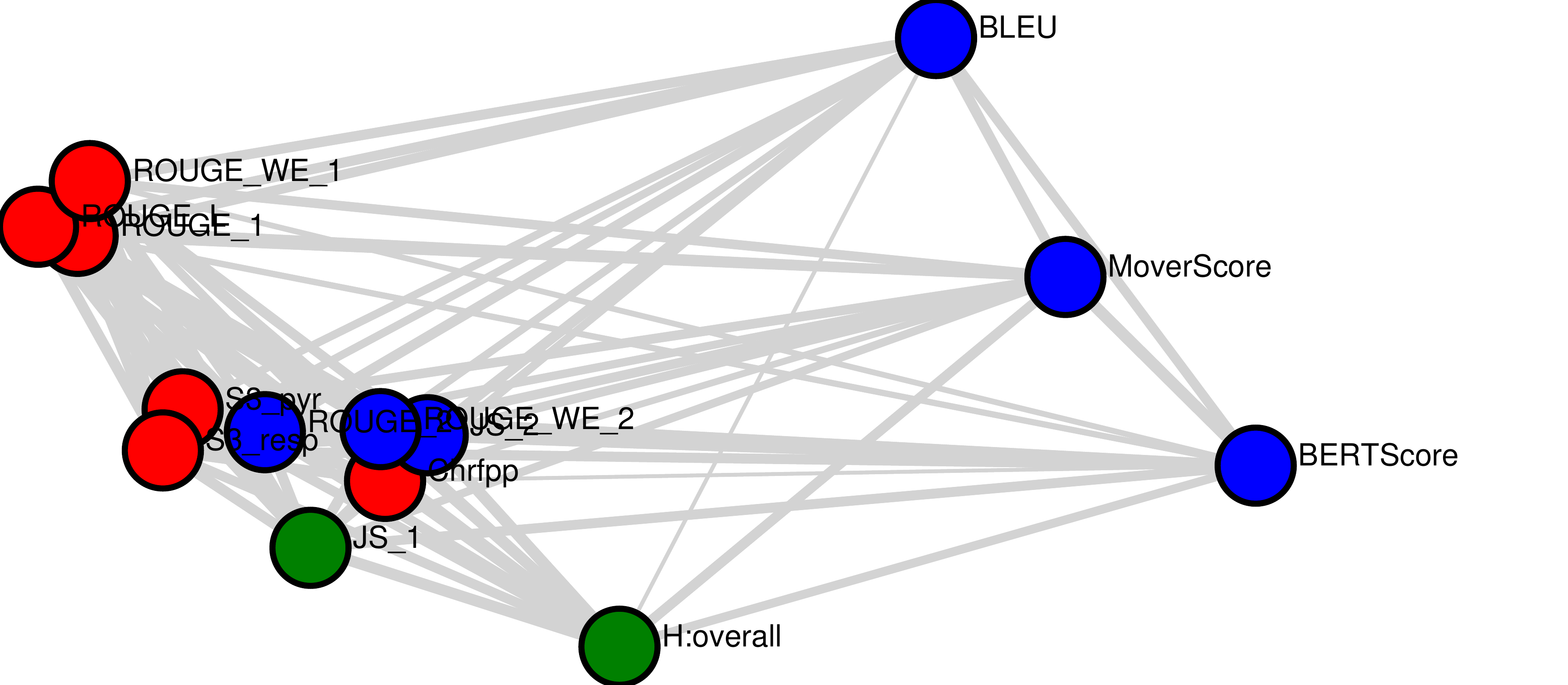}
         \caption{FLICKR}
         \label{fig:five over x}
     \end{subfigure}
 \\
             \begin{subfigure}[b]{0.3\textwidth}
         \centering
         \includegraphics[width=\textwidth]{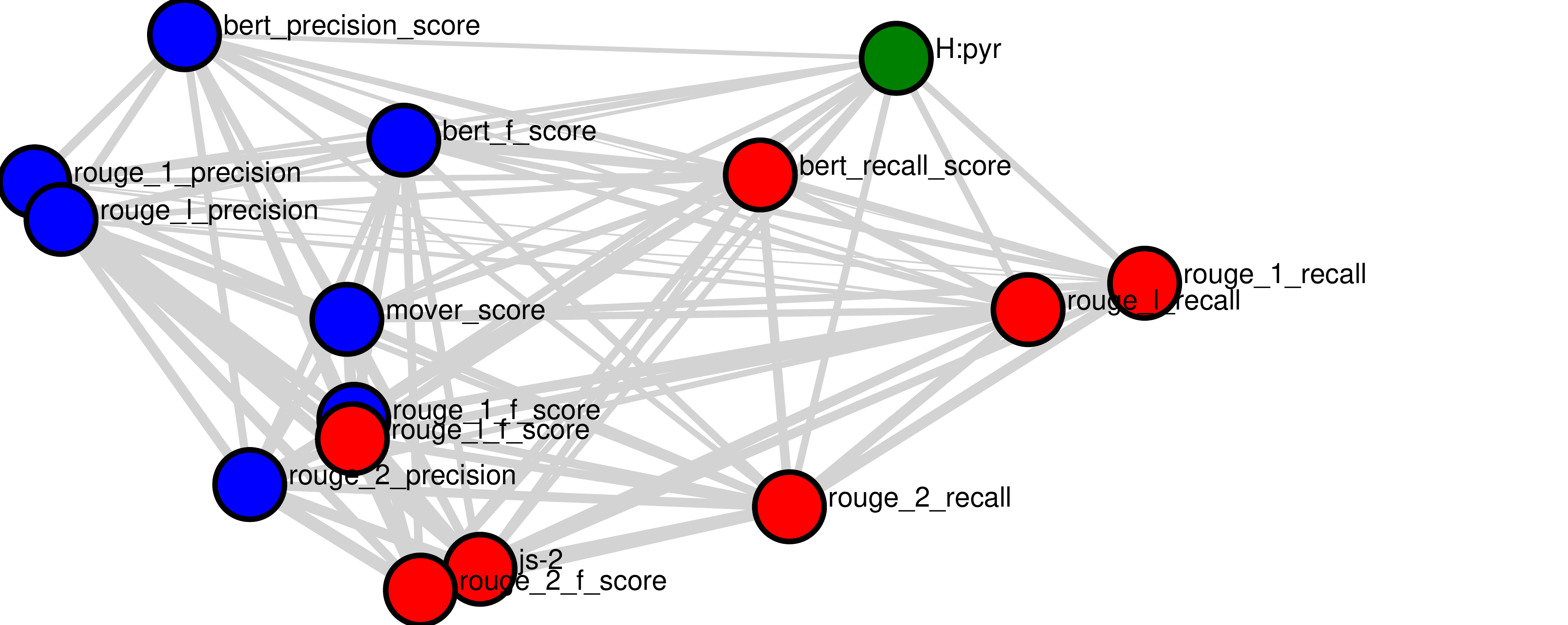}
         \caption{REAL SUM}
         \label{fig:y equals x}
     \end{subfigure}
     \hfill
     \begin{subfigure}[b]{0.3\textwidth}
         \centering
         \includegraphics[width=\textwidth]{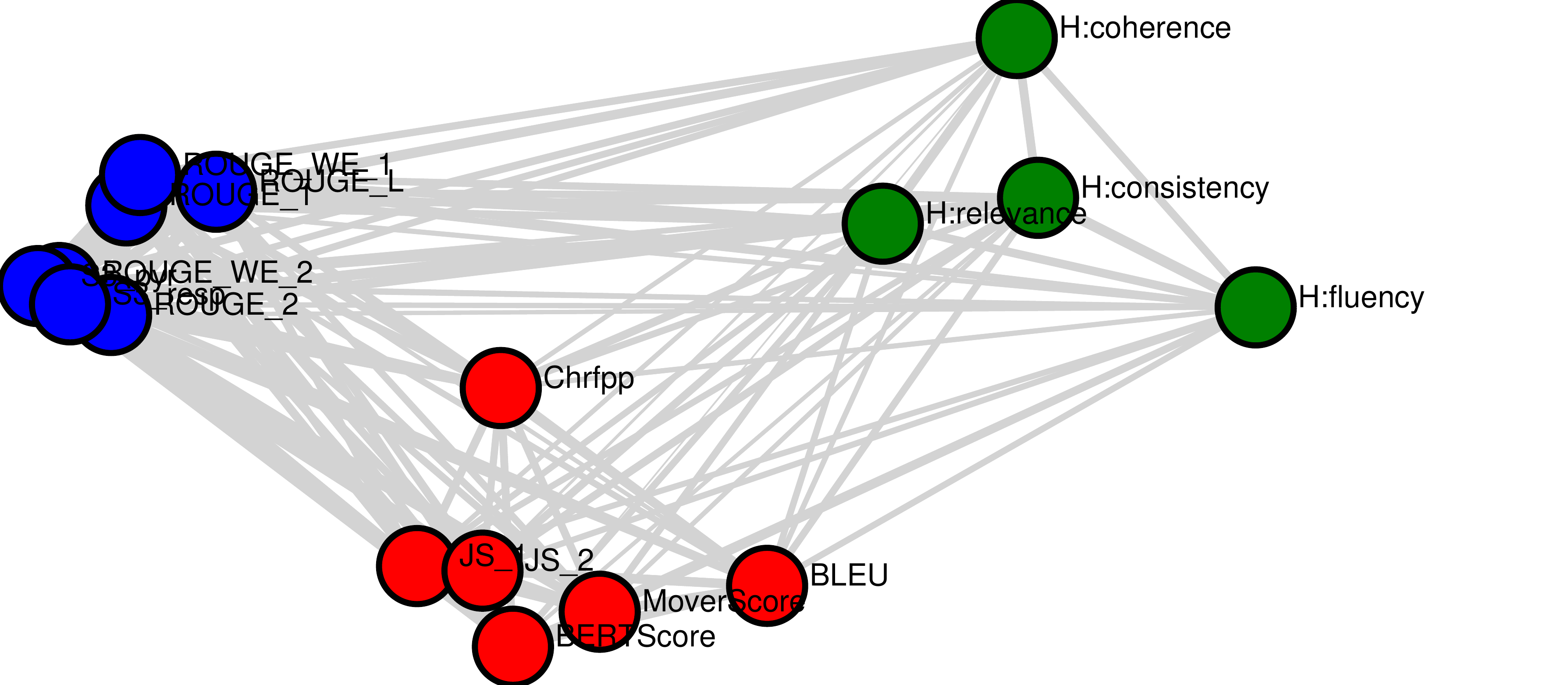}
         \caption{SUM EVAL}
         \label{fig:three sin x}
     \end{subfigure}
     \hfill
     \begin{subfigure}[b]{0.3\textwidth}
         \centering
         \includegraphics[width=\textwidth]{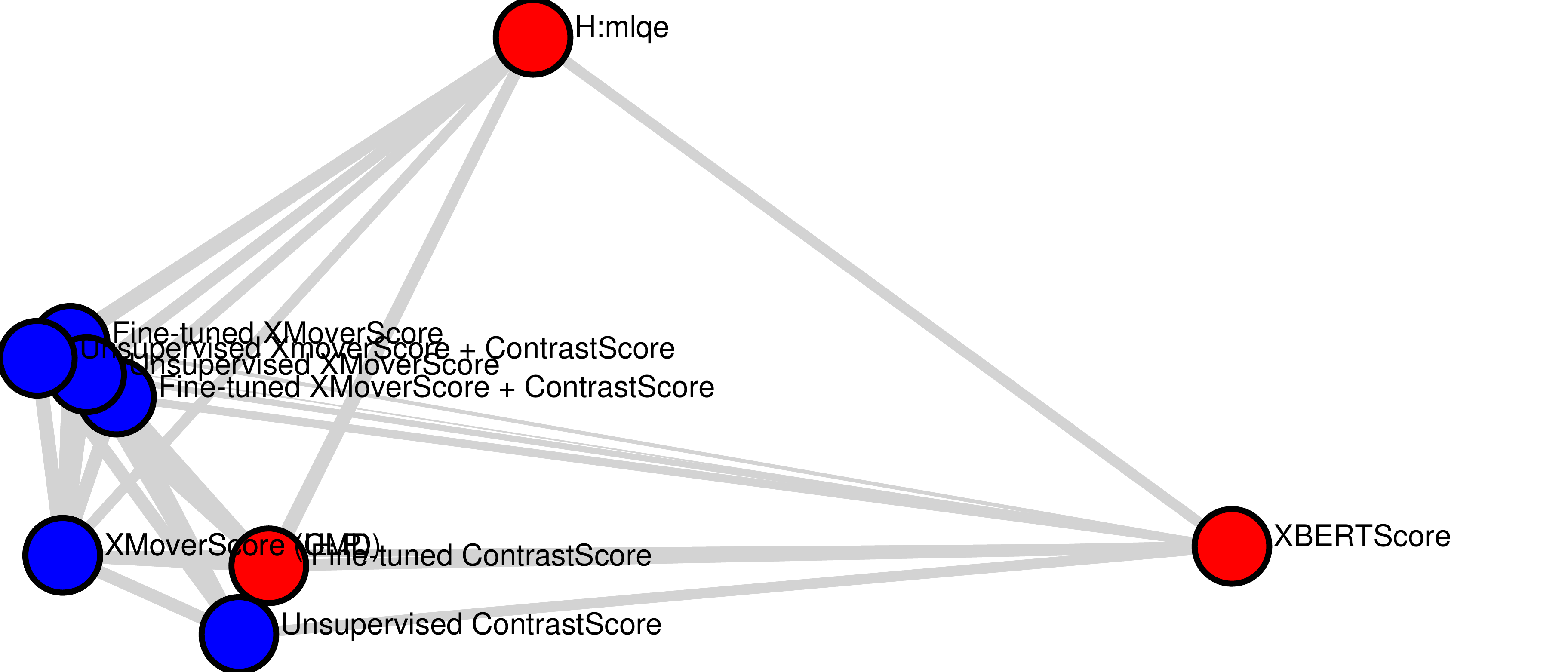}
         \caption{MLQE}
         \label{fig:five over x}
     \end{subfigure}
 \\
             \begin{subfigure}[b]{0.3\textwidth}
         \centering
         \includegraphics[width=\textwidth]{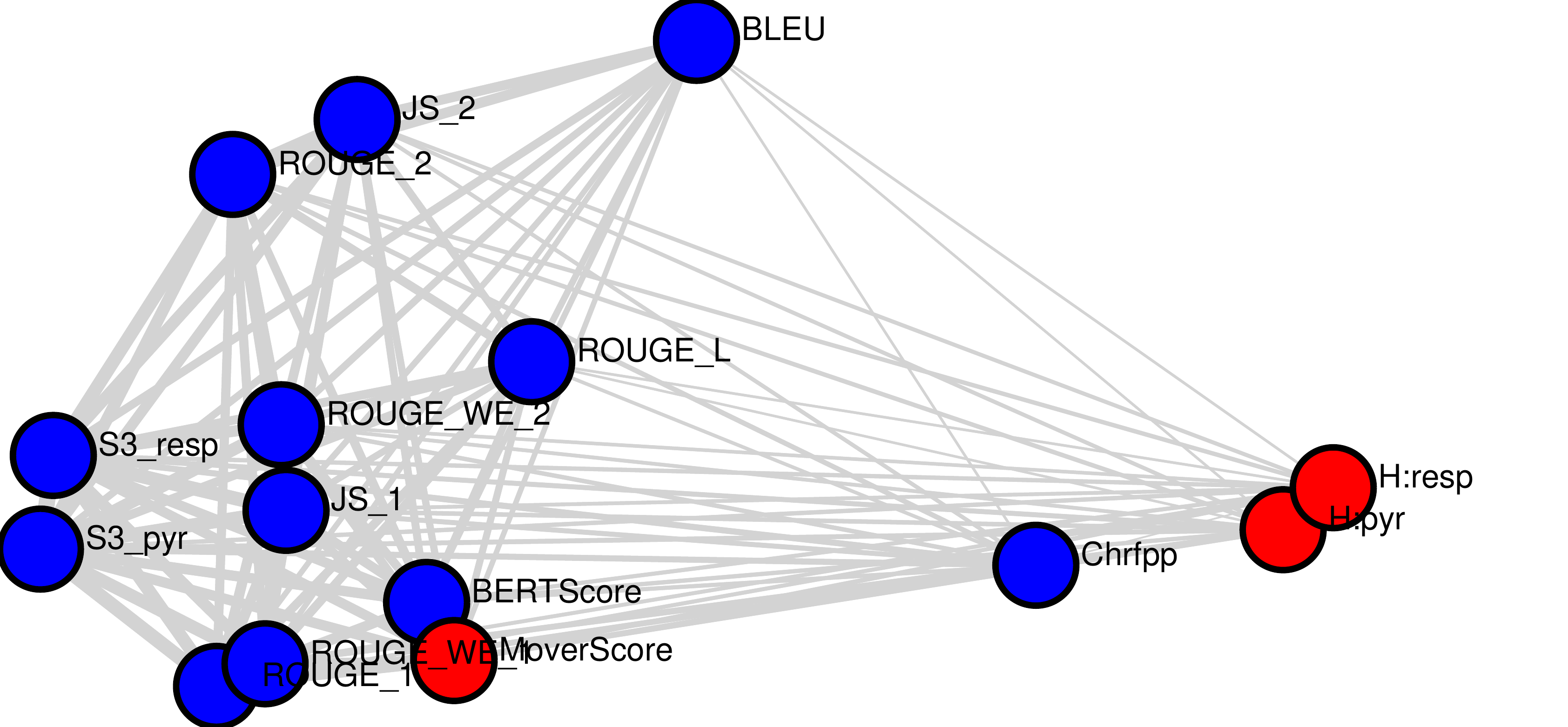}
         \caption{TAC8}
         \label{fig:y equals x}
     \end{subfigure}
     \hfill
     \begin{subfigure}[b]{0.3\textwidth}
         \centering
         \includegraphics[width=\textwidth]{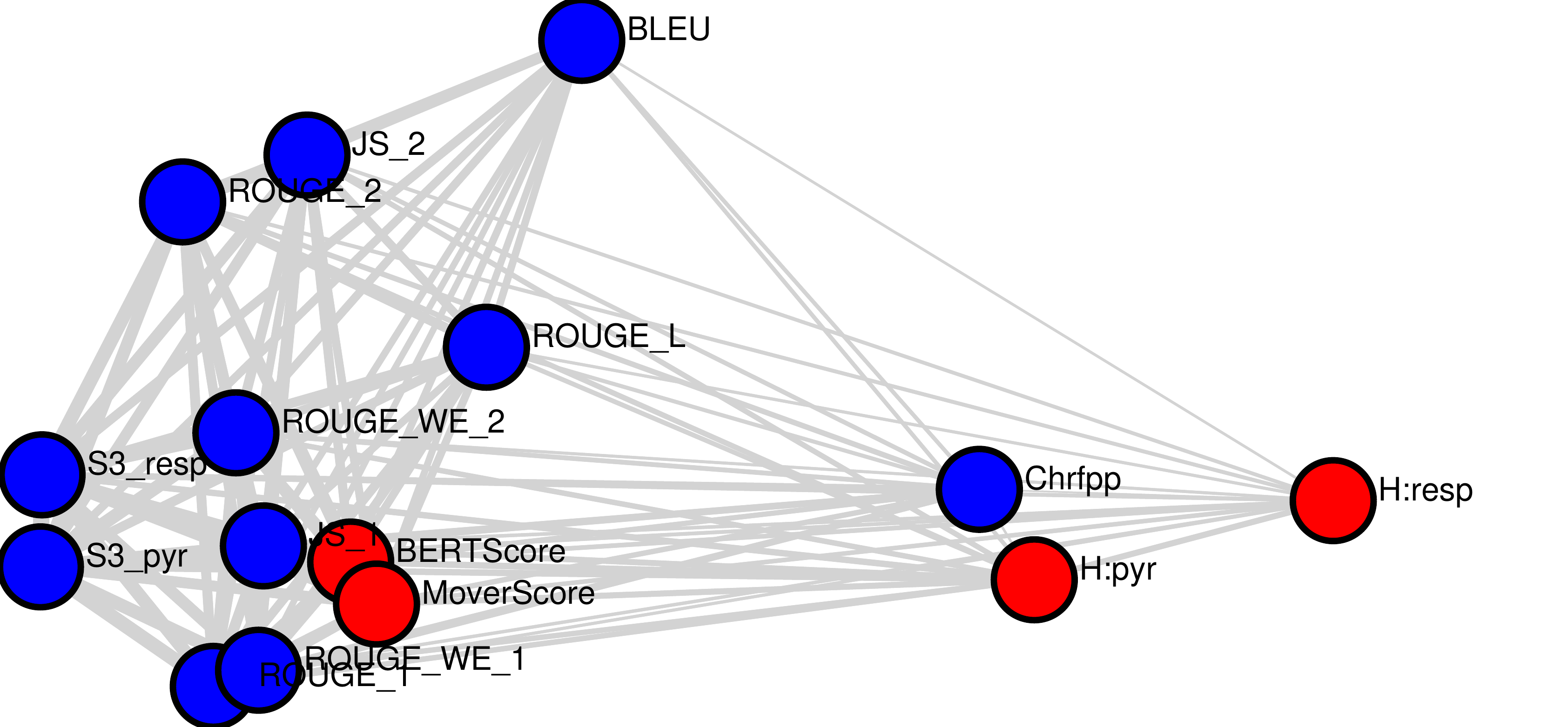}
         \caption{TAC9}
         \label{fig:three sin x}
     \end{subfigure}
     \hfill
     \begin{subfigure}[b]{0.3\textwidth}
         \centering
         \includegraphics[width=\textwidth]{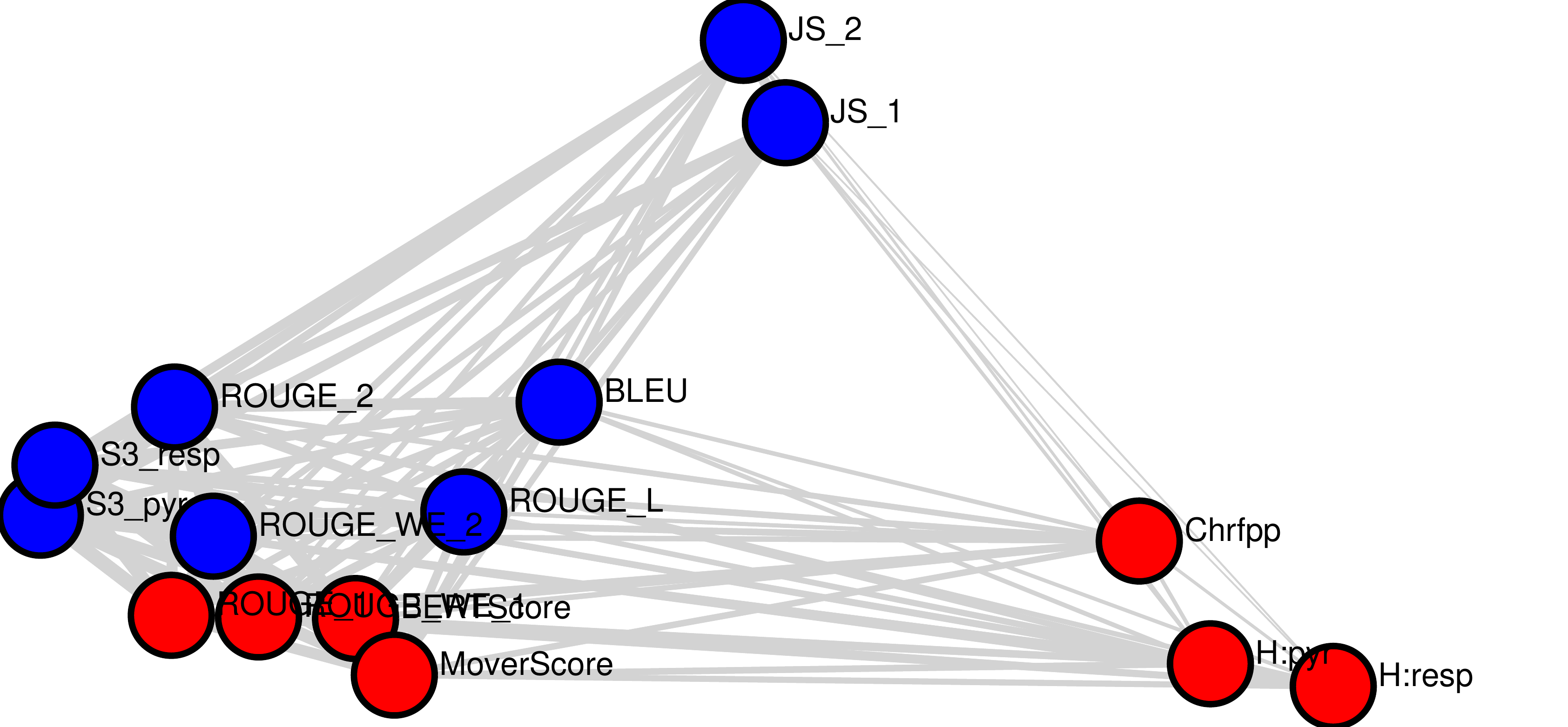}
         \caption{TAC11}
         \label{fig:five over x}
     \end{subfigure}
   \caption{\textbf{Metric visualization at the utterance level in a 2D space with clustering analysis.} For each dataset, the metric representations are obtained by considering the two first components of the PCA. To get the cluster of similar metrics, the Louvain Algorithm is applied.}  \label{fig:pca_louvain_text}
\end{figure*}

In \autoref{fig:pca_louvain_sys} and \autoref{fig:pca_louvain_text}, we represent all the considered metrics (both human and automatic) on the 2D-dimensional space corresponding to the two first components of the PCA. We cluster the metrics with the Louvain Algorithm \cite{blondel2008fast} performed on the similarity matrix between metrics.

\noindent \texttt{Analysis:} From the figure, we observe a low number of clusters, i.e., two in most cases and at most three in the case of utterance level representations. When using system-level representation, the Human metrics have their cluster in all the configurations except for FLICKR, where H:overall is in the same cluster as $JS_2$. We observe a similar trend when studying the utterance level representation: human metrics often belong to the same cluster, which contains a low number of automatic metrics. It is also worth noting that in most figures, human metrics are isolated.
%i.e., they tend to have at least one higher coordinate than the automatic metrics. 

\noindent \texttt{Takeaways:} This experiment further validates Findings 1: Automatic metrics are similar to each other much more than they are to human metric. The proposed procedure could be used in the future to find properties of newly introduced metrics and obtain visual representations of the metrics.

\subsection{Extension to other types of tasks}
In the futur we would like to incoporate more metrics such as \texttt{BaryScore} \cite{colombo2021automatic}, \texttt{InfoLM} \cite{colombo2021infolm}, \texttt{DepthScore} \cite{staerman2021depth} and apply our methodology to other tasks such as affect driven sentence generation \cite{colombo-etal-2019-affect,colombo2021beam,colombo2021learning,colombo2021novel,colombo2020guiding,colombo2021improving,witon-etal-2018-disney,colombo2022learning,chapuis2020hierarchical,chapuis2021code} or story generation \cite{chhun2022human}.

\section{Further results for Findings 2}
In this section, we provide further experiments that validate Findings 2 and provide a method for future research to understand newly introduced metrics better. Specifically, we aim to answer the following research question:
\begin{itemize}
    \item In Findings 1 we showcase that human metrics carry different information than automatic metrics. How to measure the amount of information missing between the automatic and human metrics?
    \item What metric or group of metrics are the most useful to predict a given human metric?
\end{itemize}

\subsection{Measuring the information missing in automatic metrics}

In this subsection, we extend the result provided by \autoref{fig:predicted_power}. We measure the ratio between the MSE-error of a linear regression trained with automatic metrics together with human metrics and a linear regression trained only with automatic metrics for varying regularization coefficient. For each dataset, we provide mean and variance corresponding to the prediction of available human metrics. When solely one human metric is available, the dataset is not considered. 

\begin{figure*}
     \centering
     \begin{subfigure}[b]{0.45\textwidth}
         \centering
         \includegraphics[width=\textwidth]{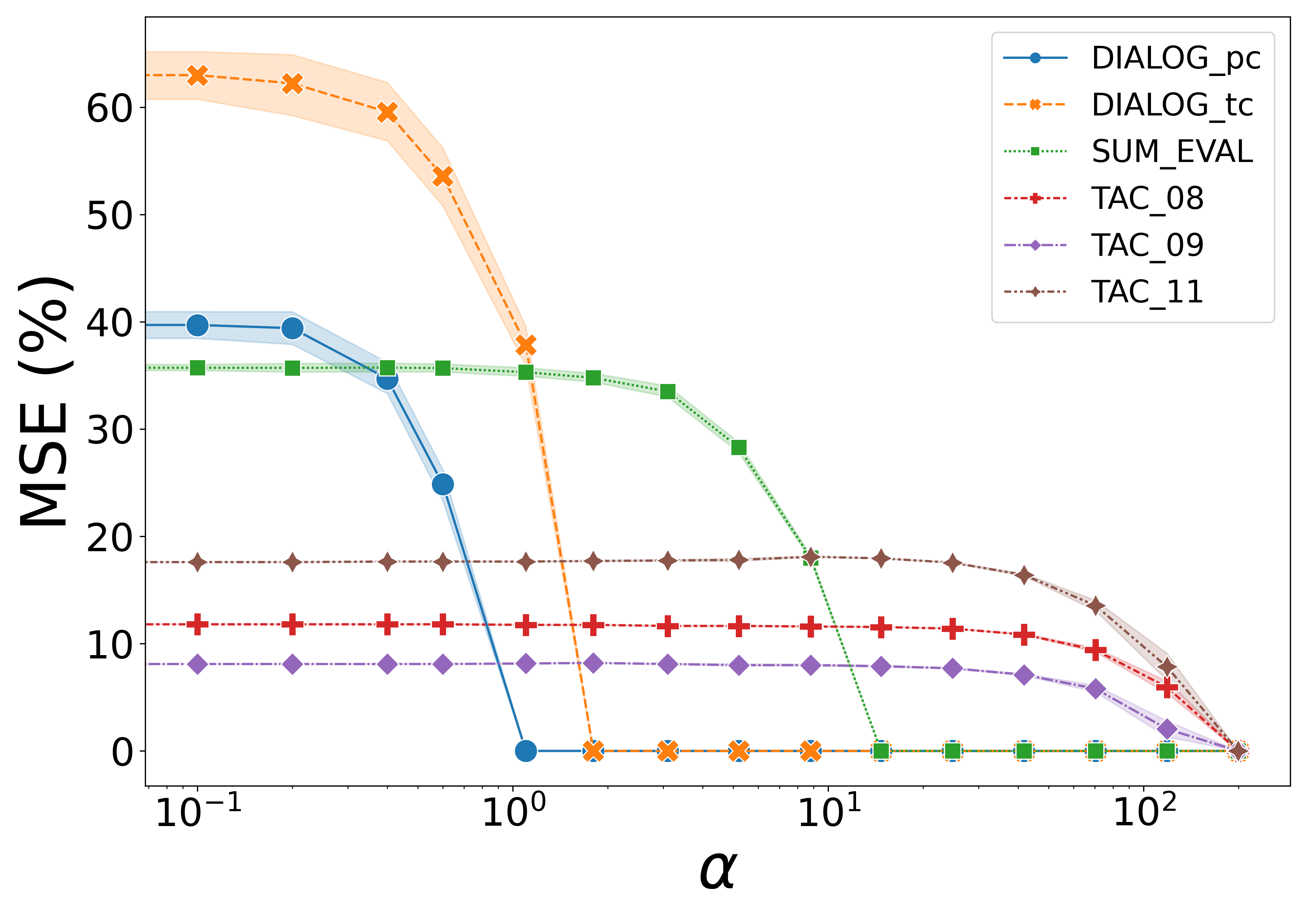}
         \caption{Aggregated score for each task when using System Level Representation.}
         \label{fig:y equals x}
     \end{subfigure}
     \hfill
     \begin{subfigure}[b]{0.45\textwidth}
         \centering
         \includegraphics[width=\textwidth]{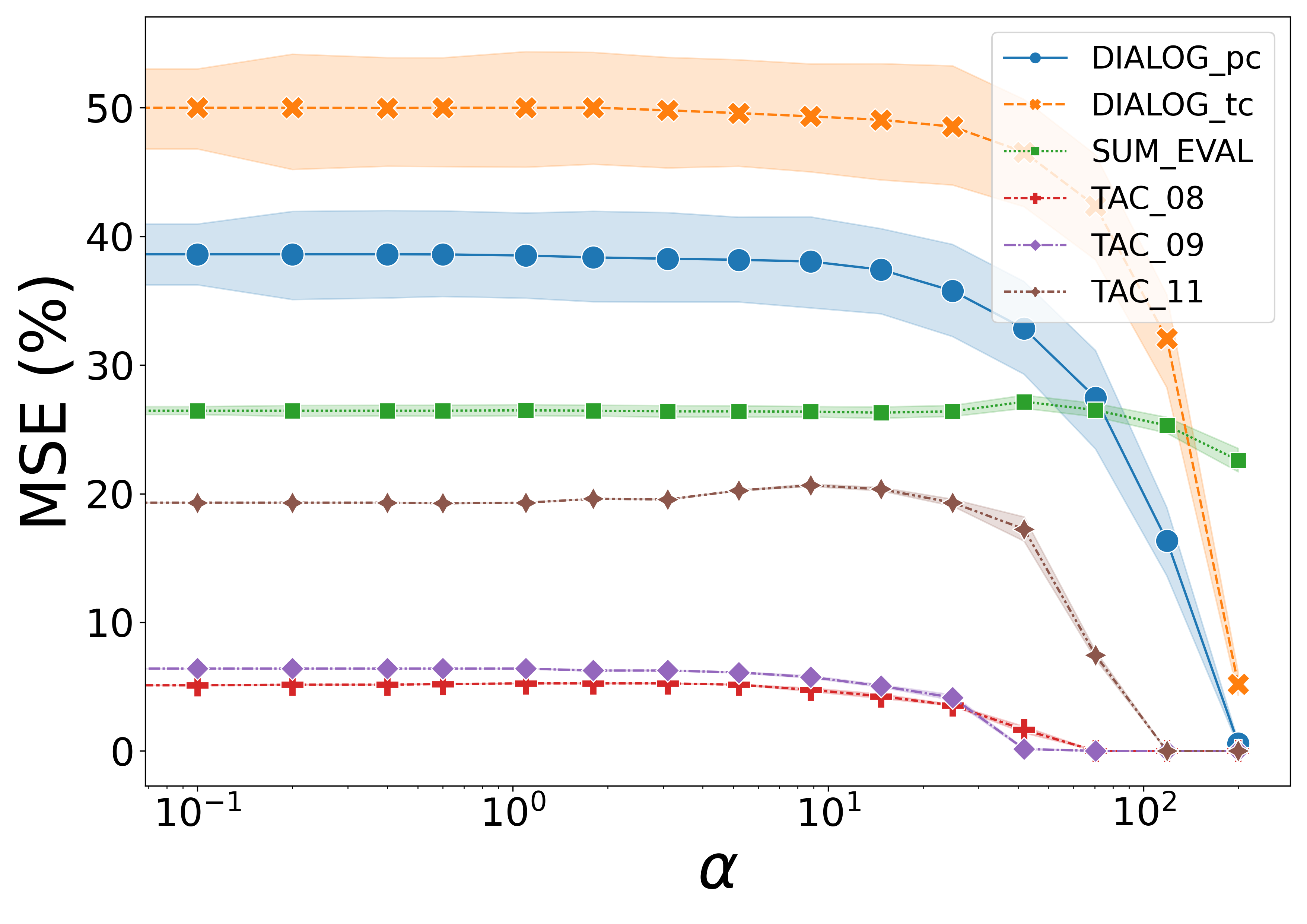}
         \caption{Aggregated score for each task when using System Level Representation}
         \label{fig:five over x}
     \end{subfigure}\\
          \centering
     \begin{subfigure}[b]{0.45\textwidth}
         \centering
         \includegraphics[width=\textwidth]{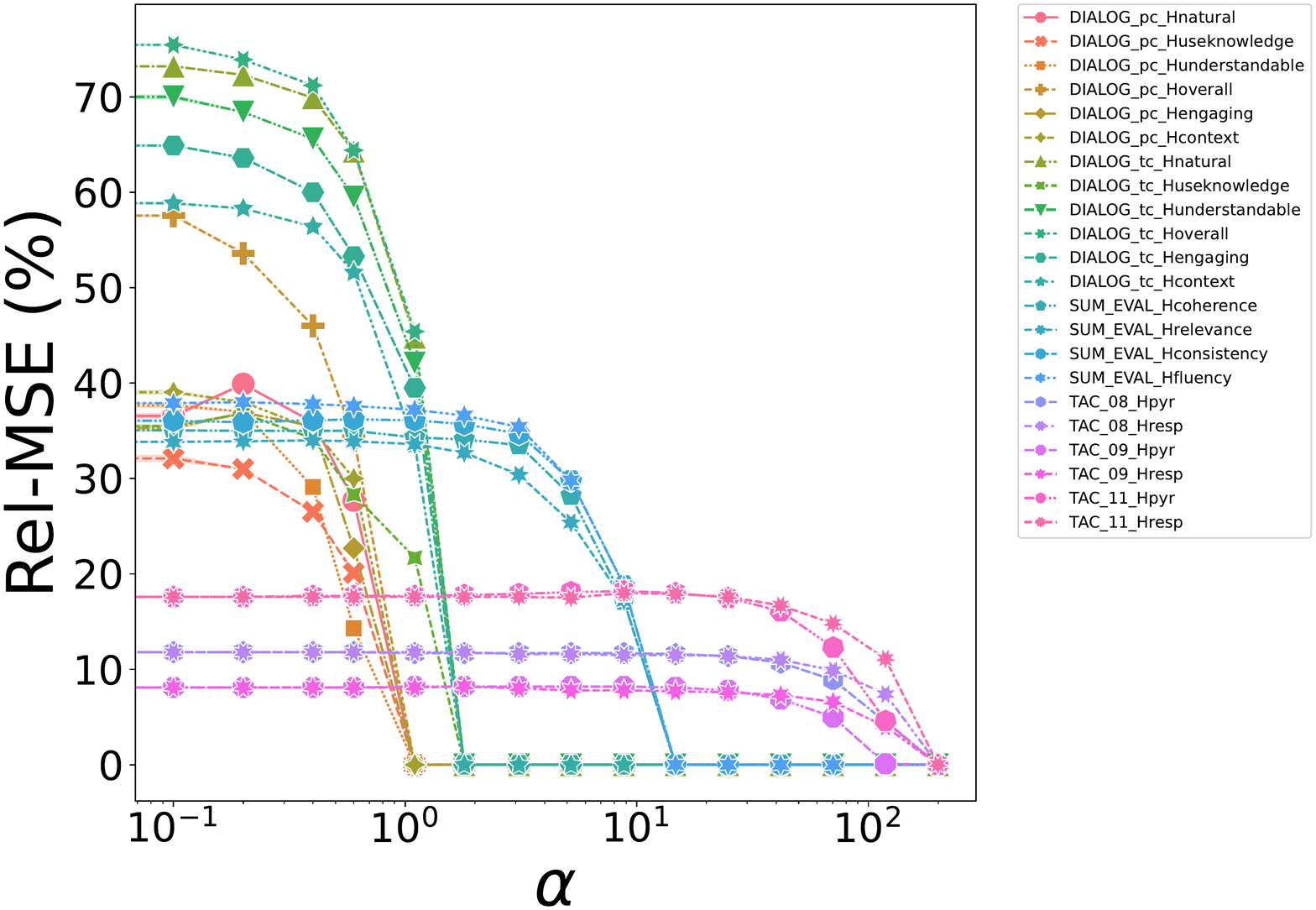}
         \caption{Detailed for each dataset when using System Level Representation}
         \label{fig:y equals x}
     \end{subfigure}
     \hfill
     \begin{subfigure}[b]{0.45\textwidth}
         \centering
         \includegraphics[width=\textwidth]{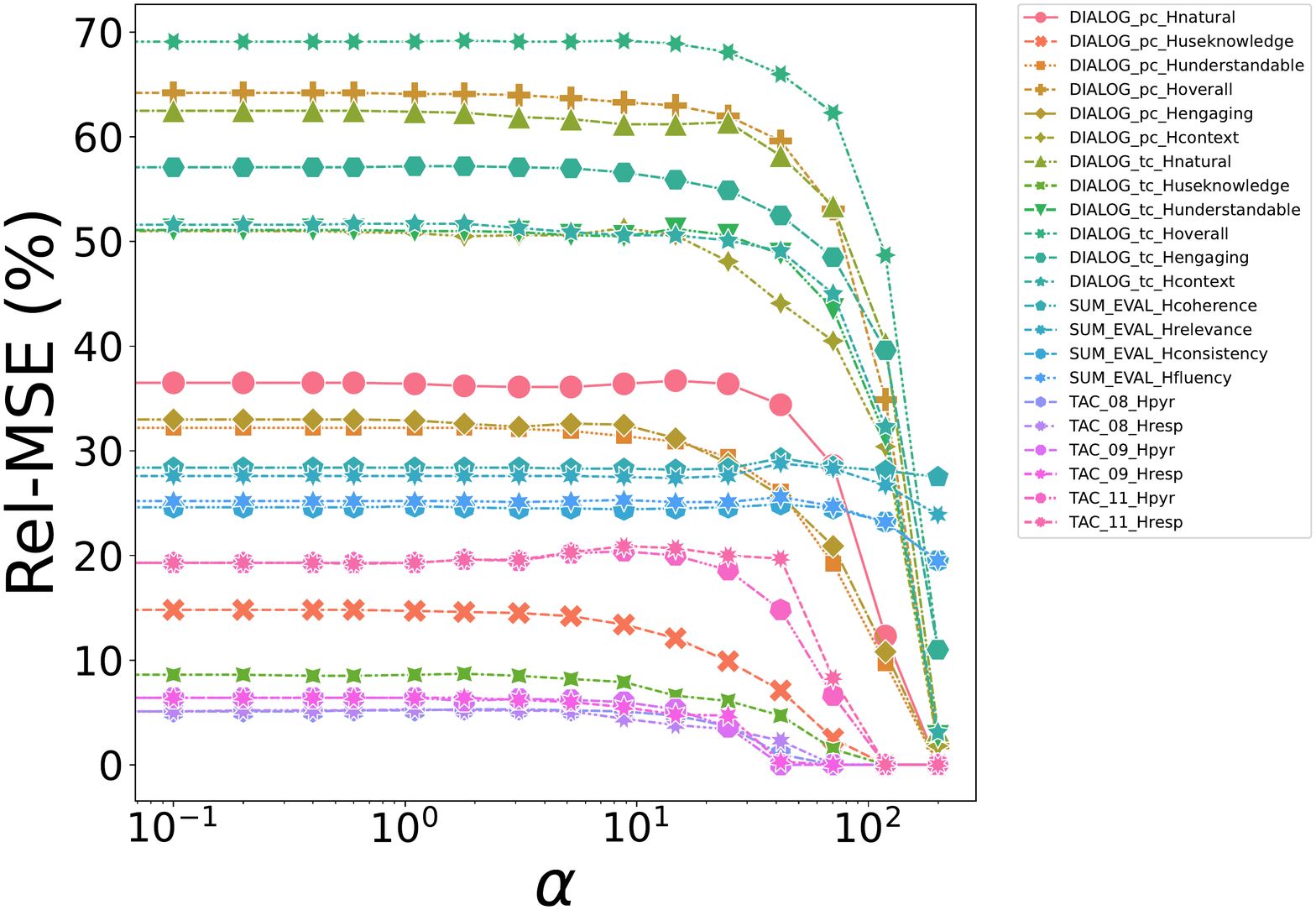}
         \caption{Detailed for each dataset when using utterance level Representation}
         \label{fig:five over x}
     \end{subfigure}
       \caption{\textbf{Human metrics contain useful information that is not in automatic metrics for predicting other human metrics.} On this plot, we report the ratio between the MSE-error of a linear regression trained with automatic metrics together with human metrics and a linear regression trained only with automatic metrics. For each dataset, we provide mean and variance corresponding to the prediction of available human metrics.}\label{fig:regression_lasso_aggregated}
 \end{figure*}

\noindent \texttt{Observations}: From the \autoref{fig:regression_lasso_aggregated}, we observe a strong decrease in error when adding human metrics to predict another human metric. When $\alpha$ increases, all the coefficients are set to 0, and the relative MSE is thus 0. It is worth noting that these observations hold for both system and utterance level representation. When observing the details per dataset, we observe a similar trend for all human metrics.

\noindent\texttt{Takeaways}: When predicting a specific human metric, other human metrics contain useful predictive information that is not present in the automatic metric.

\subsection{Which metrics are the most useful to predict human judgment at the System level?}

\noindent For this experiment we will rely on a Lasso Regression and denote the multiplier of the L1 term $\alpha$. For several values of $\alpha$ (x-axis), we report each metric's weights (y-axis) in Figures 8 and 9.

\noindent \texttt{Observations}: When increasing the weights given to the L1 penalization term, we observe that the regression weights of the human metrics are the ones that are the last to be set to 0. Human metrics contain the most relevant information. It is worth noting that this phenomenon is generic across the datasets and human criteria.

\noindent \texttt{Takeaways:} Human metrics are the most useful metrics when predicting other metrics.

\begin{figure}[!ht] \label{fig:last_fig_1}
\centering
             \begin{subfigure}[b]{0.3\textwidth}
         \centering
         \includegraphics[width=\textwidth]{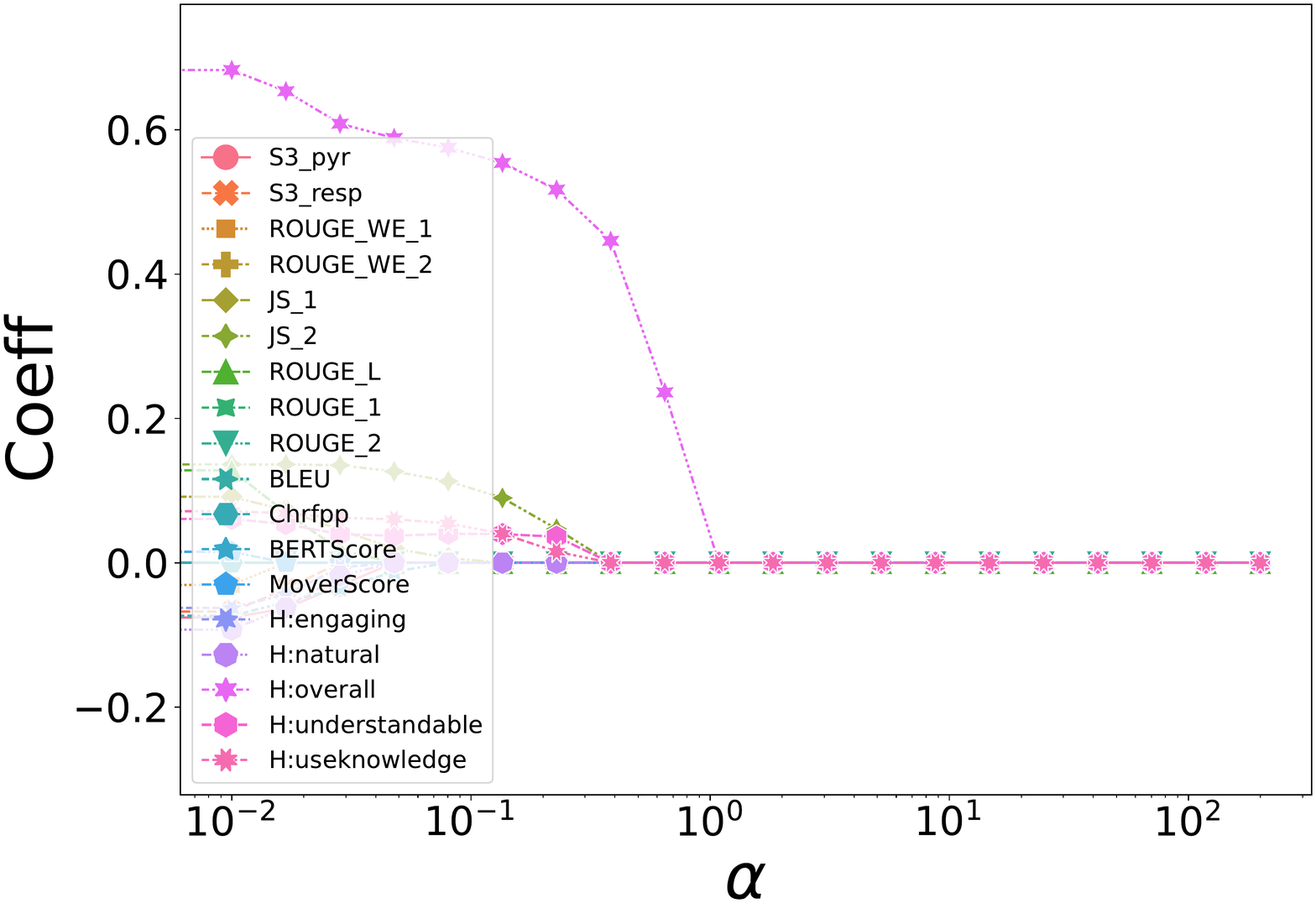}
         \caption{Dialog PC Context}
         \label{fig:y equals x}
     \end{subfigure}
      \begin{subfigure}[b]{0.3\textwidth}
         \centering
         \includegraphics[width=\textwidth]{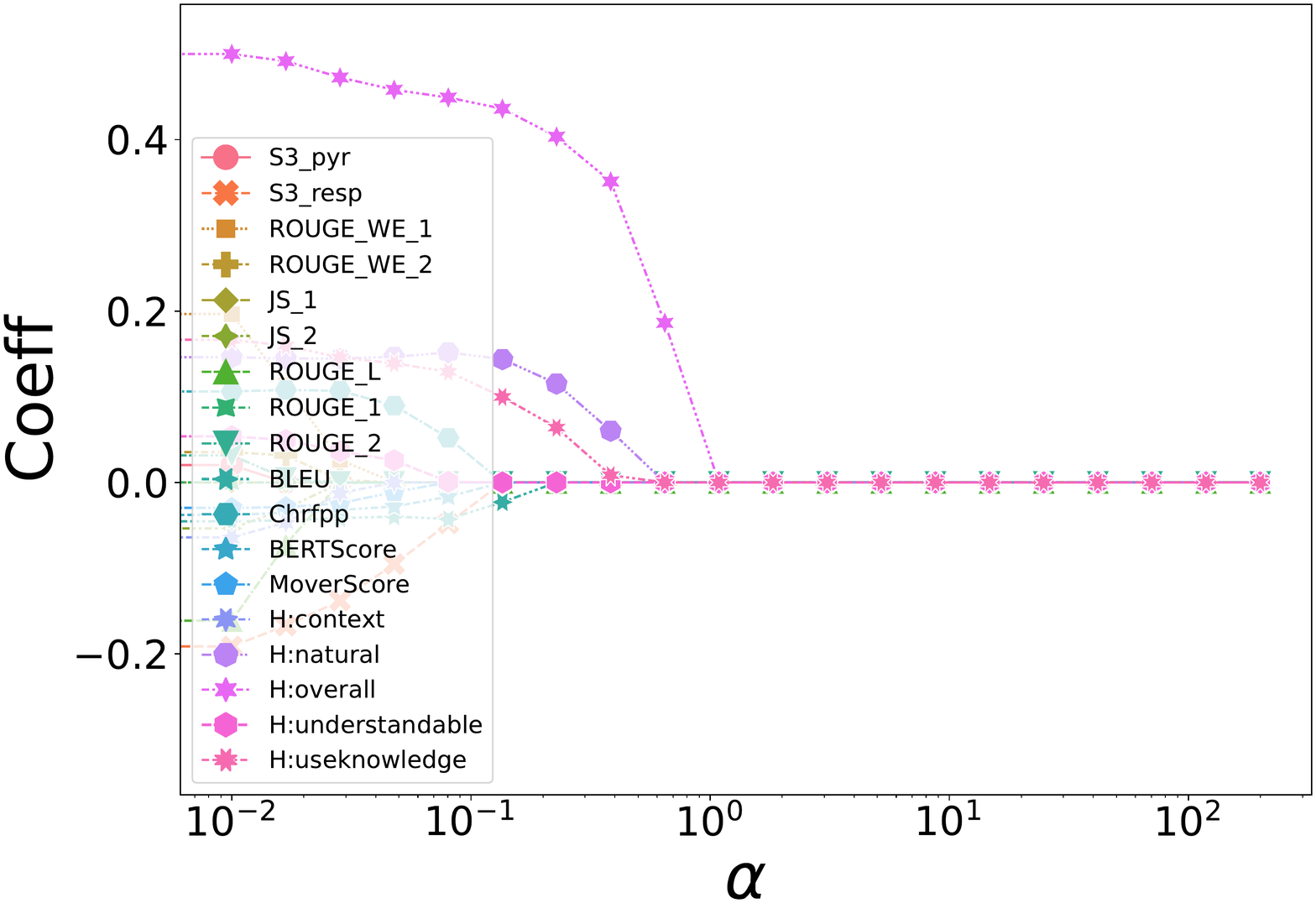}
         \caption{Dialog PC Engaging}
         \label{fig:y equals x}
     \end{subfigure}
         \begin{subfigure}[b]{0.3\textwidth}
         \centering
         \includegraphics[width=\textwidth]{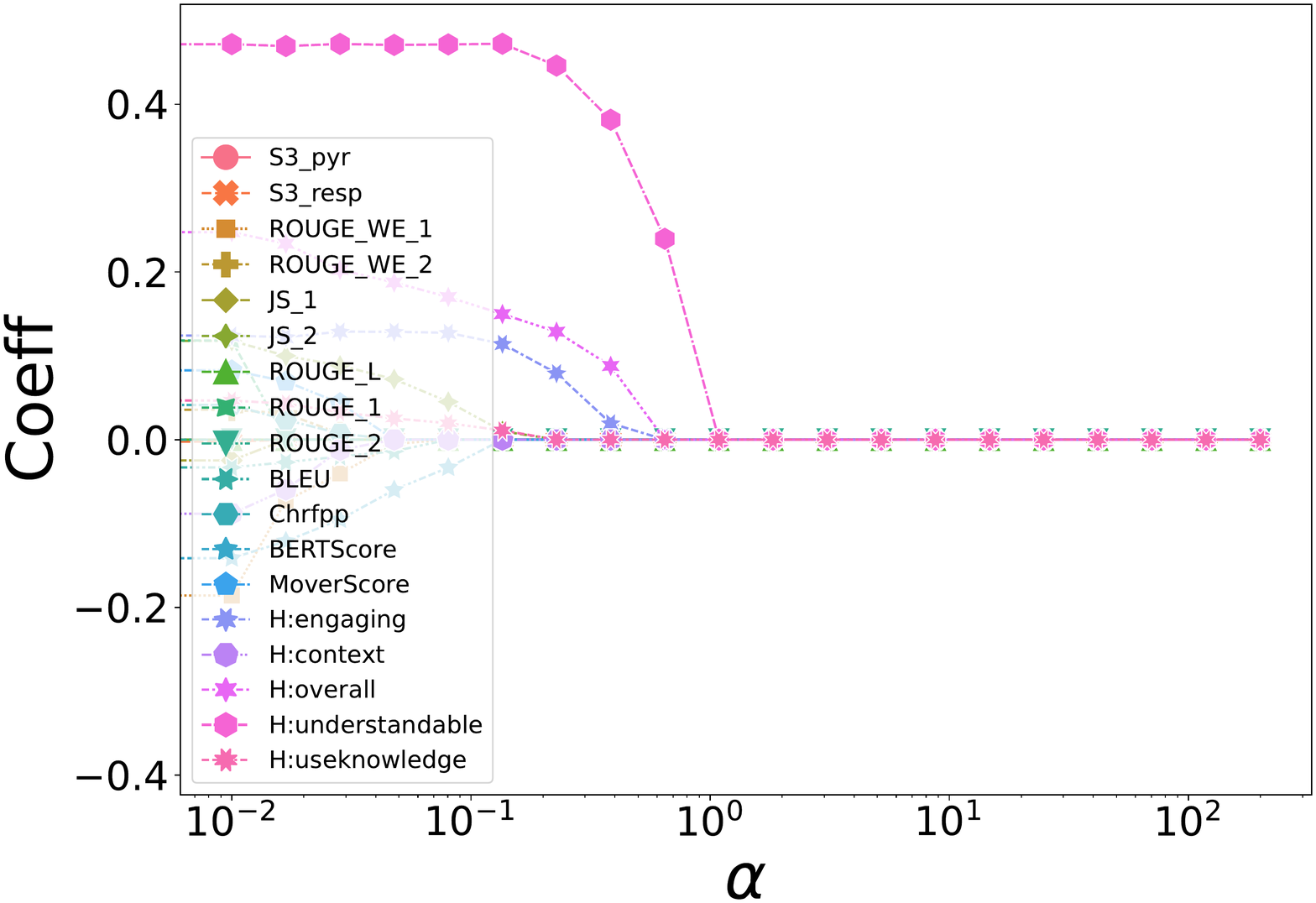}
         \caption{Dialog PC Natural}
         \label{fig:y equals x}
         
     \end{subfigure}
              \begin{subfigure}[b]{0.3\textwidth}
         \centering
         \includegraphics[width=\textwidth]{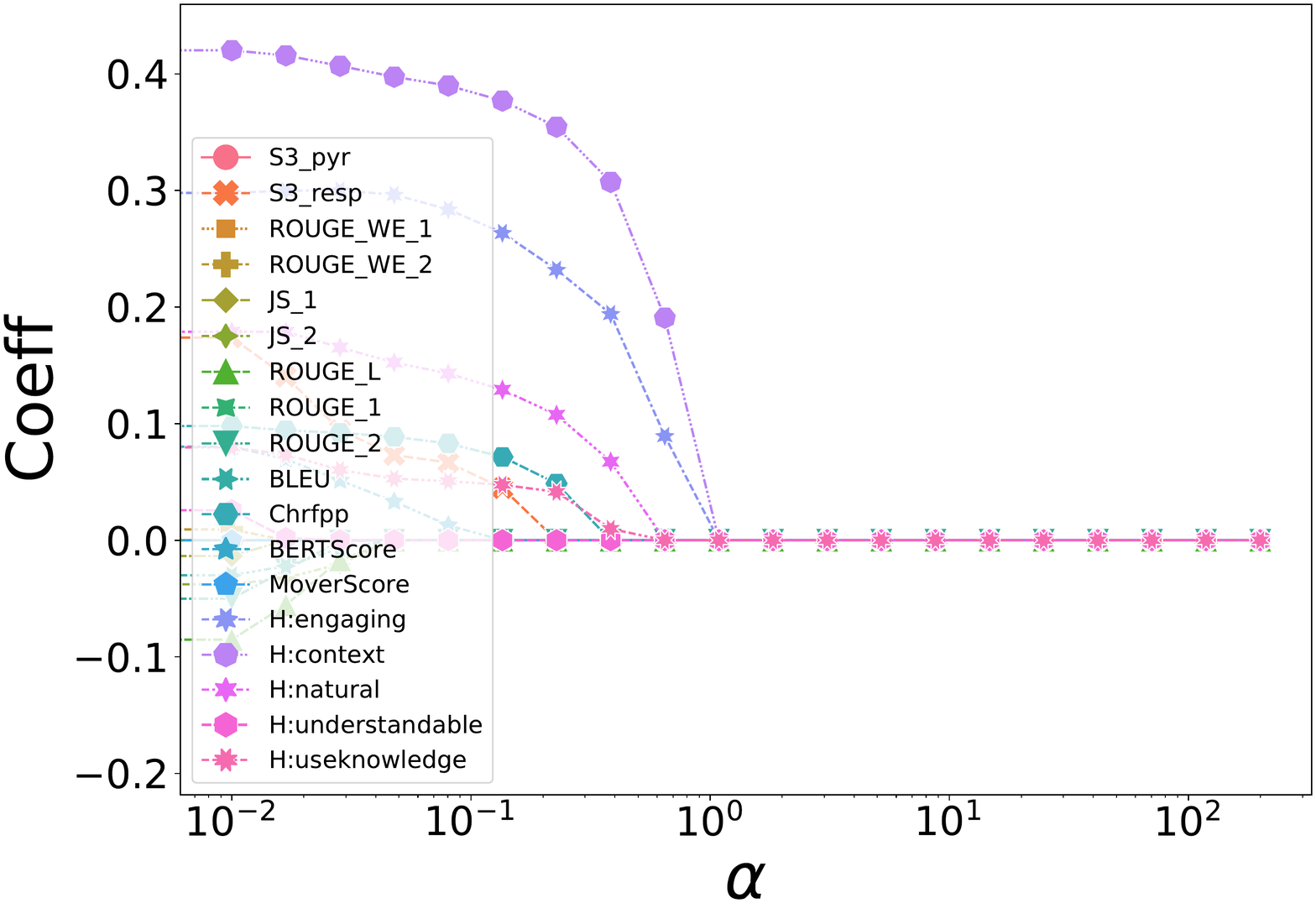}
         \caption{Dialog PC Overall}
         \label{fig:y equals x}
     \end{subfigure}
                   \begin{subfigure}[b]{0.3\textwidth}
         \centering
         \includegraphics[width=\textwidth]{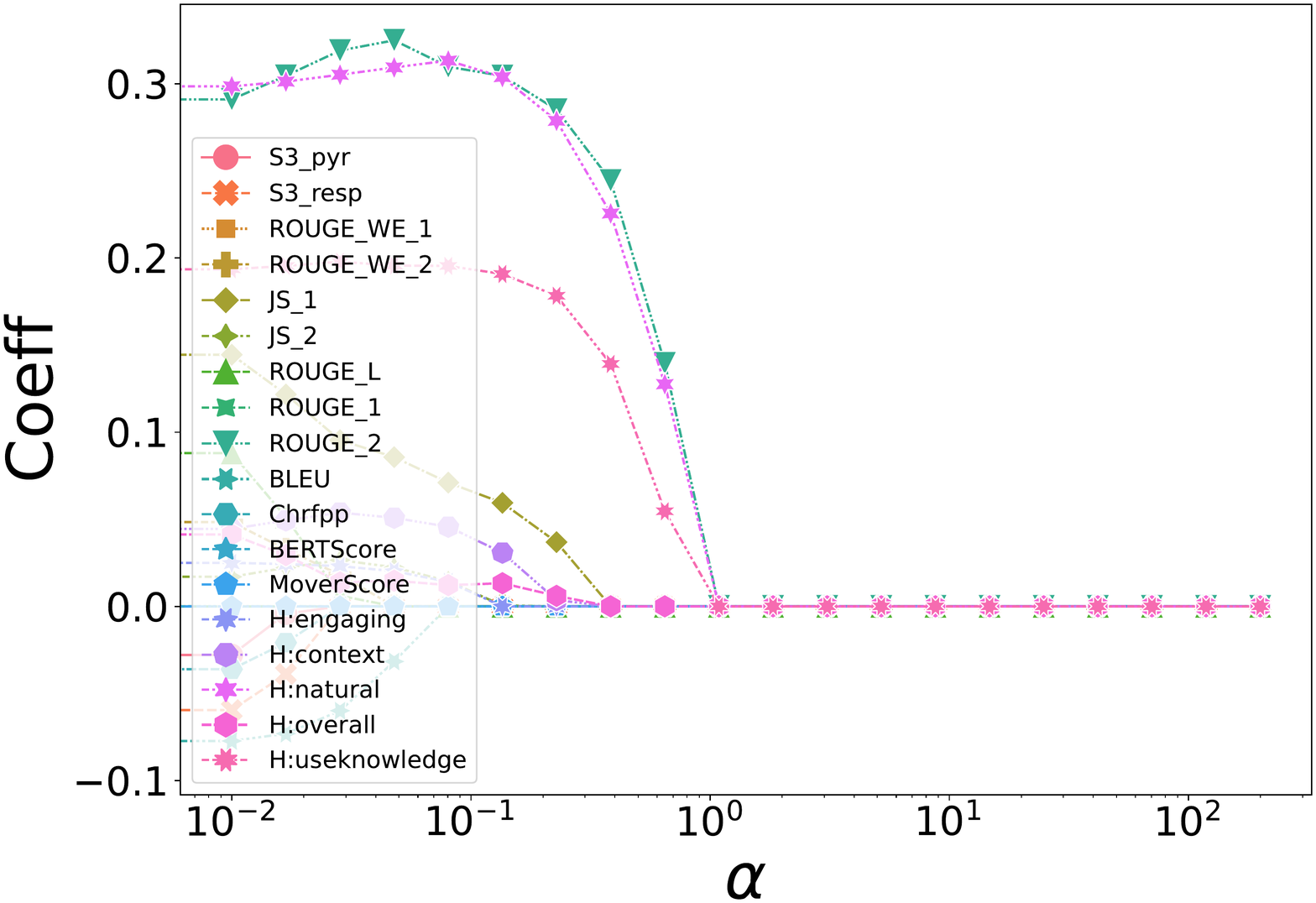}
         \caption{Dialog PC Understandable}
         \label{fig:y equals x}
     \end{subfigure}
      \begin{subfigure}[b]{0.3\textwidth}
         \centering
         \includegraphics[width=\textwidth]{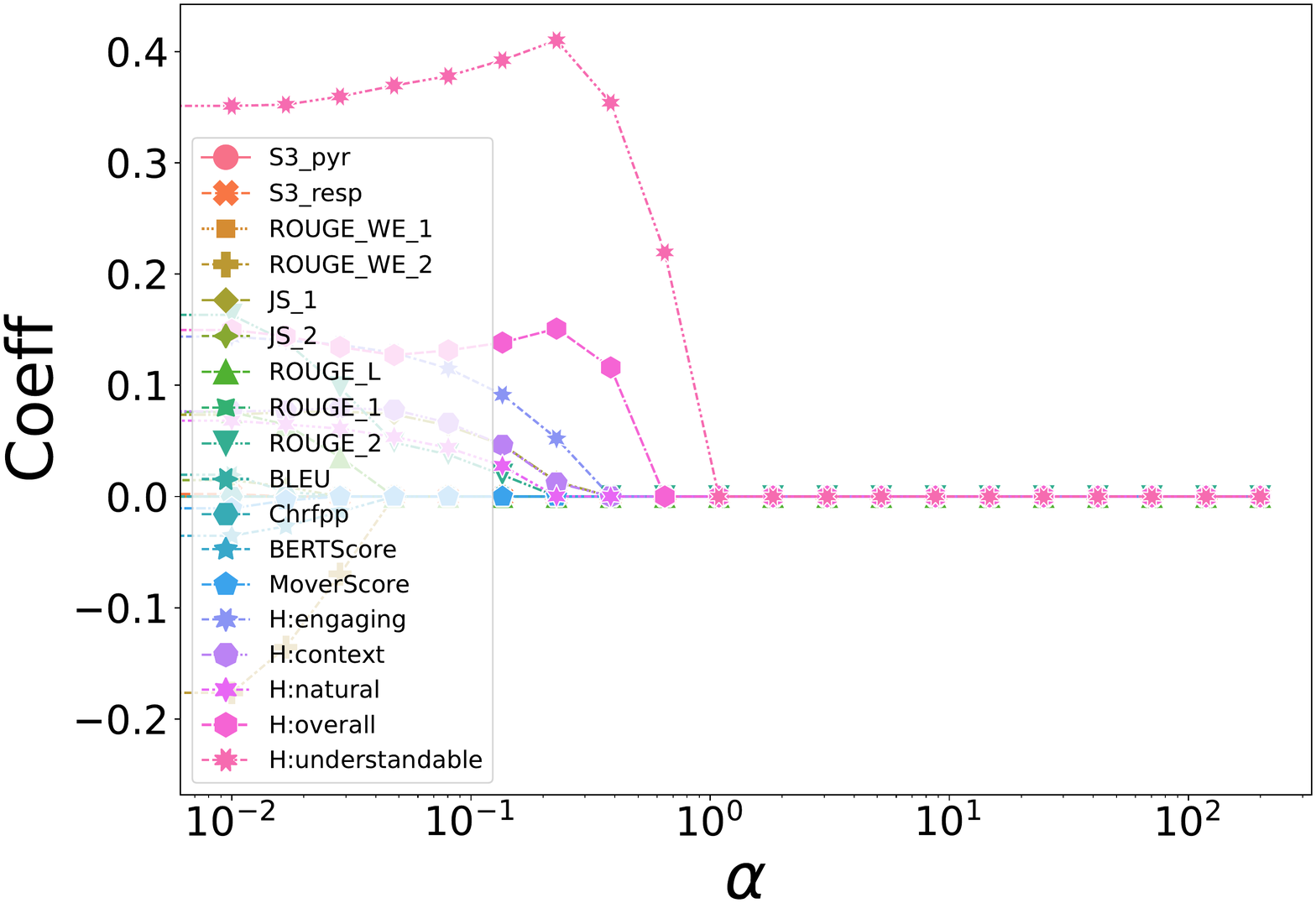}
         \caption{Dialog PC Knowledgable}
         \label{fig:y equals x}
     \end{subfigure}
     
             \begin{subfigure}[b]{0.3\textwidth}
         \centering
         \includegraphics[width=\textwidth]{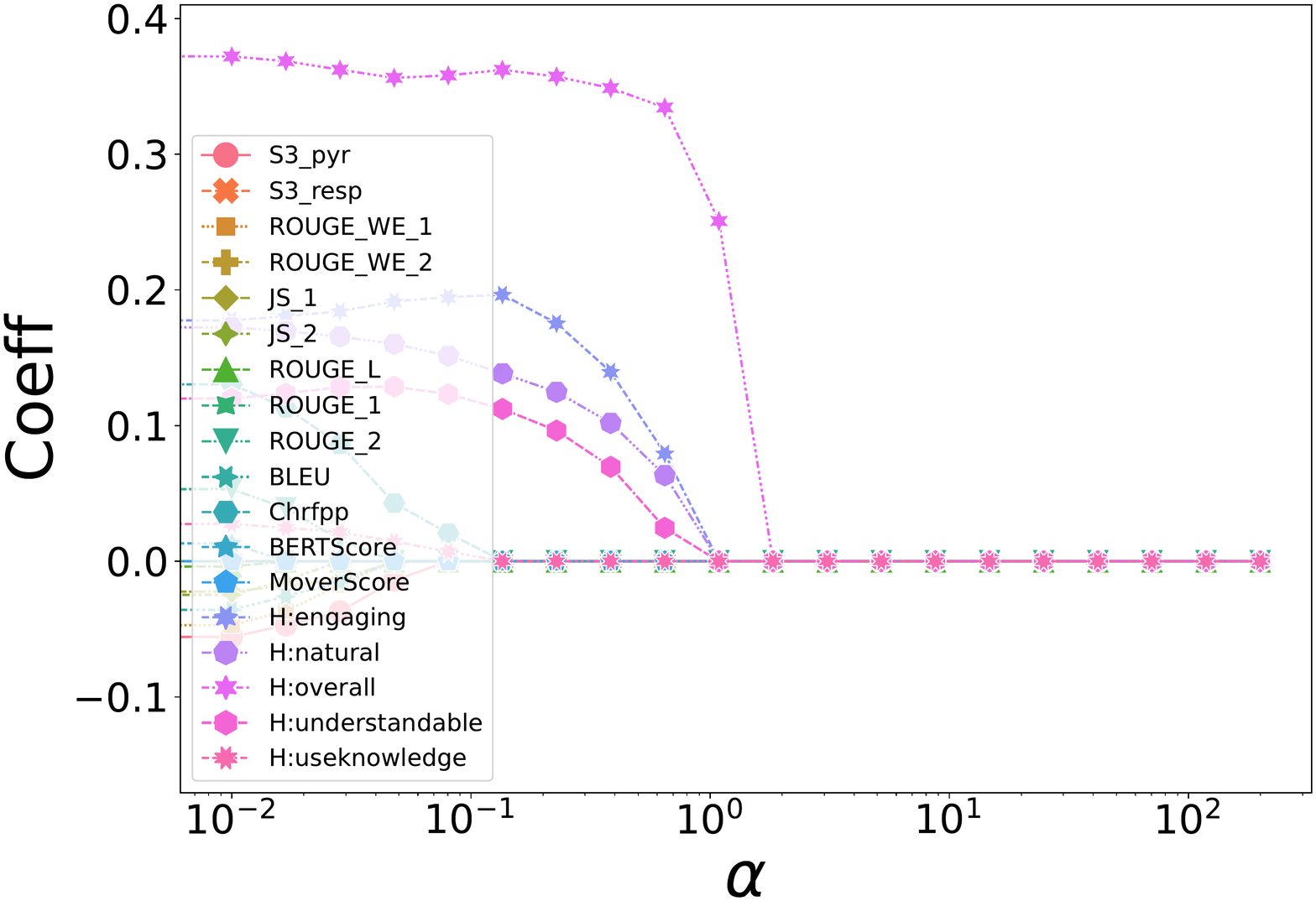}
         \caption{Dialog TC Context}
         \label{fig:y equals x}
     \end{subfigure}
      \begin{subfigure}[b]{0.3\textwidth}
         \centering
         \includegraphics[width=\textwidth]{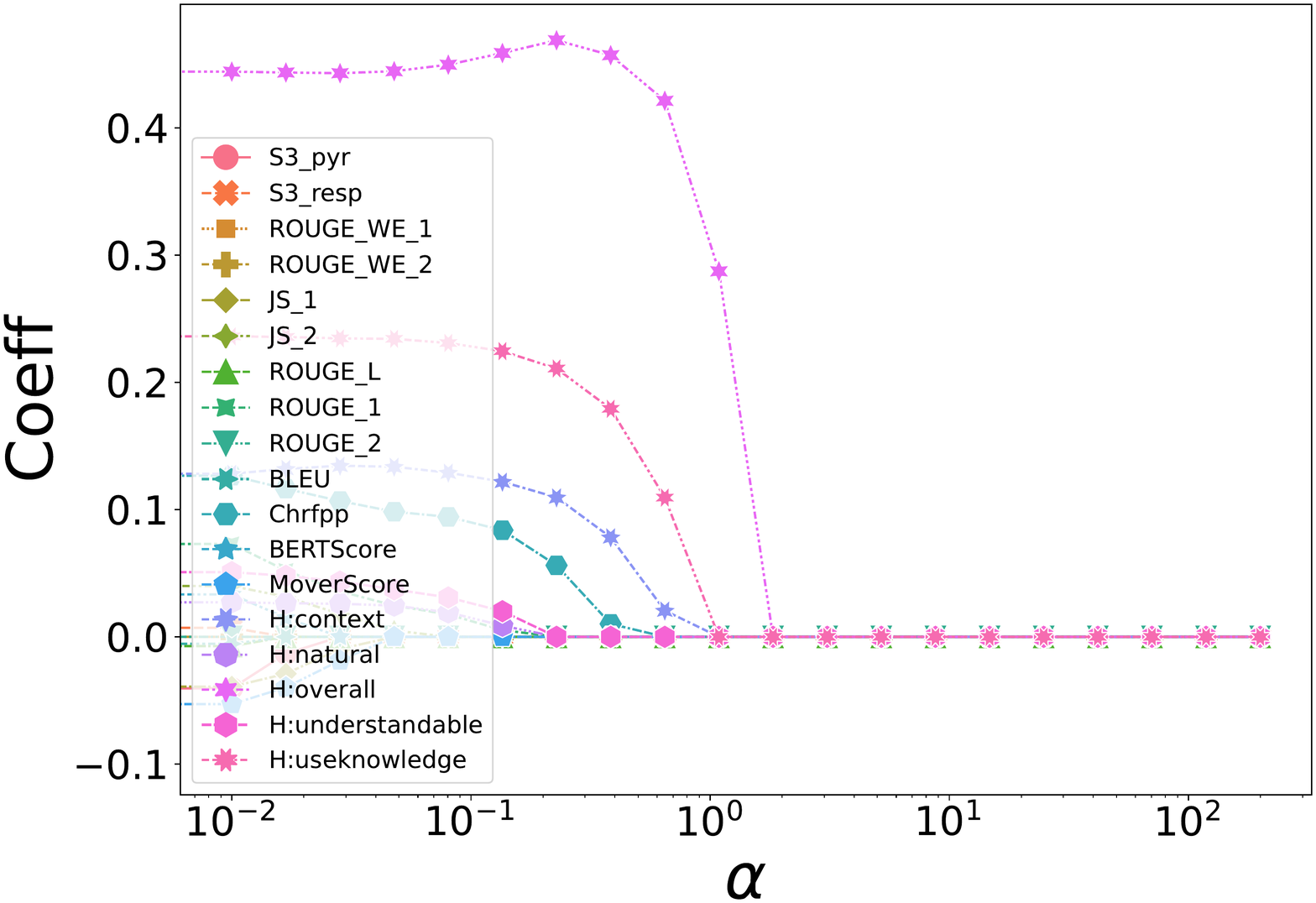}
         \caption{Dialog TC Engaging}
         \label{fig:y equals x}
     \end{subfigure}
         \begin{subfigure}[b]{0.3\textwidth}
         \centering
         \includegraphics[width=\textwidth]{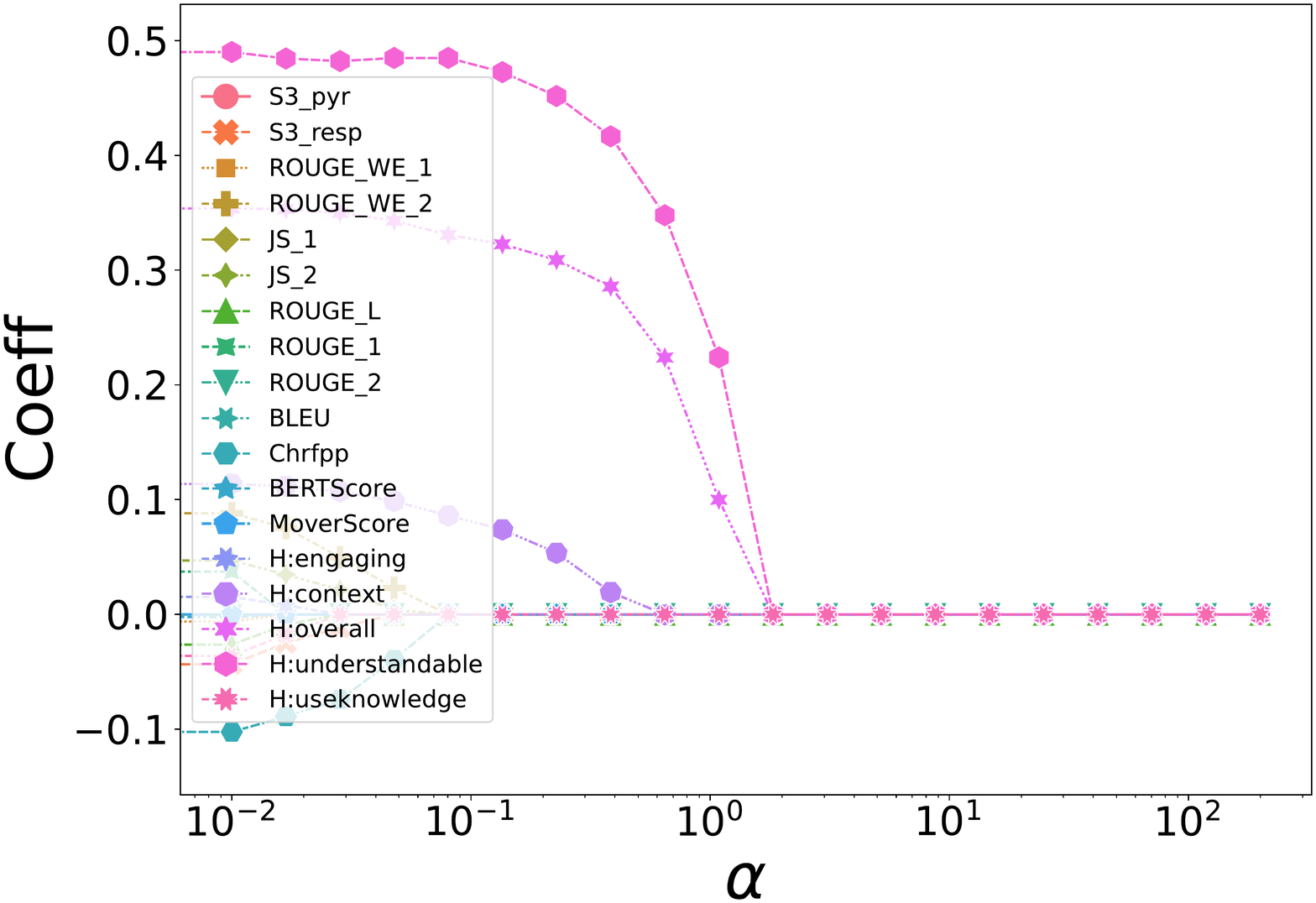}
         \caption{Dialog TC Natural}
         \label{fig:y equals x}
     \end{subfigure}
     
     \begin{subfigure}[b]{0.3\textwidth}
         \centering
         \includegraphics[width=\textwidth]{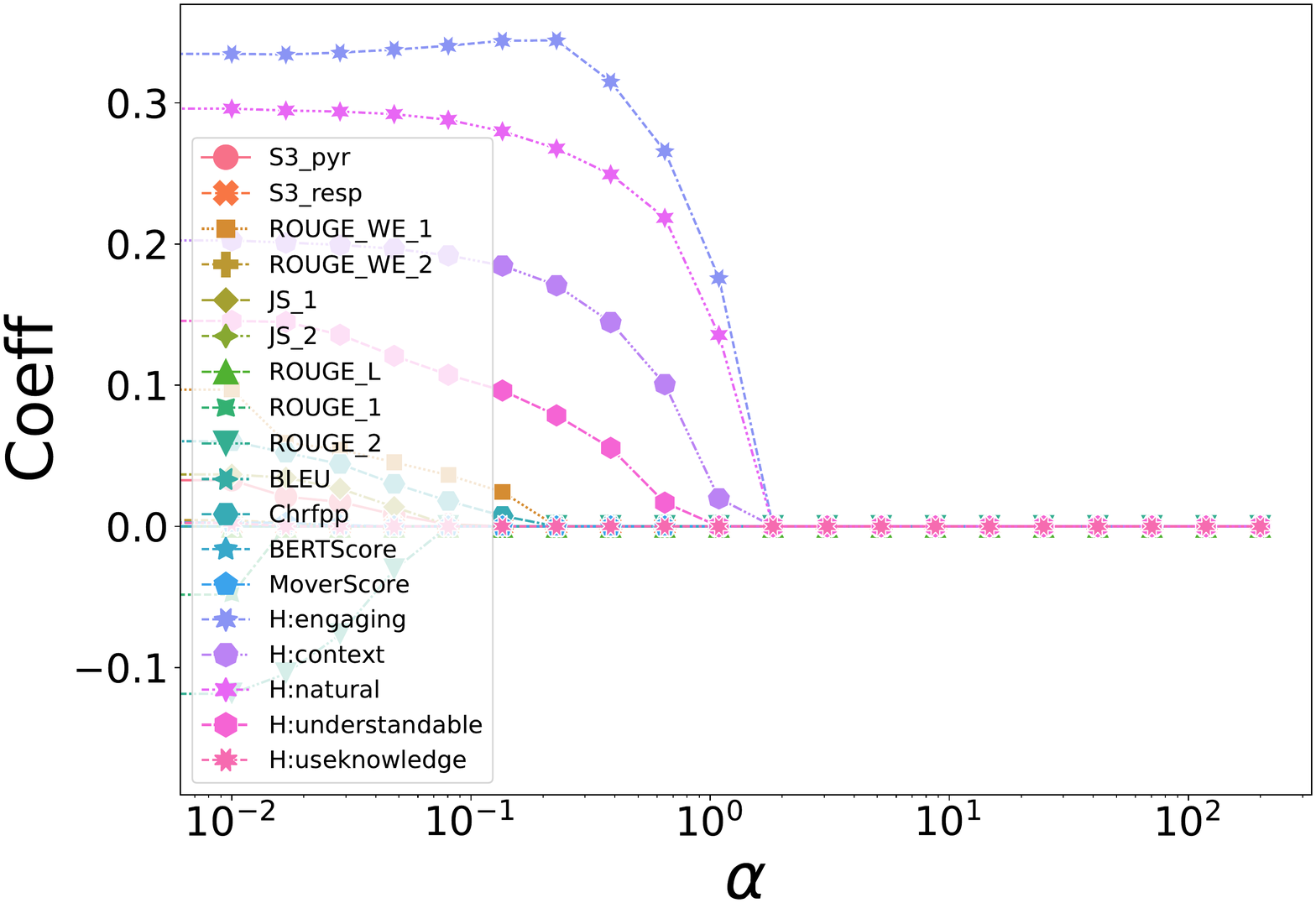}
         \caption{Dialog TC Overall}
         \label{fig:y equals x}
     \end{subfigure}\begin{subfigure}[b]{0.3\textwidth}
         \centering
         \includegraphics[width=\textwidth]{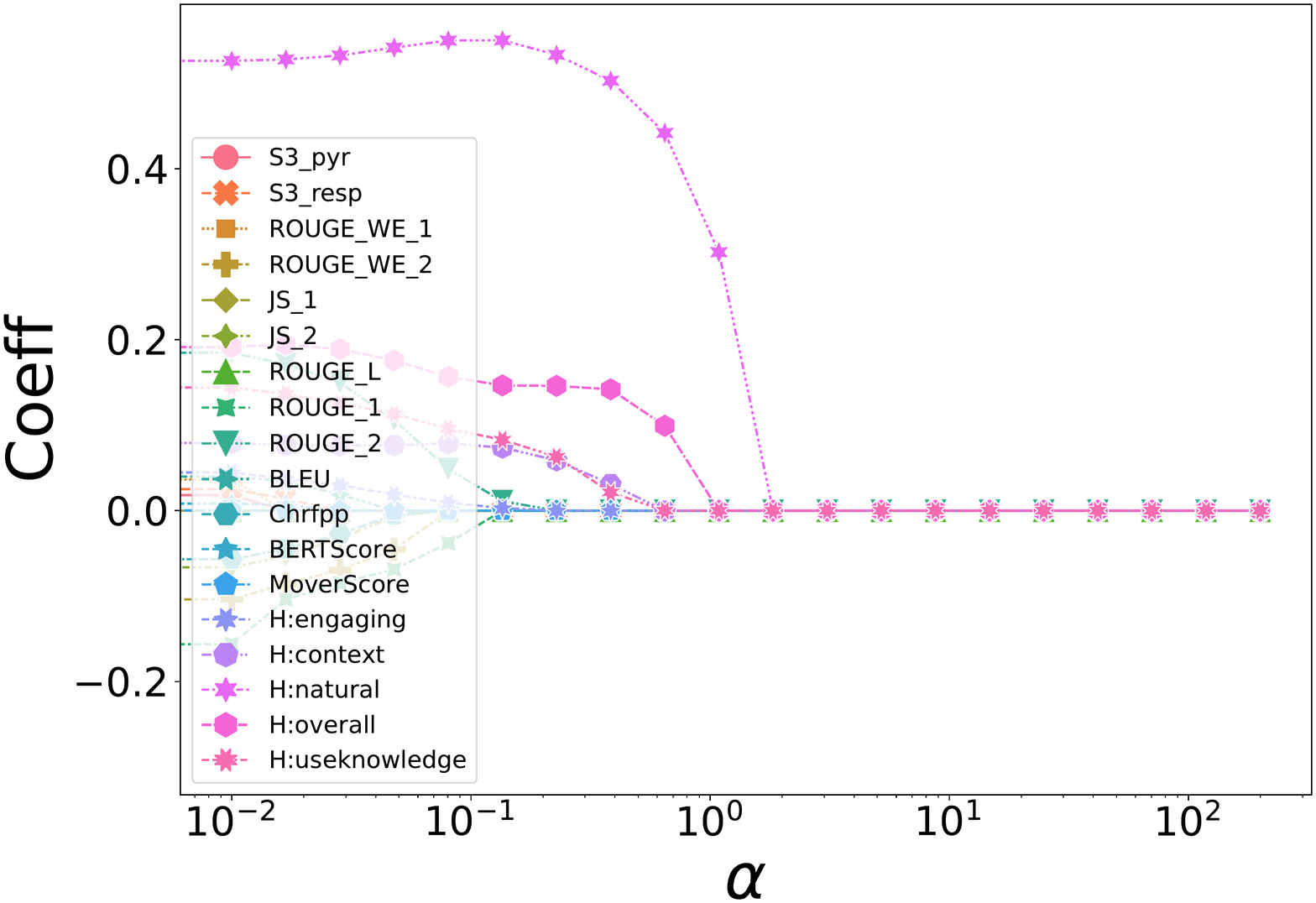}
         \caption{Dialog TC Understandable}
         \label{fig:y equals x}
     \end{subfigure}\begin{subfigure}[b]{0.3\textwidth}
         \centering
         \includegraphics[width=\textwidth]{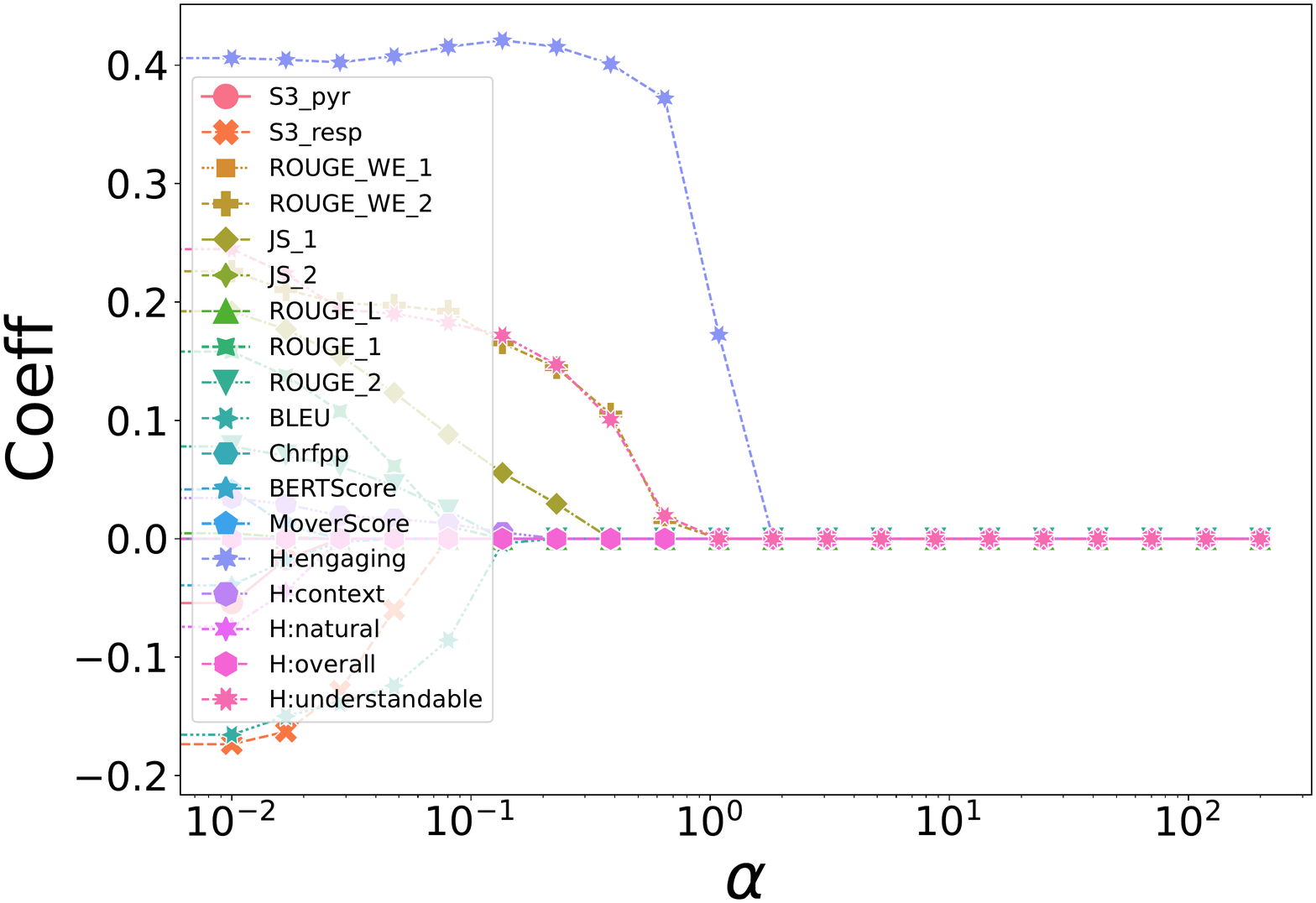}
         \caption{Dialog TC Knowledgable}
         \label{fig:y equals x}
     \end{subfigure}\label{fig:system_level_}
          \caption{\textbf{Human metrics are the most useful metrics when predicting other metrics.} Regression weights (y-axis) obtained by each metric when training a Lasso Regression to predict a human metric for different regularization coefficients (x-axis) on the system level representation of the metrics. } 
     \end{figure}

     \begin{figure}
     \centering
             \begin{subfigure}[b]{0.3\textwidth}
         \centering
         \includegraphics[width=\textwidth]{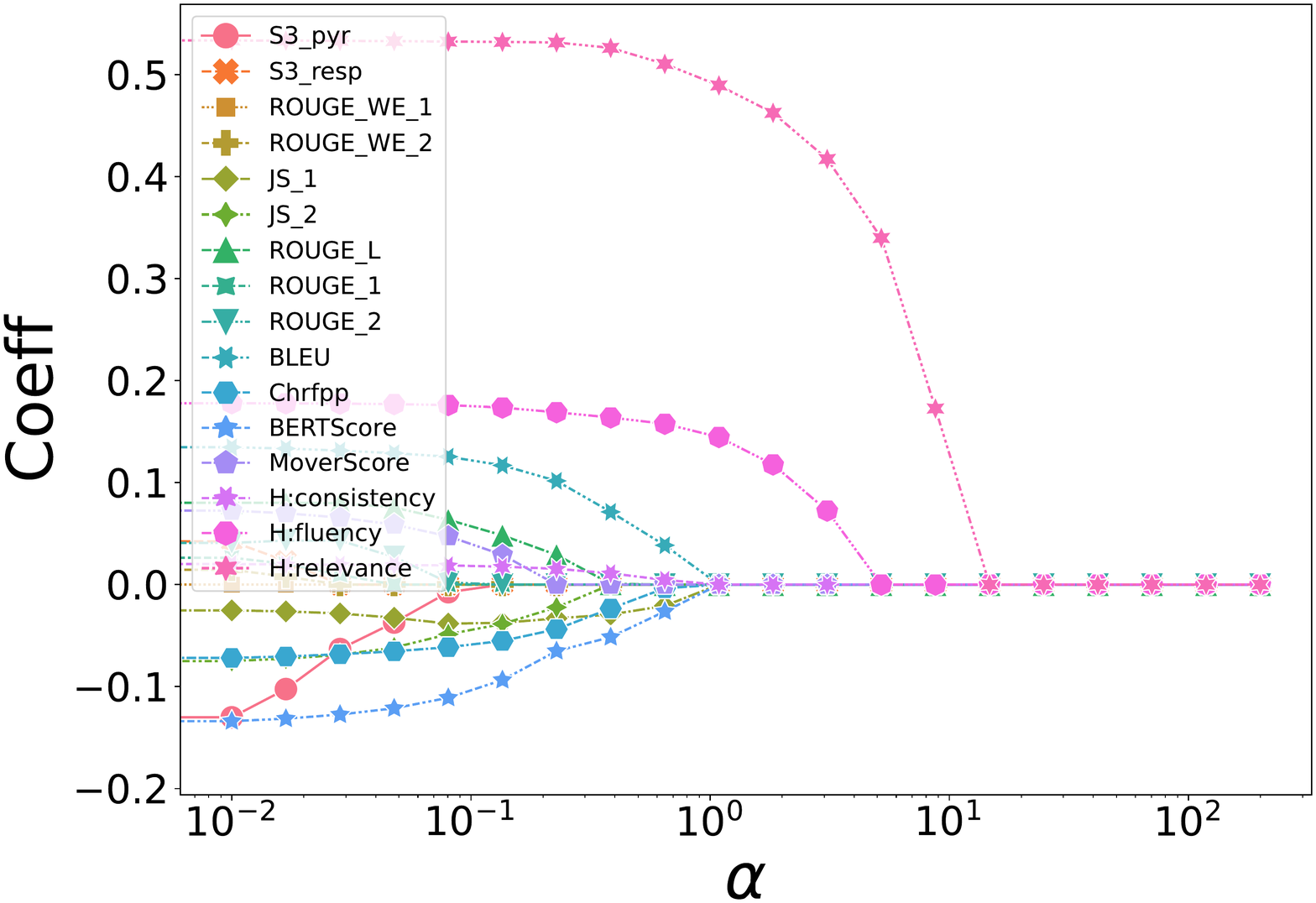}
         \caption{SUM EVAL Coherence}
         \label{fig:y equals x}
     \end{subfigure}
      \begin{subfigure}[b]{0.3\textwidth}
         \centering
         \includegraphics[width=\textwidth]{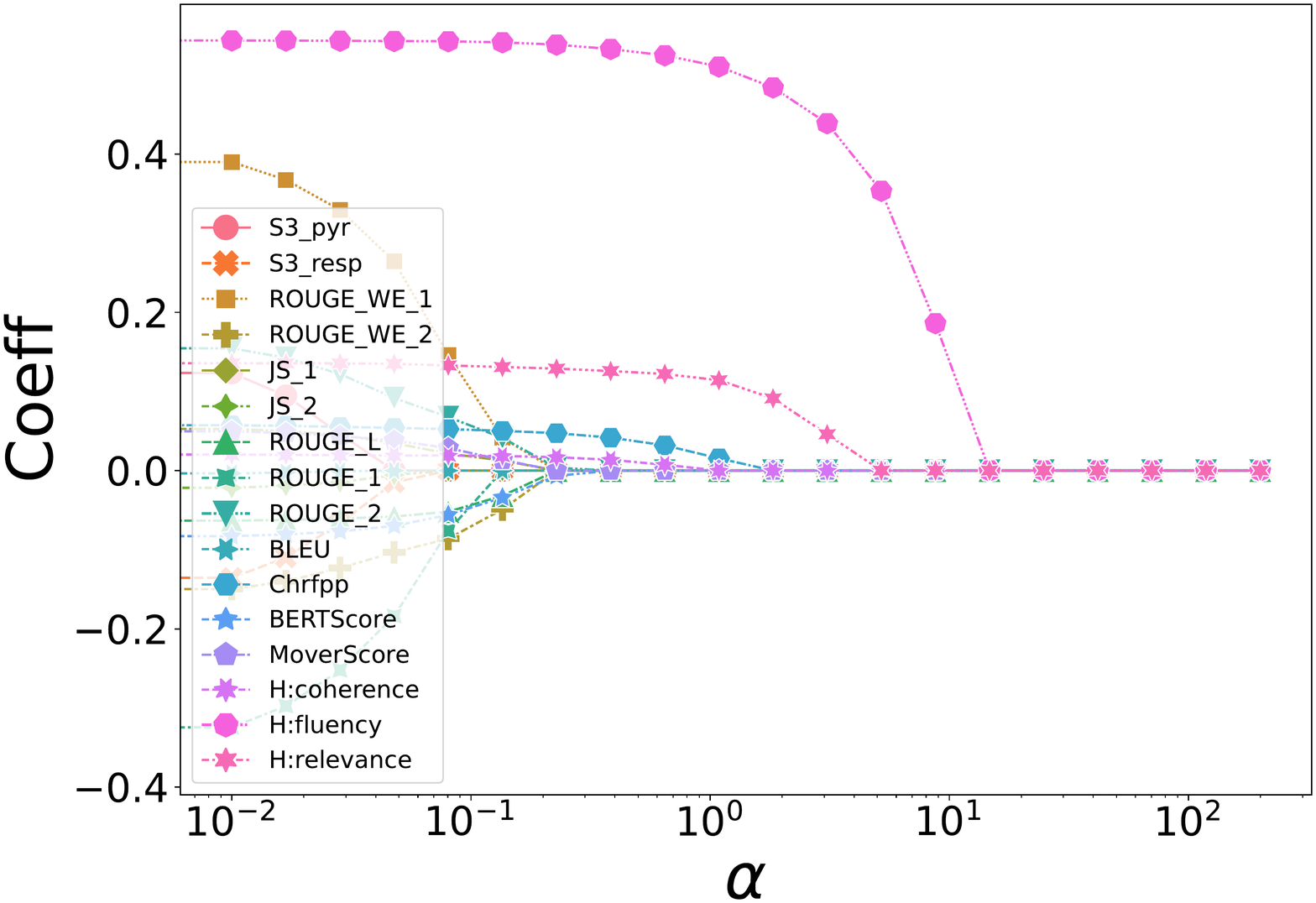}
         \caption{SUM EVAL Consistency}
         \label{fig:y equals x}
     \end{subfigure}
         \begin{subfigure}[b]{0.3\textwidth}
         \centering
         \includegraphics[width=\textwidth]{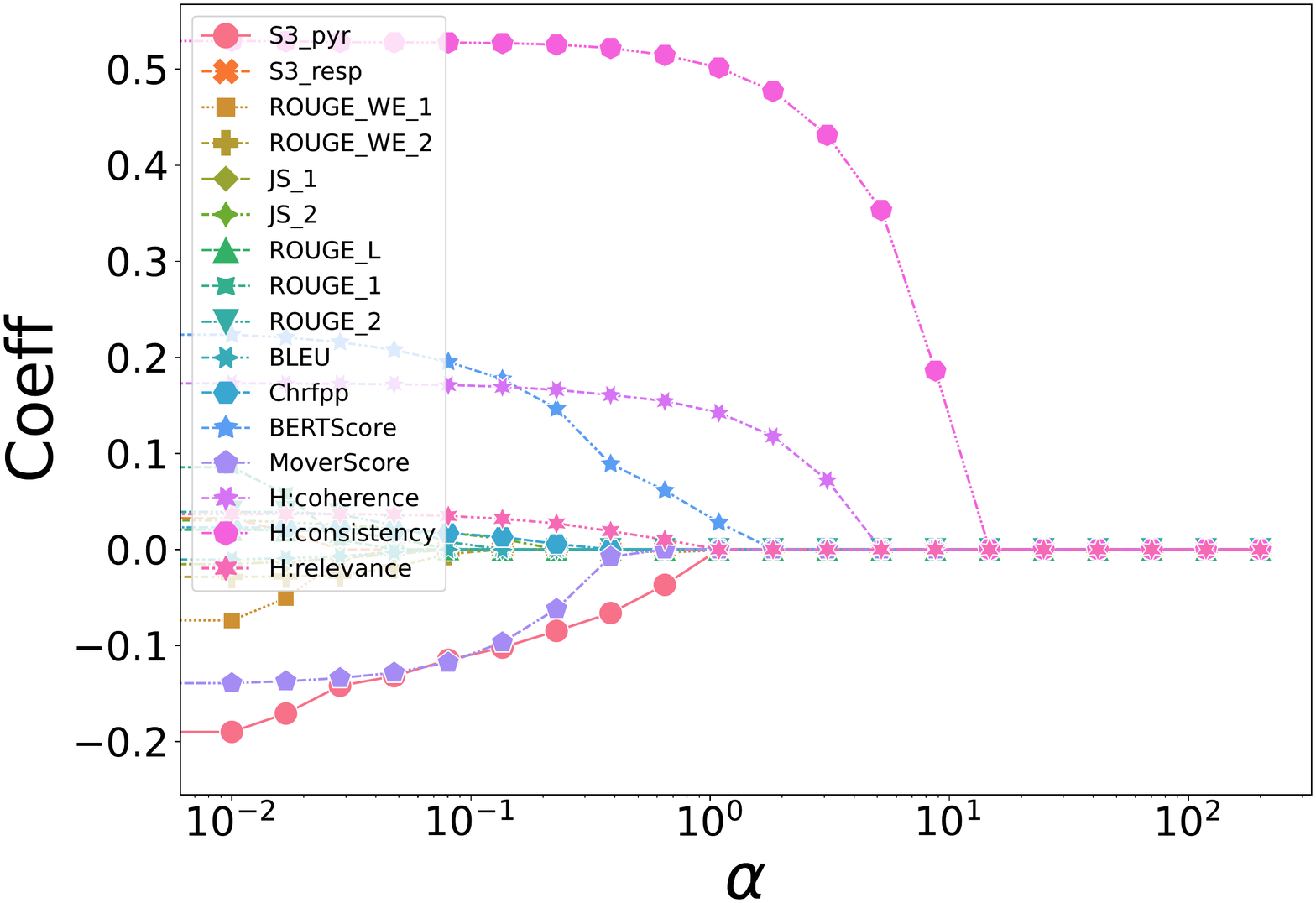}
         \caption{SUM EVAL Fluency}
         \label{fig:y equals x}
     \end{subfigure}
     
     \centering
              \begin{subfigure}[b]{0.3\textwidth}
         \centering
         \includegraphics[width=\textwidth]{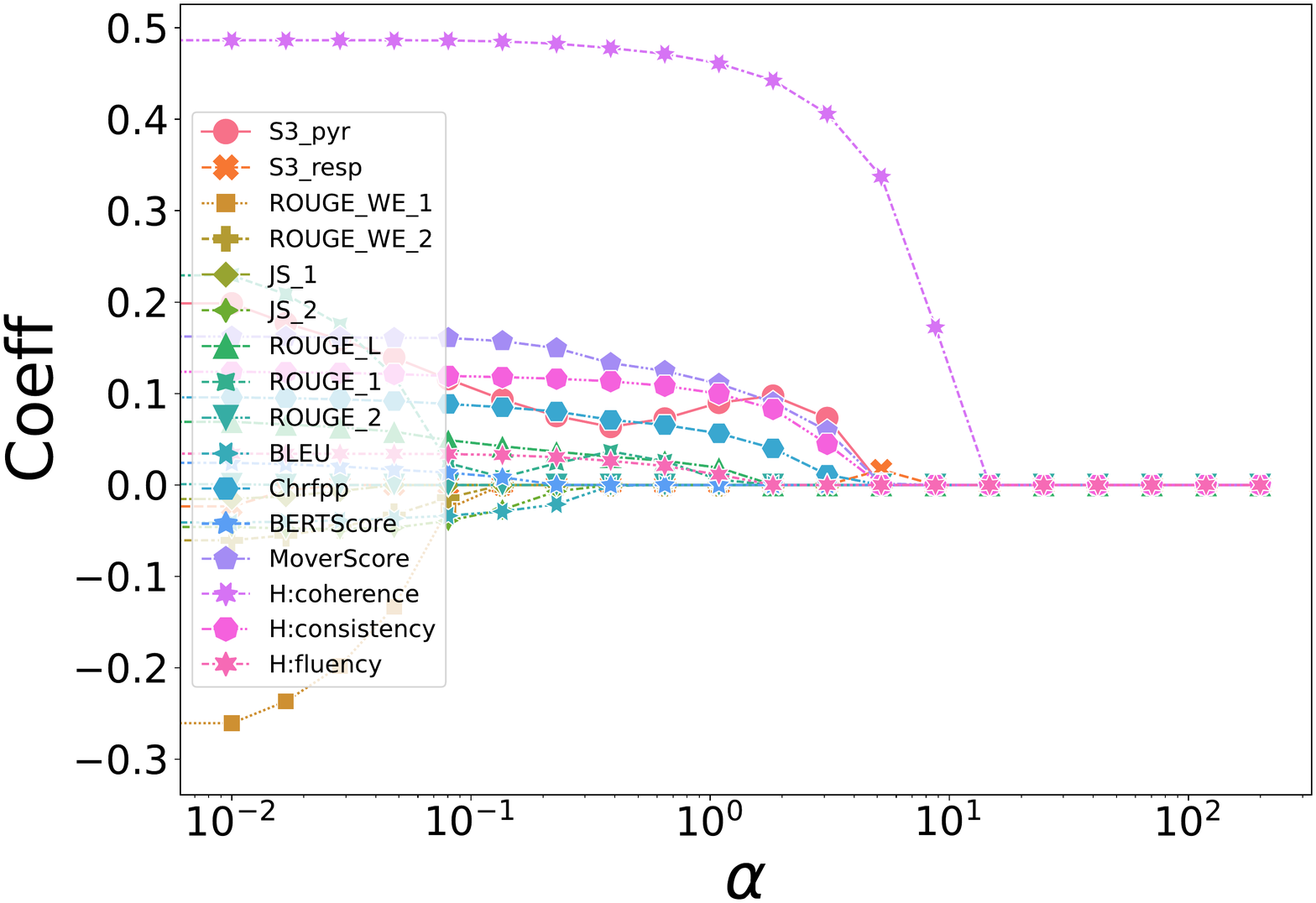}
         \caption{SUM EVAL Relevance}
         \label{fig:y equals x}
     \end{subfigure}
         \begin{subfigure}[b]{0.3\textwidth}
         \centering
         \includegraphics[width=\textwidth]{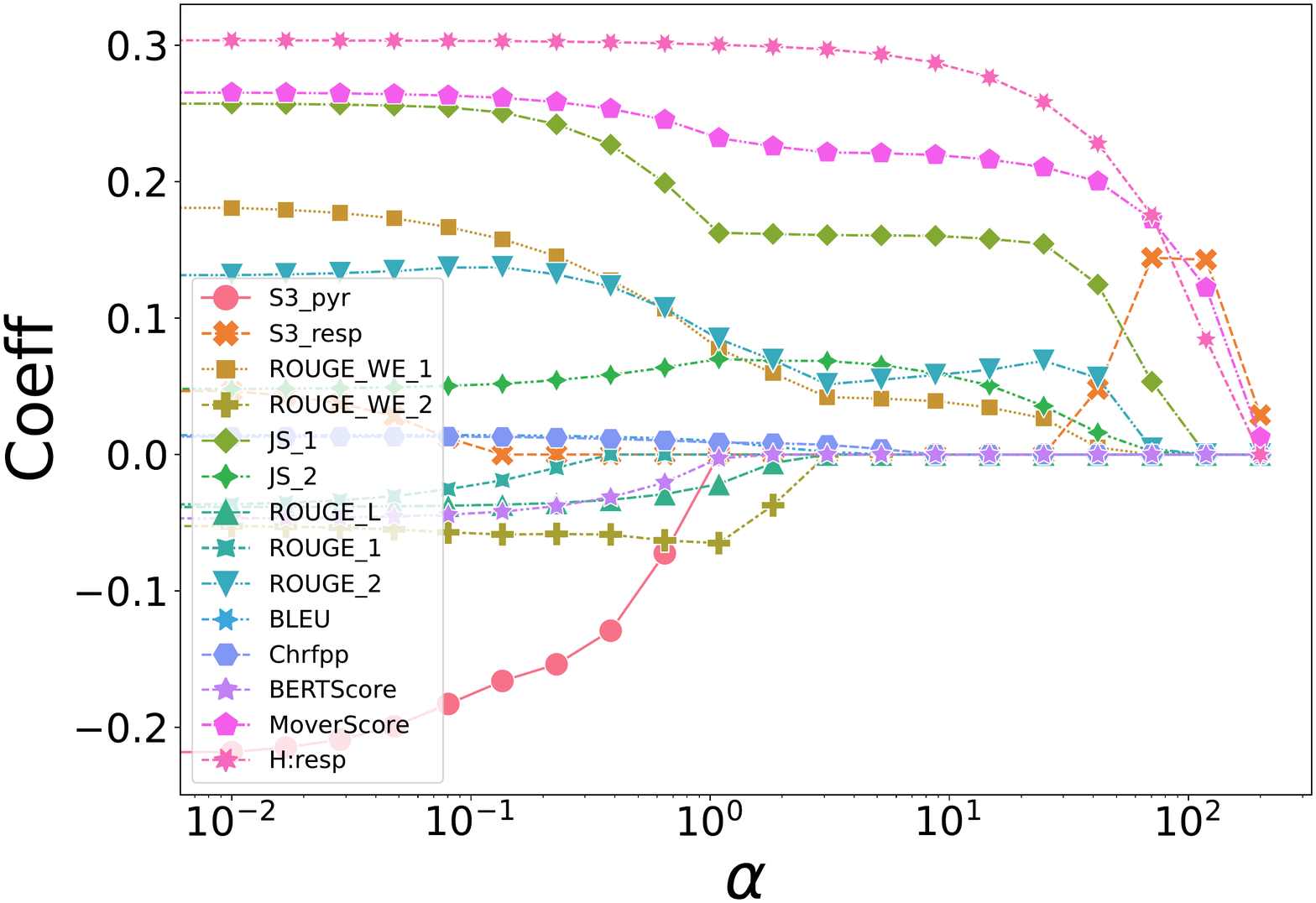}
         \caption{TAC8 Pyramide}
         \label{fig:y equals x}
     \end{subfigure}
              \begin{subfigure}[b]{0.3\textwidth}
         \centering
         \includegraphics[width=\textwidth]{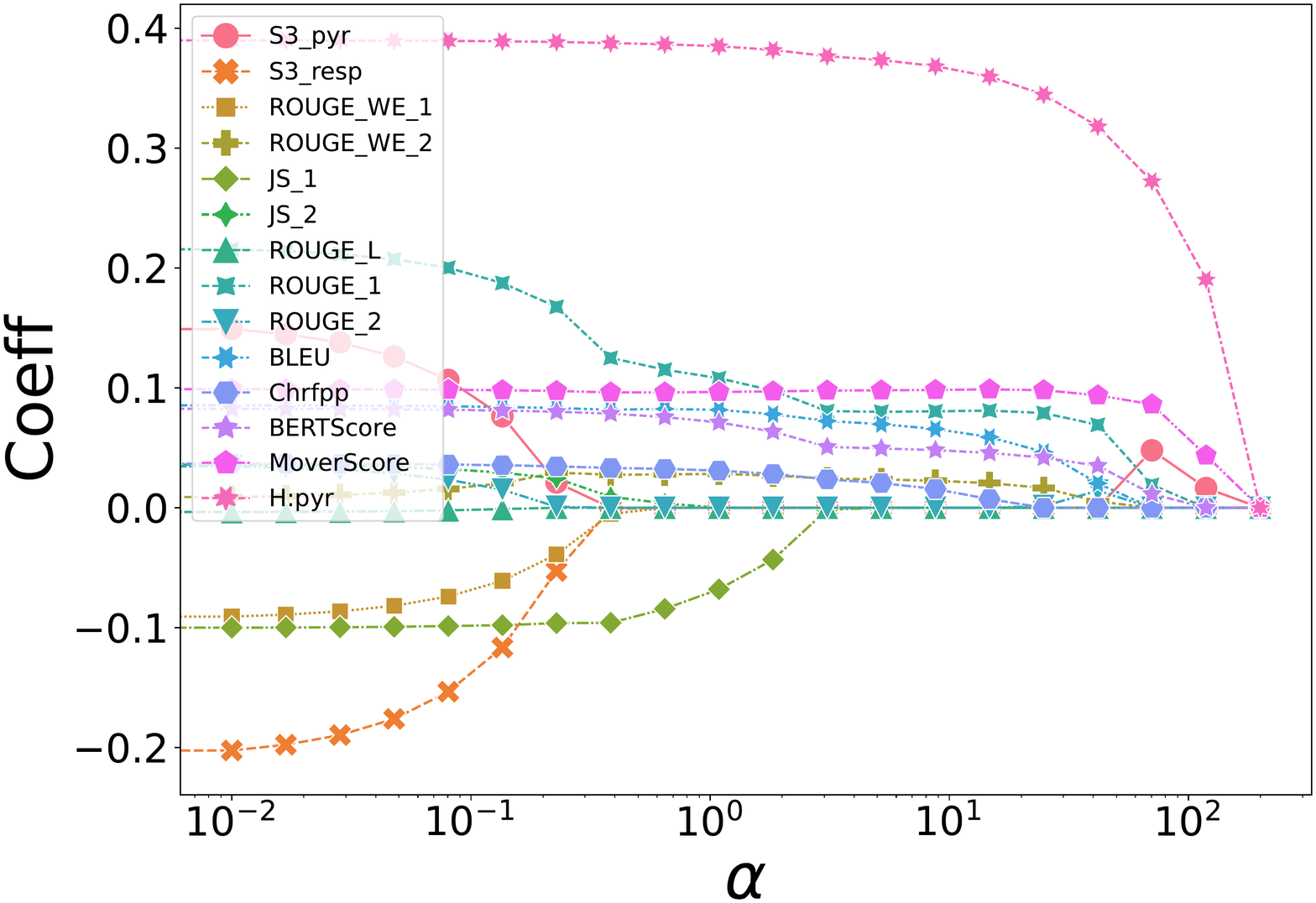}
         \caption{TAC8 Responsiveness}
         \label{fig:y equals x}
     \end{subfigure}
     
     \centering
         \begin{subfigure}[b]{0.3\textwidth}
         \centering
         \includegraphics[width=\textwidth]{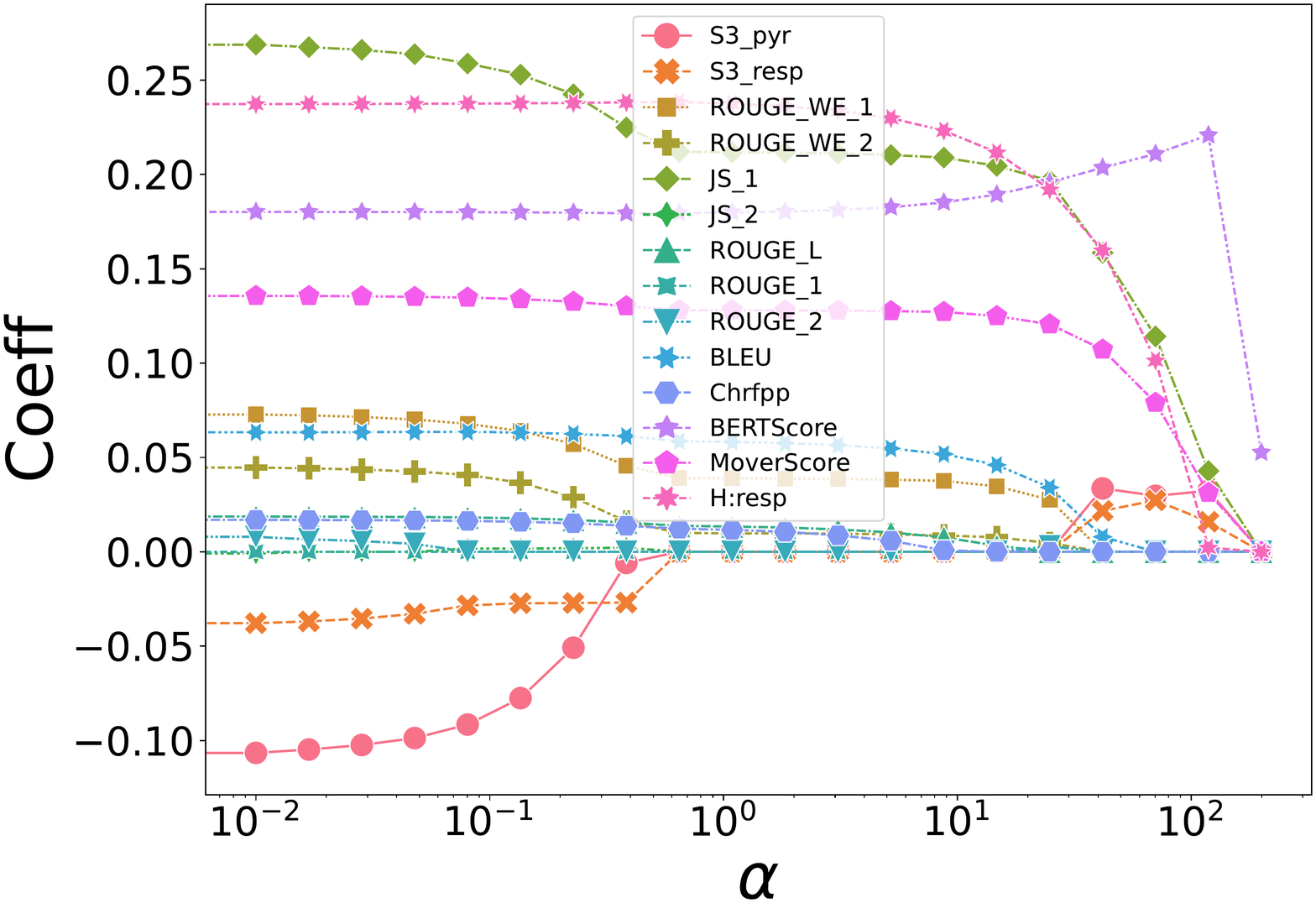}
         \caption{TAC9 Pyr}
         \label{fig:y equals x}
     \end{subfigure}
              \begin{subfigure}[b]{0.3\textwidth}
         \centering
         \includegraphics[width=\textwidth]{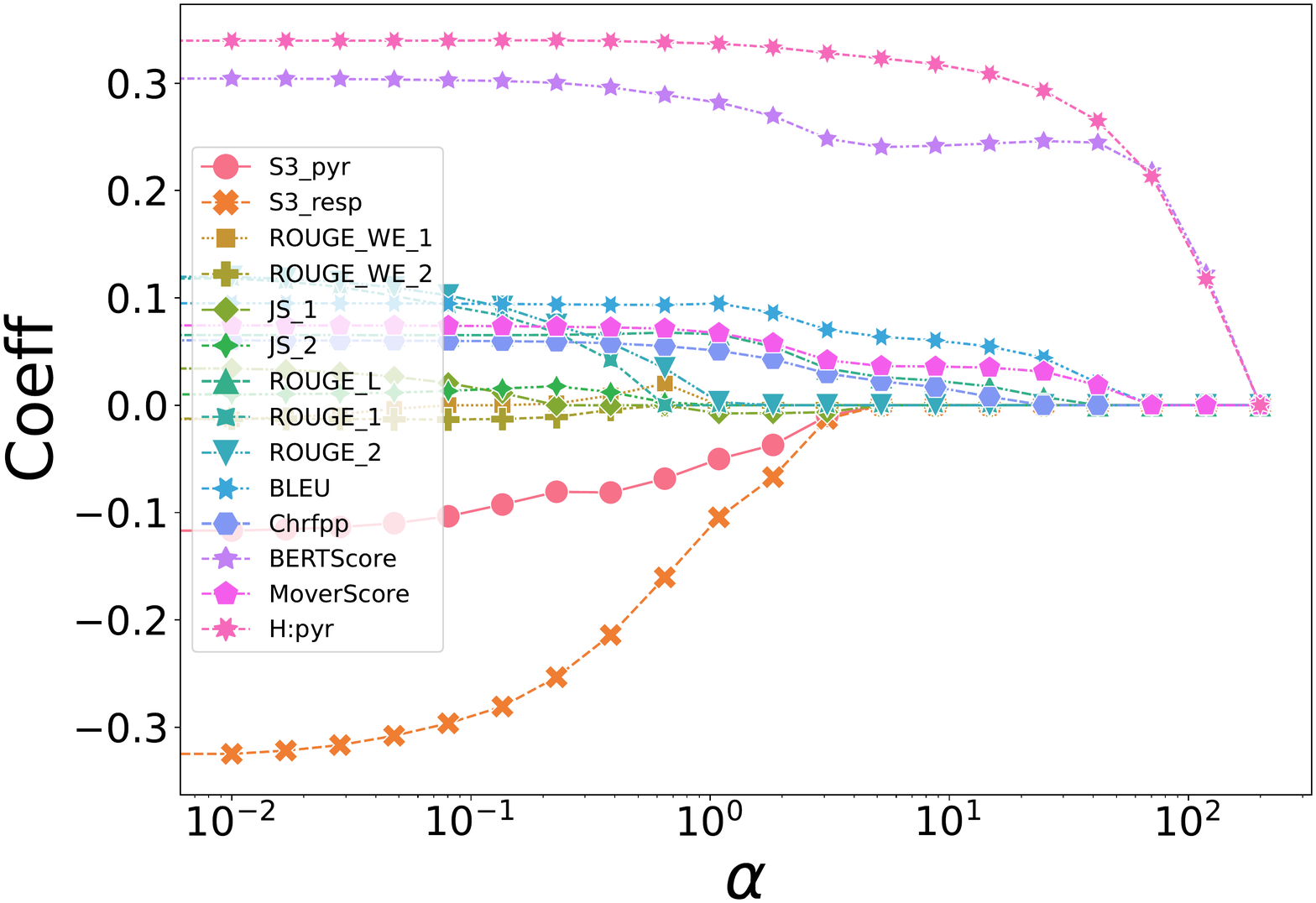}
         \caption{TAC9 Responsiveness}
         \label{fig:y equals x}
     \end{subfigure}
         \begin{subfigure}[b]{0.3\textwidth}
         \centering
         \includegraphics[width=\textwidth]{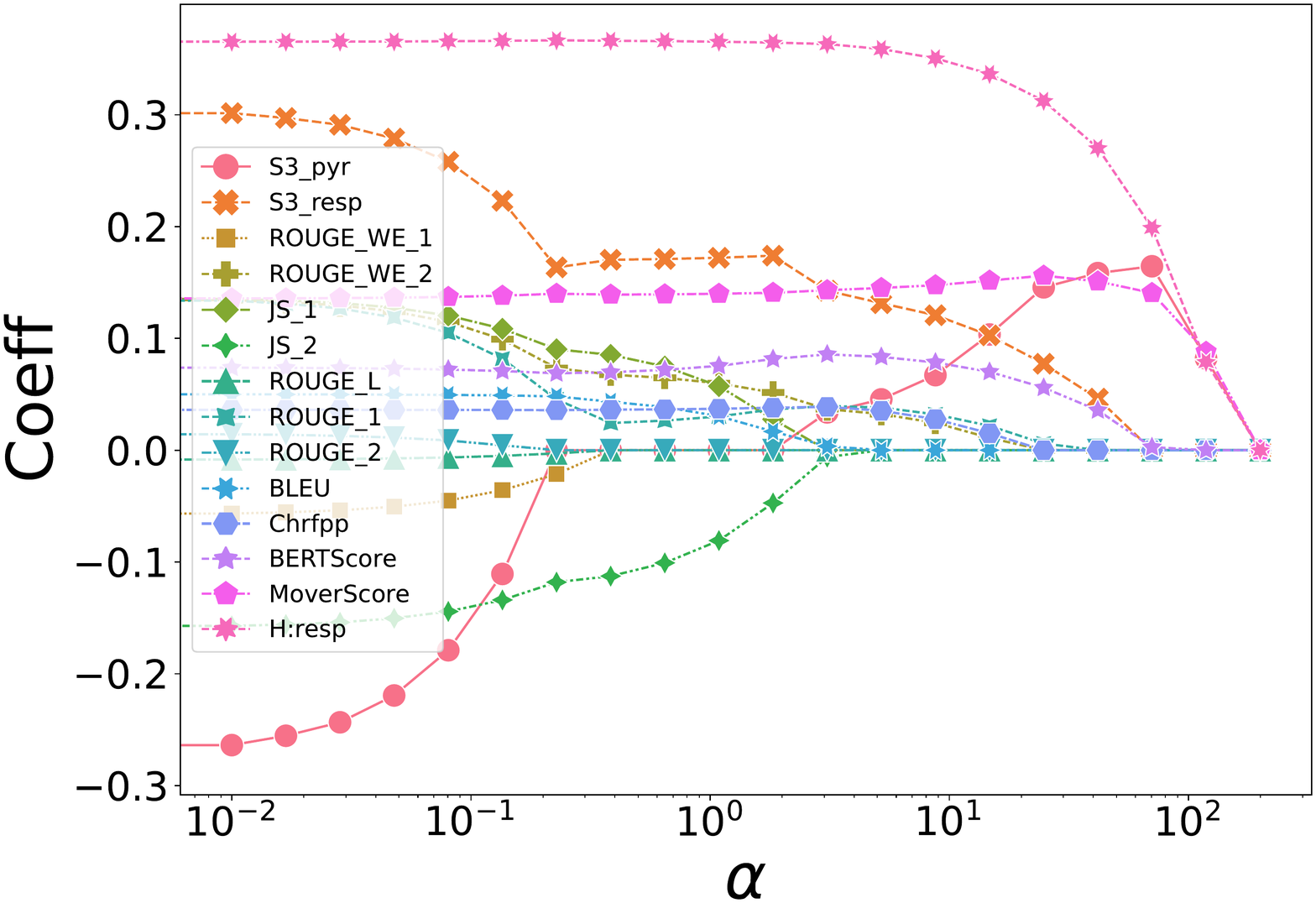}
         \caption{TAC11  Pyramide}
         \label{fig:y equals x}
     \end{subfigure}
     
     \centering
              \begin{subfigure}[b]{0.3\textwidth}
         \centering
         \includegraphics[width=\textwidth]{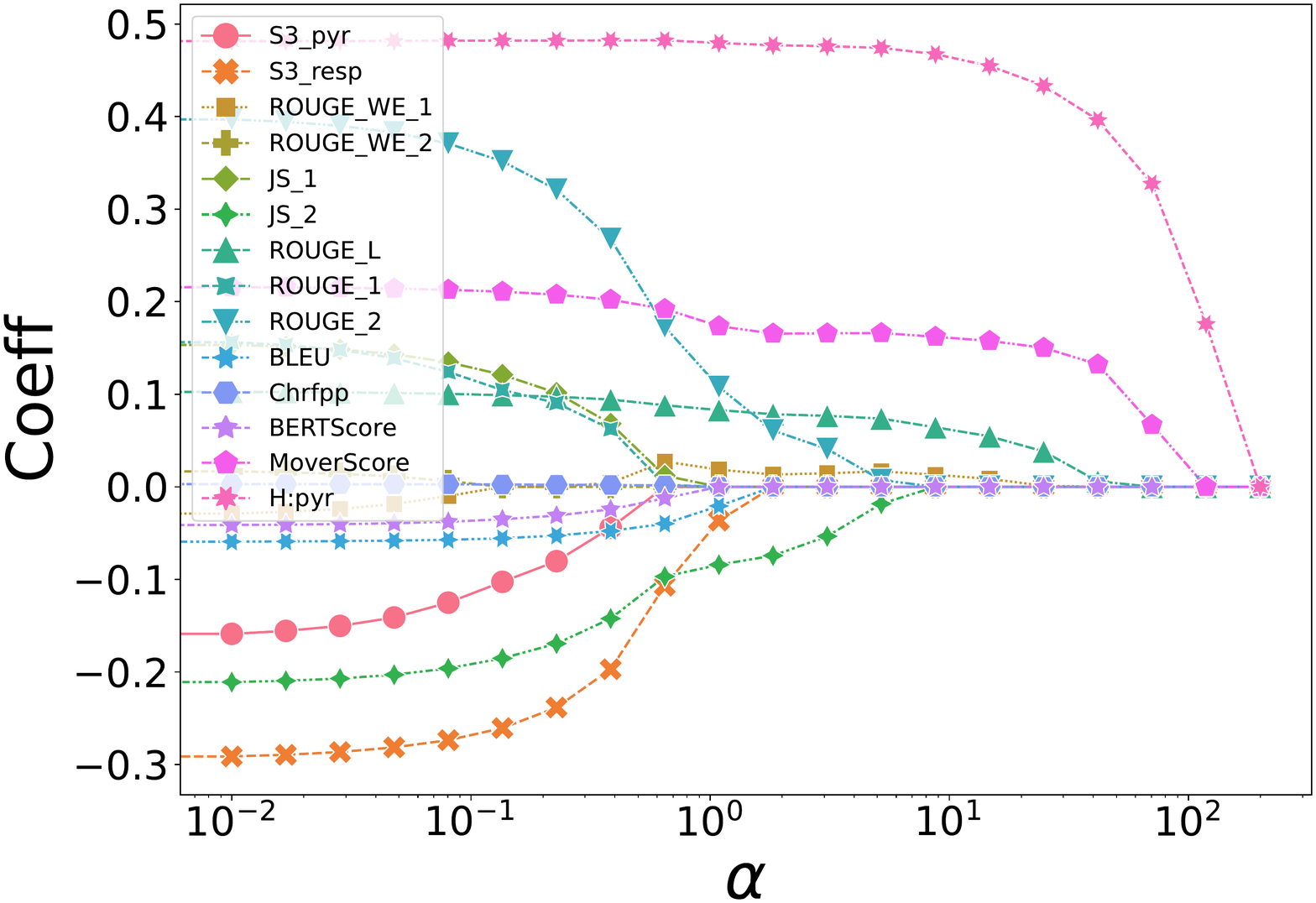}
         \caption{TAC11 Responsiveness}
         \label{fig:y equals x}
     \end{subfigure}
     \caption{\textbf{Human metrics are the most useful metrics when predicting other metrics.} Regression weights (y-axis) obtained by each metric when training a Lasso Regression to predict a human metric for different regularization coefficients (x-axis) on the system level representation of the metrics.
 }\label{fig:system_level_}
\end{figure}

\end{document}